\PassOptionsToPackage{inline}{enumitem}

\documentclass[electronics,article,accept,pdftex,moreauthors]{Definitions/mdpi}%

\firstpage{1} 
\pubvolume{1}
\issuenum{1}
\articlenumber{0}
\pubyear{2025}
\copyrightyear{2025}
\datereceived{ } 
\daterevised{ } %
\dateaccepted{ } 
\datepublished{ } 
\hreflink{https://doi.org/} %
\pdfoutput=1 %

\newlength\stextwidth
\newcommand\makesamewidth[3][c]{%
  \settowidth{\stextwidth}{#2}%
  \makebox[\stextwidth][#1]{#3}%
}

\usepackage{xfrac}

\graphicspath{{./fig/}}

\usepackage{amsthm}
\usepackage{dsfont}

\usepackage{booktabs}
\usepackage{multirow}

\newcommand{\coder}{%
\url{https://github.com/So-Cool/bLIMEy/tree/master/ELECTRONICS_2025}
}%

\newcommand{\coderepo}{\footnote{%
\coder%
}}

\newcommand{\IR}{\ensuremath{\mathit{IR}}}

\DeclareMathOperator*{\argmin}{arg\,min}
\DeclareMathOperator*{\argmax}{arg\,max}

\usepackage[ruled,vlined]{algorithm2e}

\Title{\textsc{LIMEtree}: Consistent and Faithful Surrogate Explanations of Multiple Classes}

\TitleCitation{LIMEtree: Consistent and Faithful Surrogate Explanations of Multiple Classes}

\Author{Kacper Sokol $^{*\dagger}$\orcidA{} and Peter Flach \orcidB{}}%

\AuthorNames{Kacper Sokol and Peter Flach}

\isAPAStyle{%
       \AuthorCitation{Sokol, K., \& Flach, P.}
         }{%
        \isChicagoStyle{%
        \AuthorCitation{Kacper Sokol, and Peter Flach.}
        }{
        \AuthorCitation{Sokol, K.; Flach, P.}
        }
}

\address[1]{%
Intelligent Systems Laboratory, University of Bristol, United Kingdom; \{k.sokol, peter.flach\}@bristol.ac.uk%
}

\firstnote{Current address: Department of Computer Science, ETH Zurich, Switzerland}%

\corres{Correspondence: k.sokol@bristol.ac.uk or kacper.sokol@inf.ethz.ch%
}

\abstract{%
Explainable artificial intelligence provides tools to better understand predictive models and their decisions, but many such %
methods are limited to producing insights with respect to a single class. %
When generating explanations for several classes, reasoning over them to obtain a comprehensive view may be difficult %
since they can present competing or contradictory evidence. %
To address this challenge we introduce the novel paradigm of \emph{multi-class explanations}. %
We outline the theory behind such techniques and propose a local surrogate model based on multi-output regression trees -- called {\sc LIMEtree} -- that offers \emph{faithful} and \emph{consistent} explanations of multiple classes for individual predictions while being post-hoc, model-agnostic and data-universal. %
On top of strong fidelity guarantees, our implementation delivers a range of diverse explanation types, including counterfactual statements favoured in the literature. %
We evaluate our algorithm with respect to explainability desiderata, through quantitative experiments and via a pilot user study, on image and tabular data classification tasks, comparing it to LIME, which is a state-of-the-art surrogate explainer. %
Our contributions demonstrate the benefits of multi-class explanations and wide-ranging advantages of our method across a diverse set of scenarios. %
}


\begin{document}

\section{Introduction\label{sec:intro}}

Explainability of predictive systems based on artificial intelligence (AI) algorithms has become one of their most desirable properties~\cite{rudin2019stop,sokol2021explainability,longo2024explainable}. %
While a wide array of explanation types -- supplemented by numerous techniques to generate them -- is available~\cite{guidotti2018survey}, contrastive statements are dominant~\cite{miller2018explanation,wachter2017counterfactual,poyiadzi2020face,romashov2022baycon,waa2018contrastive,verma2024counterfactual}. %
Their particular realisation in the form of \emph{counterfactual examples} is the most ubiquitous given its everyday usage among humans and solid foundations in social sciences~\cite{miller2018explanation} as well as its compliance with various legal frameworks~\cite{wachter2017counterfactual}. %
Such insights are usually of the form: ``Had certain aspects of the given case been different, the predictive model would behave like so instead.'' %
The conditional part of this proposition usually prescribes a modification of the feature vector of a particular data point, whereas the hypothetical fragment of the statement tends to capture the resulting change in class prediction. %

While offering a very appealing recipe for swaying an automated decision, these explanations are intrinsically restricted to a pair of outcomes, which may impact their utility, effectiveness and comprehensibility. %
They can either highlight an \emph{explicit contrast} between two classes -- ``Why \(A\) rather than \(B\)?'' -- or be \emph{implicit} instead -- ``Why \(A\) (as opposed to anything else)?'' %
As a result, counterfactuals, but also \emph{single-class explanations} more broadly, have been shown to simply \emph{justify} conclusions of AI systems, which may be counterproductive as it implicitly limits the number of possibilities that the explainees consider, thus bias their perception, impede independent reasoning and yield unwarranted reliance on AI or prevent trust from developing altogether~\cite{byrne2023good}. %

In human explainability this limitation can be overcome with follow-up questions, progressively exploring and narrowing down the scope of the lack of understanding until finally eliminating it. %
One could imagine generating multiple counterfactuals across all the possible outcomes to mimic this process, e.g., ``Why \(A\) (and not \(B\) or \(C\))?'', ``Why \(A\) rather than \(B\)?'', ``Why \(A\) instead of \(C\)?'', ``Why \(B\) (and not \(A\) or \(C\))?'', ``Why \(B\) instead of \(A\)?'', etc., for three outcomes \(A\), \(B\) and \(C\). %
Other explainability methods could also be employed in this scenario to provide a wider gamut of insights varying in scope, complexity and explanation target. %
Such approaches embody the recent \emph{hypothesis-driven decision support} conceptualisation of explainable AI (XAI), which aims to provide diverse evidence for data-driven predictions instead of offering a recommendation to simply accept or reject a pre-selected AI decision~\cite{miller2023explainable}; %
this process keeps the explainees engaged instead of displacing them, utilises their expertise, and mitigates over- and under-dependence on automation. %

However, implementing this paradigm with current XAI tools is likely to fall short given that they %
tend to generate independent insights whose one-class limitation prevents them from capturing and communicating a congruent bigger picture. %
The lack of a single origin and shared context may yield insights that do not overlap or are outright contradictory -- e.g., different conditionals used by counterfactuals and disparate pieces of evidence output by other techniques -- preventing the explainees from drawing coherent conclusions and adversely affecting their trust and decision-making capabilities~\cite{weld2018challenge}. %
While a promising research direction, %
to the best of our knowledge the challenge of generating inherently consistent explanations of multiple classes has neither been addressed for counterfactuals nor any other explanation type. %
In this paper we fill this gap by introducing the novel concept of \textbf{multi-class explanations}, where individual insights pertaining to different predictions (of a selected instance) originate from a single explanatory source. %

To this end, we:%
\begin{enumerate}[label=(\roman*)]
    \item define a multi-class explainability optimisation objective;%
    \item operationalise it in the form of a local surrogate;%
    \item offer an algorithm for building multi-class explainers; and%
    \item implement it with \emph{multi-output} regression trees. %
\end{enumerate}
We evaluate our method -- called \textsc{LIMEtree} -- along three dimensions: %
    an \emph{analytical assessment} of human-centred XAI desiderata; %
    a series of \emph{quantitative experiments} on tabular and image data measuring explainer \emph{fidelity}; and %
    a \emph{qualitative user study} capturing explainees' preferences. %
We choose to demonstrate multi-class explainability with a \emph{surrogate}~\citep{craven1996extracting} since this design yields an explainer that is %
\emph{post-hoc} -- i.e., capable of being retrofitted to pre-existing AI systems -- \emph{model-agnostic} -- i.e., compatible with any predictive algorithm -- and \emph{data-universal} -- i.e., suitable for tabular, text and image domains. %
Additionally, by using a \emph{(binary) decision tree}~\citep{breiman1984classification} as the surrogate, %
{\sc LIMEtree} offers a broad range of explanation types such as model structure visualisation, feature importance, exemplars, logical rules, what-ifs and, most importantly, \emph{counterfactuals}~\cite{sokol2021towards}. %
This suite of investigative mechanisms supports diverse explanation scopes spanning model simplification, sub-space approximation and prediction rationales. %

{\sc LIMEtree} offers solutions to many shortcomings of currently available surrogate explainers in addition to addressing limitations found across the social and technical dimensions of XAI~\citep{saeed2023explainable}. %
Specifically, by using (shallow) binary regression trees as surrogate models, it can guarantee \emph{full fidelity} of the explanations with respect to the investigated black box under certain conditions, thus addressing one of the major criticisms of post-hoc approaches~\cite{rudin2019stop,retzlaff2024post}. %
The flexible explanation generation process additionally enables it to comply with a range of desiderata such as feasibility and actionability~\cite{poyiadzi2020face} as well as facilitate algorithmic recourse~\cite{karimi2021algorithmic}, to name just a few~\cite{sokol2020explainability,guidotti2024counterfactual}. %
The availability of multiple diverse explanation types also allows it to provide explainability to a broad range stakeholders and satisfy their diverse needs~\citep{meske2022explainable}. %
With all of these contributions we hope to launch multi-class explainability as a novel, highly beneficial XAI research direction. %

\section{Related Work and Background\label{sec:background}}

{\sc LIMEtree} builds upon two prominent findings in XAI: \emph{counterfactuals}~\cite{miller2018explanation,wachter2017counterfactual} and \emph{surrogate explainers}~\cite{craven1996extracting,ribeiro2016should,sokol2019blimey,sokol2020towards,sokol2021towards,sokol2022what}. %
As noted earlier, the former are lauded for their \emph{human-centred} aspects, and the latter exhibit numerous appealing \emph{technical} properties, making them one of the most flexible type of explainers. %
In a nutshell, surrogates mimic the behaviour of more complex, hence opaque, predictive systems either locally or globally with simpler, inherently interpretable models, thereby offering human-comprehensible insights into their operation~\cite{craven1996extracting,ribeiro2016should}. %
Unlike surrogates and counterfactuals, \emph{multi-class explainability} is a largely under-explored topic. %
While counterfactual explanations can be generated for multiple classes~\cite{carlevaro2023multi}, such insights may not present a coherent perspective given that they can be conditioned on different sets of features. %
One of the very few pieces of work, if not the only, that directly addresses this challenge %
expands Generalised Additive Models (GAMs~\cite{hastie1986generalized}) -- which are inherently transparent and powerful predictors popular in high stakes domains~\cite{lou2012intelligible} -- to multiple classes~\cite{zhang2019axiomatic}. %

LIME~\cite{ribeiro2016should} is one of the most popular surrogate approaches; it uses sparse linear regression to explain (probabilistic) black-box predictions. %
It augments the classic paradigm of surrogate explainers with \emph{interpretable representations} (IR) of raw data, making them compatible with a variety of data domains (such as images and text) %
and extending their applicability beyond inherently interpretable features (of tabular data). %
High modularity and flexibility of these explainers~\cite{sokol2019blimey} encouraged the research community to compose their different variants, some of which use decision trees as the (local) surrogate model~\cite{waa2018contrastive,shi2019explaining,sokol2020towards,sokol2021towards}. %
For example, \citet{waa2018contrastive} showed how a local one-vs-rest classification tree can be used to produce contrastive explanations; and \citet{shi2019explaining} fitted a local shallow regression tree whose structure constitutes an explanation. %
Interpretability of decision trees and their ensembles has also been investigated outside of the surrogate explainability context~\cite{tolomei2017interpretable,sokol2018glass,sokol2020one,sokol2021towards}. %
\citet{sokol2018glass,sokol2020one} demonstrated how to interactively extract personalised counterfactuals from a decision tree; and \citet{tolomei2017interpretable} introduced a method to explain predictions made by tree ensembles also with counterfactuals. %

More specifically, %
LIME builds a local surrogate model \(g \in \mathcal{G}\) to explain the prediction of an instance \(\mathring{x} \in \mathcal{X}\) with respect to a selected class \(c\) for a \emph{probabilistic} black box \(f: \mathcal{X} \mapsto \mathcal{Y}\), where \(\mathcal{G}\) is the space of (sparse linear) surrogate models, \(\mathcal{X}\) is the input data domain, \(\mathcal{Y}\) is the space of \(n\)-dimensional probability vectors, \(n \in \mathcal{N}^+\) is the number of target classes, and \(c \in [1, \ldots, n]\). %
To this end, it employs %
a \emph{user-defined} interpretable representation transformation function \(\IR: \mathcal{X} \mapsto \mathcal{X}^\prime\), which encodes presence (\(1\)) and absence (\(0\)) of \(d \in \mathcal{N}^+\) selected human-comprehensible concepts found in a data point \(x \in \mathcal{X}\), i.e., \(\mathcal{X}^\prime = \{0, 1\}^d\). %
Additionally, \(\IR\) is defined such that the explained instance is assumed to have all of the concepts present, i.e., \(\IR(\mathring{x}) = \mathring{x}^\prime = [1,\ldots,1]\), which is an all-\(1\) vector. %
This step allows us to generate ``conceptual'' variations of \(\mathring{x}\) by drawing a collection of binary vectors \(X^\prime = \{x^\prime : x^\prime \in \mathcal{X}^\prime\}\). %

Next, \(X^\prime\) is converted back to the original data domain \(\mathcal{X}\) using the inverse of the interpretable representation transformation function \(\IR^{-1}: \mathcal{X}^\prime \mapsto \mathcal{X}\), i.e., \(X = \{\IR^{-1}(x^\prime) : x^\prime \in X^\prime\}\), which facilitates predicting these instances with the explained black box \(f\), focusing on the probabilities of the selected class \(c\), i.e., \(Y_{c} = \{ f_{c}(x) : x \in X\} \). %
These predictions capture the influence of (the presence of) each human-comprehensible concept on the (change in) prediction of class \(c\). %
We can quantify this dependence by fitting \emph{sparse linear regression} to the binary sample \(X^\prime\) and probabilities \(Y_{c}\). %
This procedure can be focused on a specific aspect of the data sample by computing its \emph{distance} \(\ell\) to the explained instance either in the original or interpretable representation -- i.e., \(\ell : \mathcal{X} \times \mathcal{X} \mapsto \mathbb{R}\) or \(\ell : \mathcal{X}^\prime \times \mathcal{X}^\prime \mapsto \mathbb{R}\) -- then transformed into a similarity measure by passing it through a \emph{kernel} \(\kappa : \mathbb{R} \mapsto \mathbb{R}\) and used as weight factor for training the surrogate model. %
This step allows to prioritise smaller changes to the instance, e.g., give more significance to samples with fewer alterations in the concept space. %

LIME optimises \emph{fidelity} of the surrogate, i.e., its ability to approximate the predictive behaviour of the explained black box, and \emph{complexity} of the resulting explanation, i.e., its hu\-man-com\-pre\-hen\-si\-bil\-i\-ty; %
this objective \(\mathcal{O}\) is formalised in Equation~\ref{eq:lime:objective}. %
Complexity \(\Omega\), in case of linear models, is computed as the number of non-zero (or significantly larger than zero) coefficients \(\Theta_g\) of the surrogate \(g\) -- see Equation~\ref{eq:lime:complexity}. %
High \emph{fidelity} entails small empirical loss \(\mathcal{L}\) -- Equation~\ref{eq:lime:loss} -- calculated between the outputs of the black box \(f\) and the surrogate \(g\) %
using data sampled ``around'' the explained instance. %
Individual loss components are weighted by similarity scores -- \(\omega(x ; \; \mathring{x})\) for \(x \in X\) or \(\omega(x^\prime ; \; \mathring{x}^{\prime})\) for \(x^{\prime} \in X^{\prime}\) depending on the domain -- derived by kernelising distance between the explained instance and sampled data. %
This loss is inspired by \emph{Weighted Least Squares}, where the weights are similarity scores. %

\begin{equation}\label{eq:lime:objective}
    \mathcal{O}(\mathcal{G}; \; f) = %
    \argmin_{g \in \mathcal{G}} %
    \underbrace{\mathcal{L}(f, g)}_{\text{\makesamewidth[c]{XXXXXXXX}{fidelity}}} %
    + %
    \underbrace{\Omega(g)}_{\text{\makesamewidth[c]{XXXXXXXX}{complexity}}}
\end{equation}
\begin{equation}\label{eq:lime:complexity}
    \Omega(g) = \sum_{\theta \in \Theta_g} \mathds{1}(\vert\theta\vert > 0) \; / \; \vert\Theta_g\vert %
\end{equation}
    \begin{align}\begin{split}
    \label{eq:lime:loss}%
        \mathcal{L}(f, g; \; X^\prime, \mathring{x}, c)  = & %
        \frac{1}{\sum_{x^\prime \in X^\prime} \omega\left(x^\prime; \; \IR(\mathring{x})\right)} \\ %
        & \sum_{x^\prime \in X^\prime} %
        \omega\left(x^\prime; \; \IR(\mathring{x})\right) \; %
        \left(f_{c}\left(\IR^{-1}(x^\prime)\right) - %
        g(x^\prime)\right)^2 \\
        & \text{where} \quad %
    \omega(x^\prime ; \; \mathring{x}^{\prime}) = \kappa\left(\ell\left(x^{\prime}, \; \mathring{x}^\prime\right)\right)%
    \end{split}\end{align}

\begin{figure}[t]
    \centering
\begin{adjustwidth}{-\extralength}{0cm}
        \subfloat[Interpretable representation.]{%
\makebox[0.2285\fulllength][c]{%
    \includegraphics[height=0.2244\fulllength]{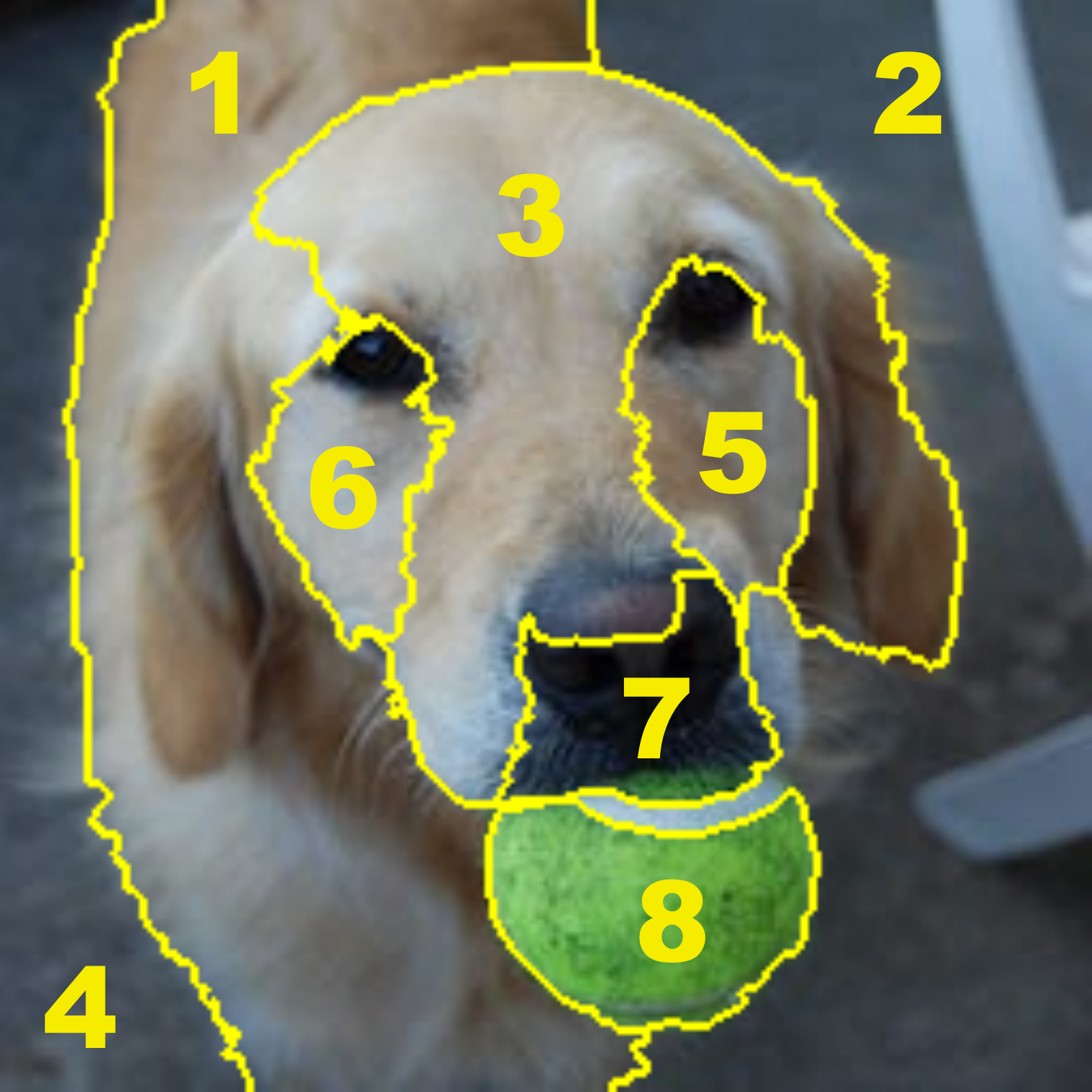}%
}
    \label{fig:lime:image}%
    }
    \hspace{0.03025\textwidth}%
        \subfloat[\emph{Tennis ball} (99.28\%).]{%
\makebox[0.2285\fulllength][c]{
    \includegraphics[height=0.2244\fulllength]{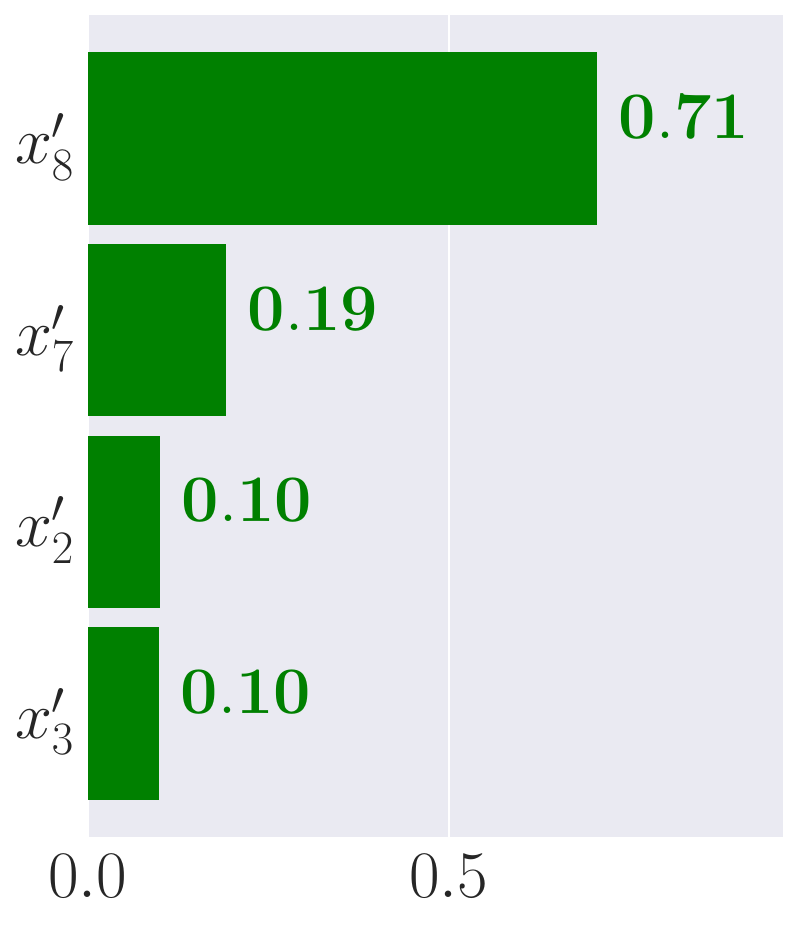}%
}
    \label{fig:lime:ball}%
    }
    \hspace{0.03025\textwidth}%
        \subfloat[\emph{Golden retriever} (0.67\%).]{%
\makebox[0.2285\fulllength][c]{
    \includegraphics[height=0.2244\fulllength]{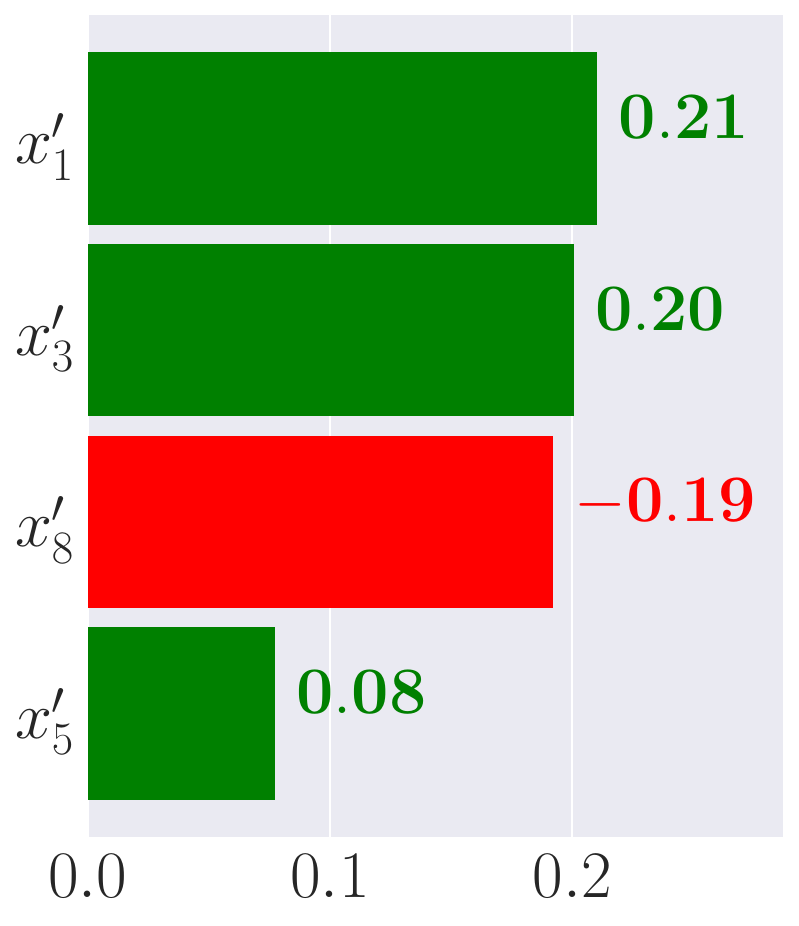}%
}
    \label{fig:lime:labrador}%
    }
    \hspace{0.03025\textwidth}%
        \subfloat[\emph{Labrador retriever} (0.04\%).]{%
\makebox[0.2285\fulllength][c]{
    \includegraphics[height=0.2244\fulllength]{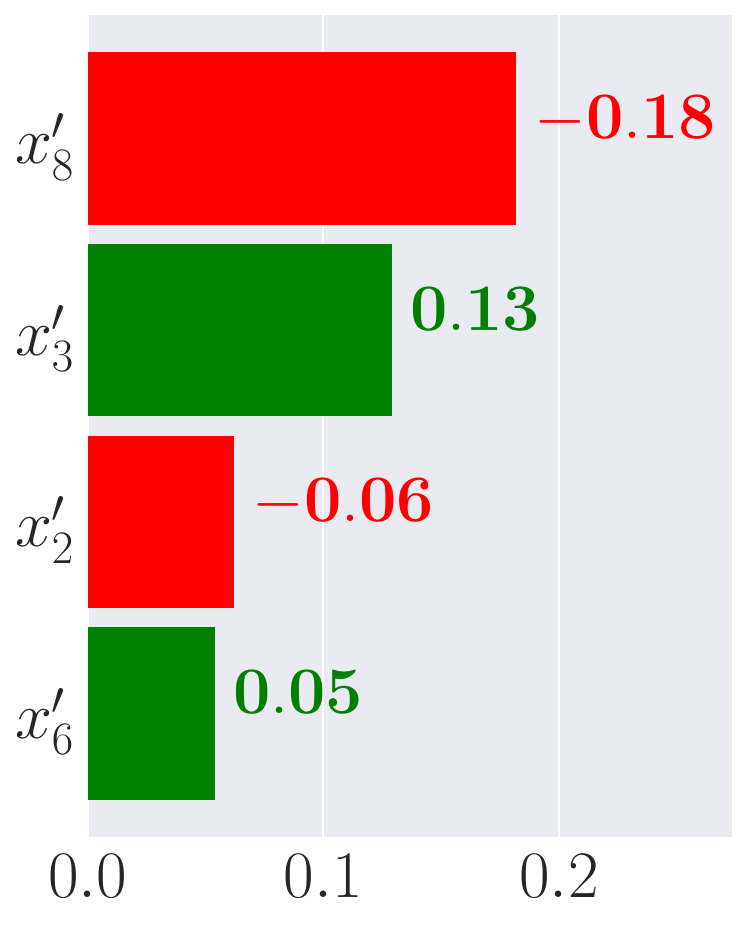}%
}
    \label{fig:lime:golden}%
    }
\end{adjustwidth}
    \caption{%
    LIME explanations for the top three classes predicted by a black-box model. %
    Panel~\protect\subref*{fig:lime:image} shows the super-pixel interpretable representation of the explained image with \(d=8\) segments. %
    Panels~\protect\subref*{fig:lime:ball}, \protect\subref*{fig:lime:labrador} and \protect\subref*{fig:lime:golden} are LIME explanations; %
    they capture the positive or negative influence of (the presence of) interpretable features  %
    on the prediction (probability) of a selected class.%
    }%
    \label{fig:lime}%
\end{figure}

The precise definitions of the interpretable representation transformation function \(\IR\) and its inverse \(\IR^{-1}\) depend on the data domain. %
For \emph{text}, \(\IR\) splits it into \(d\) tokens, e.g., using the bag-of-words approach, whose presence (\(1\)) or absence (\(0\)) is encoded by \(\mathcal{X}^{\prime}\); %
setting a component of this domain to \(0\) is thus equivalent to removing a token from a text excerpt. %
For \emph{images}, this domain transformation %
relies on super-pixel partition of a picture into \(d\) non-overlapping patches whose binary vector encoding indicates whether a particular segment is \emph{preserved} (\(1\)) or \emph{discarded} (\(0\)); %
since parts of an image cannot be removed directly, an \emph{occlusion proxy} that replaces selected patches with a predetermined colour is used. %
Figure~\ref{fig:lime} shows an interpretable representation of an image and its LIME explanations for the top three predictions. %
LIME explanations of text follow a similar pattern with it being split into tokens (its interpretable representation) whose influence on a prediction -- e.g., the positive or negative sentiment of a sentence -- is quantified through the coefficients of the corresponding surrogate linear model. %

For \emph{tabular data}, the \(\IR\) function is more complex; %
continuous features are first discretised and then, together with any remaining categorical attributes, binarised. %
The latter step assigns, separately for every feature, \(1\) to the discrete partition where the explained instance is located, with all the other partitions merged and represented by \(0\). %
As a result, %
the mapping between \(\mathcal{X}^{\prime}\) and \(\mathcal{X}\) tends to be \emph{non-deterministic}, unlike the corresponding \(\IR^{-1}\) transformation for image and text data. %
Further information about surrogate explainers -- including their generalisation and in-depth analysis of their individual building blocks in the context of text, image and tabular data domains -- can be found in the literature~\cite{sokol2019blimey,sokol2022what,sokol2020towards,sokol2021towards}. %

\section{\textsc{LIMEtree}\label{sec:limetree}}%

LIME fits
a separate surrogate model to the probabilities of each class of interest. %
This makes the process of discovering the dependencies between multiple classes challenging as each explanation needs to be interpreted in isolation. %
A surrogate fitted to class \(A\) is implicitly a \emph{one-vs-rest} explainer since it can only answer questions about the probability of this single class, with the complementary probability \(p(\neg A) = 1 - p(A)\) modelling the union of all the other classes \(\neg A \equiv B \cup C \cup \cdots\). %
Interpreting the magnitude of the probability \(p(A)\) output by a surrogate trained for class \(A\) can also be problematic when explaining multi-class black boxes. %
For example, if \(p(A) \leq 0.5\), we cannot be certain whether there is a single class \(B\) with \(p(B) > p(A)\), or alternatively the combined probability of all the complementary classes \(p(\neg A)\) is greater than or equal to \(p(A)\) with no single class dominating over \(p(A)\). %

Moreover, linear predictors -- thus such surrogates as well -- are unable to model target variables that are \emph{non-linear} with respect to input features~\citep{flach2012machine} -- a property that does not necessarily hold for high-level features such as the concepts encoded by IRs~\citep{sokol2020towards}. %
Their high inter-dependence %
may also have adverse effects on explanation quality. %
Additionally, modelling probabilities with linear regression risks confusing the explainees who expect an output bounded between \(0\) and \(1\) but may be given a numerical prediction outside of this range. %

\begin{figure}[t]
    \centering
	\begin{adjustwidth}{-\extralength}{0cm}
    \includegraphics[width=1.0\fulllength]{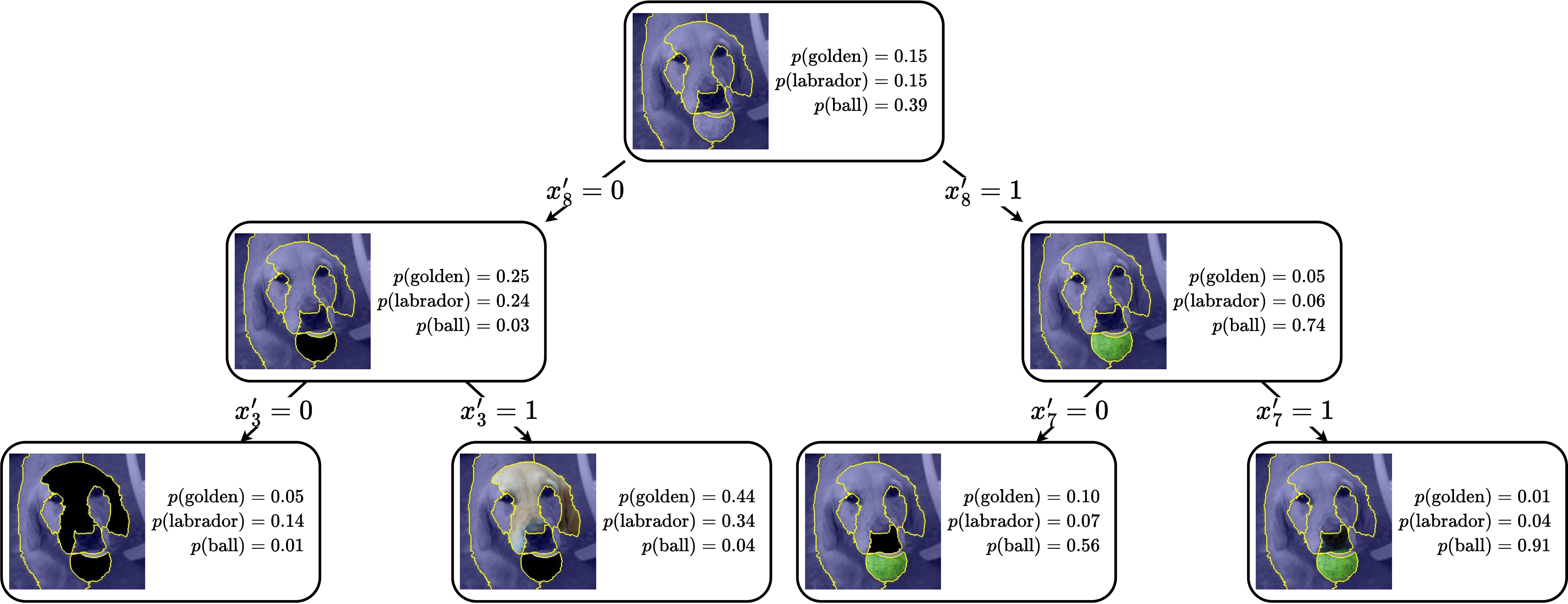}%
	\end{adjustwidth}
    \caption{%
Surrogate multi-output binary regression tree explaining the top three classes -- \emph{tennis ball}, \emph{golden retriever} and \emph{Labrador retriever} -- predicted by a black box for the image shown in Figure~\ref{fig:lime:image}. %
The segments marked in \emph{blue} do not influence the explanation at a given tree node, i.e., they can either be preserved or discarded for the explanation to hold. %
Super-pixels whose value in the interpretable representation is \(1\) are preserved and those with \(0\) are ``removed'' by occluding them with black patches. %
The class probabilities estimated by each node of the surrogate tree may not sum up to \(1\) as these values capture a subset of the modelled classes and are a result of numerical regression, hence they should not be treated as probabilities per se.%
\label{fig:mort}}
\end{figure}

We address the challenge of simultaneously explaining multiple predicted classes of an instance output by a probabilistic model by proposing a first-of-a-kind surrogate explainer based on \textbf{binary multi-output regression trees}. %
It facilitates \emph{multi-class} modelling in a \emph{regression} setting, allowing the surrogate to capture the \emph{interactions} between multiple classes, hence explain them coherently. %
Each node of such a tree approximates the probabilities of every explained class -- a level of detail that is impossible to achieve with surrogate \emph{multi-class classifiers} -- thus reflecting how individual interventions in the interpretable domain affect the predictions. %
Figure~\ref{fig:mort} shows an %
example of a surrogate multi-output regression tree. %
This is a significant improvement over training a separate regression surrogate for each explained class, which may produce diverse, inconsistent, competing or contradictory explanations -- thus risk confusing the explainees and put their trust at stake -- whenever these models do not share a common tree structure or split on different feature subsets. %
Our contributions establish a new direction in XAI research -- concerned with consistent and faithful explanations of multiple classes -- and offer a pioneering method to address this challenge.%

Moreover, employing decision trees as surrogates overcomes the shortcomings identified when linear models are used to this end~\cite{sokol2019blimey,sokol2020towards,sokol2021towards}. %
Trees neither presuppose independence of features nor existence of a linear relationship between them and the target variable~\citep{flach2012machine}. %
While surrogate regression trees that approximate the probability of a single class are guaranteed to output a number within the \([0, 1]\) range -- since the estimate is calculated as an average (default tree behaviour) -- this may not necessarily hold for multi-output trees. %
Approximating probabilities of multiple classes by calculating the mean of their respective predictions across a number of instances may yield averages whose sum is greater than \(1\), nonetheless these values can be rescaled to avoid confusing the explainees.%

While surrogates based on linear models are limited to %
(interpretable) feature influence explanations %
-- see Figure~\ref{fig:lime} -- employing trees offers a broad selection of diverse explanation types. %
These include: %
\begin{enumerate*}[label=(\arabic*)]
\item
    visualisation of the tree structure; %
\item
    tree-based (interpretable) feature importance (\emph{Gini importance}~\cite{breiman2001random}); %
\item
    logical conditions extracted from root-to-leaf paths; %
\item
    exemplar explanations taken from the training data assigned to the same leaf; %
\item
    answers to what-if questions generated based on the tree structure (e.g., by querying the model); and %
\item
    \textbf{counterfactuals} retrieved by comparing and applying logical reasoning to different tree paths~\cite{sokol2021towards}. %
\end{enumerate*}
The first two explanation types uncover the behaviour of a black box in a given data sub-space; the remainder targets specific predictions. %
Since all the six explanation types -- see Section~\ref{sec:discussion} for their examples -- are derived from a single (surrogate) model, they are guaranteed to be coherent and their diversity should appeal to a wide range of audiences. %

To ensure low complexity and high fidelity of our multi-output regression trees, we employ the optimisation objective \(\mathcal{O}\) from Equation~\ref{eq:lime:objective}. %
Since we are using surrogate trees, we modify the model complexity function \(\Omega\) to measure the \emph{depth} or \emph{width} (number of leaves) of the tree as given by Equation~\ref{eq:blimey_complexity}, where \(d\) is the dimensionality of the binary interpretable domain \(\mathcal{X}^\prime\). %
This choice depends on the type of the explanation that we want to extract from the surrogate tree; e.g., depth may be preferred when visualising the tree structure or extracting decision rules. %
In some cases, such as unbalanced trees, optimising for width or a mixture of the two may be more desirable. %
We also adapt the loss function \(\mathcal{L}\) to account for the surrogate tree \(g\) outputting multiple values in a single prediction as shown in Equation~\ref{eq:blimey_loss}, where \(C \subseteq [1,\ldots,n]\) are the classes to be explained by \(g\), for which the \(c\) subscript in \(g_c(x^\prime)\) indicates the prediction of a selected class \(c \in C\) for the data point \(x^\prime\).%

\begin{equation}\label{eq:blimey_complexity}
    \Omega(g; \; d) = \frac{\text{depth}(g)}{d}%
    \qquad \text{or} \qquad%
    \Omega(g; \; d) = \frac{\text{width}(g)}{2^d}%
\end{equation}

    \begin{align}\begin{split}
    \label{eq:blimey_loss}
    \mathcal{L}(f, g; \; X^\prime, \mathring{x}, C) = & %
    \frac{1}{\sum_{x^\prime \in X^\prime} \omega(x^\prime; \; \IR(\mathring{x}))} %
    \\
    & \sum_{x^\prime \in X^\prime} \left(%
    \frac{%
    \omega(x^\prime; \; \IR(\mathring{x})) %
    }{1 + \mathds{1}( \vert C \vert > 1 )} \; %
    \sum_{c \in C}
    \left(f_c\left(\IR^{-1}(x^\prime)\right) - %
    g_c(x^\prime)\right)^2 %
    \right)%
    \end{split}\end{align}

\begin{figure}[t]%
    \centering
\begin{adjustwidth}{-\extralength}{0cm}
        \includegraphics[width=\fulllength]{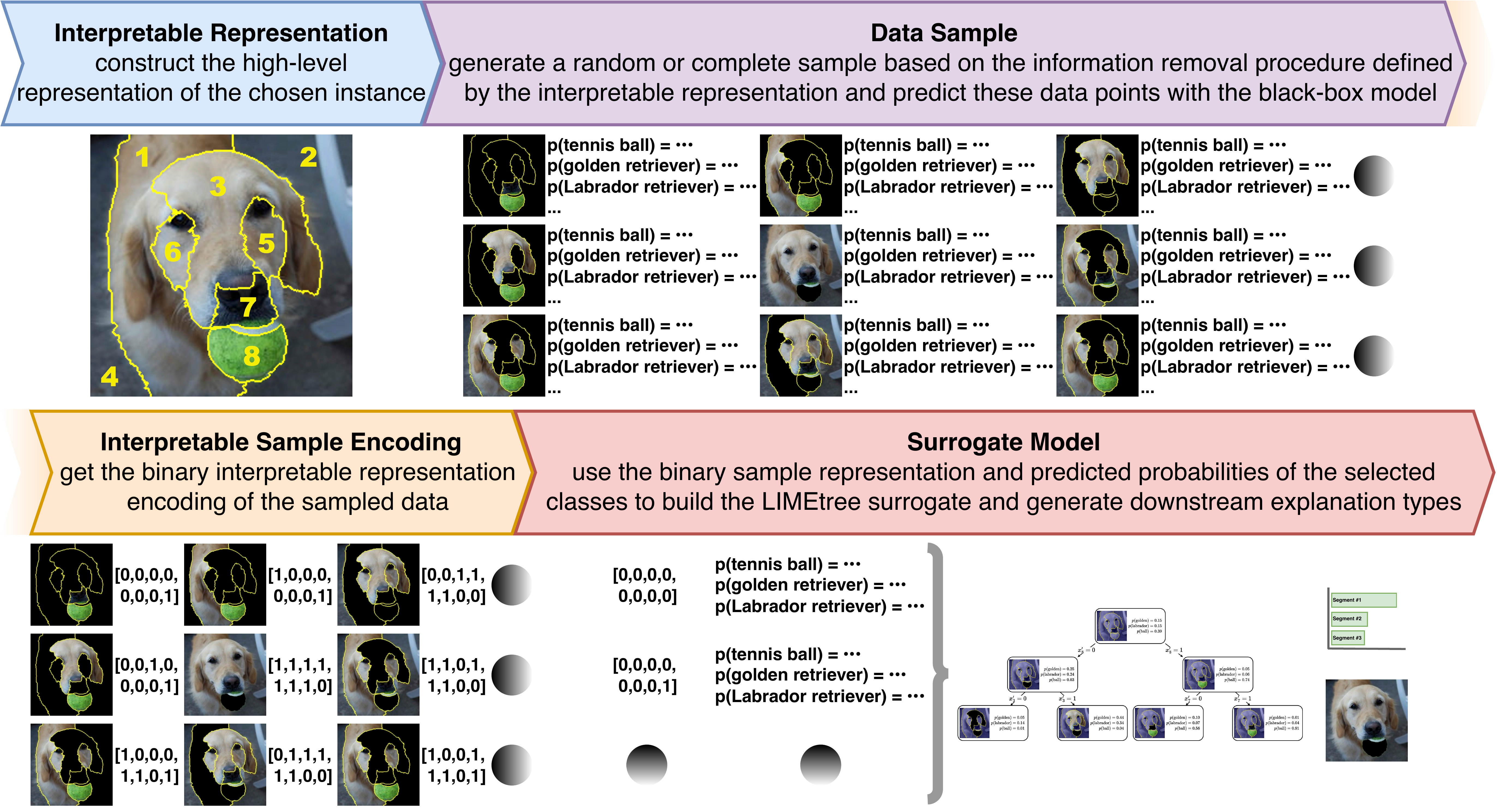}%
\end{adjustwidth}
        \caption{High-level overview of {\sc LIMEtree}.\label{fig:limetree_overview}}%
\end{figure}

Note that the inner sum over the explained classes \(\sum_{c \in C}\) is normalised by \((1 + \mathds{1}( \vert C \vert > 1 ))^{-1}\), which is \(1\) when the surrogate is built for a single class and becomes \(\sfrac{1}{2}\) for more classes. %
The loss given by Equation~\ref{eq:blimey_loss} is thus equivalent to the one in Equation~\ref{eq:lime:loss} in the former case, and in the latter the scaling factor ensures that the inner sum is between \(0\) and \(1\) since the biggest squared difference is \(2\), which happens when the predictions of \(f\) and \(g\) assign a probability of \(1\) to two different classes, e.g., \([1, 0, 0]\) and \([0, 0, 1]\). %
An additional assumption is that the sum of values predicted by each leaf of the surrogate tree is at most \(1\), which as noted earlier may in some cases require normalisation. %
In practice, the surrogate explainer is built by iteratively adding splits to a multi-output regression tree -- thus incrementally increasing its complexity \(\Omega(g; \; d)\) but also improving its predictive power -- which allows to progressively minimise the loss \(\mathcal{L}\) and optimise the objective \(\mathcal{O}\). %
This procedure %
-- captured by Algorithm~\ref{algo:LIMEtree} given in Appendix~\ref{apx:algos} -- %
terminates when the loss \(\mathcal{L}\) (calculated with Equation~\ref{eq:blimey_loss}) reaches a certain, user-defined level \(\epsilon \in [0, 1]\), which corresponds to the fidelity of the local surrogate, i.e., \(\mathcal{L}\left(f, g; \; X^\prime, \mathring{x}, C\right) \leq \epsilon\). %
Figure~\ref{fig:limetree_overview} provides a high-level overview of {\sc LIMEtree}. %

\section{Fidelity Guarantees\label{sec:fidelity}}%

The flexibility of surrogate explainers -- they are post-hoc, model-agnostic and, often, data-universal -- also contributes to the instability and occasional unreliability of their explanations~\cite{laugel2018defining,zhang2019should,sokol2020towards,sokol2019blimey,sokol2022what}. %
Their subpar \emph{fidelity}, i.e., predictive coherence with respect to the explained black box, is thus a major barrier for their uptake~\cite{rudin2019stop}. %
In addition to remedying the shortcomings of linear surrogates, {\sc LIMEtree} comes with strong fidelity guarantees, which can be achieved in practice while preserving low explanation complexity. %

To imbue {\sc LIMEtree} with near-full or full fidelity, %
we identify the \emph{minimal} IR set \(X^\prime_{\textit{min},T} \subseteq \mathcal{X}^\prime\). %
It is unique to a tree \(T\) and composed of binary vectors \(x^\prime_{\textit{min},t}  \) drawn from the IR -- one per leaf \(t \in T\) of the surrogate tree -- that have the least number of \(0\) components while still being assigned to the leaf \(t\). %
The construction of this set is formalised in Definition~\ref{def:minimal_x_star} and can be understood as seeking instances with the highest number of human-interpretable concepts being present, e.g., minimal occlusion for images, for each leaf. %

\begin{Definition}[Minimal Representation]\label{def:minimal_x_star}
    Assume a binary decision tree \(g \in \mathcal{G}\) fitted to a binary \(d\)-dimensional data space \(\mathcal{X}^\prime = \{0, 1\}^d\), with \(T\) denoting its set of leaves. %
    This tree assigns a leaf \(t \in T\) to a data point \(x^\prime \in \mathcal{X}^\prime\) with the function \(g_\mathit{id}(x^\prime) = t\). %
    For a given tree leaf \(t\), its \emph{unique} \emph{minimal} data point \(x^\prime_{\textit{min}, t}\) is given by %
    \[%
    x^\prime_{\textit{min}, t} = \argmax_{x^\prime \in \mathcal{X}^\prime} \sum_{i=1}^d x^\prime_i%
    \quad\text{s.t.}\quad%
    g_\mathit{id}(x^\prime) = t%
    \text{,}%
    \]%
    where \(x^\prime_i\) is the \(i\)\textsuperscript{th} component of the binary vector \(x^\prime\). %
    We can further define a \emph{minimal} set of data points \(X^\prime_{\textit{min}, T} \subseteq \mathcal{X}^\prime\) -- uniquely representing a tree \(g\) and the set of its leaves \(T\) -- that is composed of all the \emph{minimal} data points for this tree as %
    \[%
    X^\prime_{\textit{min}, T} = \{x^\prime_{\textit{min}, t} : t \in T\}%
    \text{.}%
    \]%
\end{Definition}

Next, we transform this minimal representation set \(X^\prime_{\textit{min}, T}\) from the interpretable into the original domain using the inverse of the IR transformation function: \(X_{\textit{min}, T} = \{\IR^{-1}(x^\prime_{\textit{min}, t}) : x^\prime_{\textit{min}, t} \in X^\prime_{\textit{min}, T}\}\). %
We then predict class probabilities for each instance in \(X_{\textit{min}, T}\) with the black box \(f\) and \textbf{replace} the values estimated by the surrogate tree with these probabilities for each leaf \(t \in T\), i.e., modify the surrogate tree by overriding its predictions. %
Doing so is only feasible for the tree leaves as the \emph{minimal data points} for some of the splitting nodes are indistinguishable; e.g., all the nodes on the root-to-leaf path that decides every interpretable feature to be \(1\) are non-unique and all would be represented by the unmodified explained instance. %
We additionally assume that the explained predictive model is \emph{deterministic}, therefore it always outputs the same prediction for a given instance. %

This variant of {\sc LIMEtree} -- called \underline{{\sc TREE}} and codified by Algorithm~\ref{algo:LIMEtree_star} provided in Appendix~\ref{apx:algos} -- %
guarantees \textbf{full fidelity} of the surrogate tree with respect to the \emph{explanations derived from the tree structure} such as counterfactuals and root-to-leaf decision rules %
(see Section~\ref{sec:discussion} for their examples). %
However, for this property to hold the function \(\IR\) transforming data from their original domain into the interpretable representation has to be \emph{deterministic}~\cite{sokol2020towards}, which holds for image and text but not for tabular data (refer back to Section~\ref{sec:background}). %
The rationale behind this claim is %
outlined in Lemma~\ref{lemma:local_tree_fidelity} (proof in Appendix~\ref{apx:proofs}), which follows from the %
subsequent discussion. %

\begin{Lemma}%
[Structural Fidelity]%
    \label{lemma:local_tree_fidelity}
    A surrogate tree can achieve \textbf{full fidelity} with respect to the explanations derived from its \emph{structure} -- i.e., model-driven explanations -- if the interpretable representation transformation function %
    \(\IR\) %
    is \textbf{deterministic}. %
    Therefore, an instance \(x \in \mathcal{X}\) can be translated into a unique point \(\IR(x) = x^\prime \in \mathcal{X}^\prime\) and vice versa \(\IR^{-1}(x^\prime) = x\), %
i.e., the mapping is one-to-one. %
\end{Lemma}

Lemma~\ref{lemma:local_tree_fidelity} guarantees that each leaf in the surrogate tree is associated with only one data point \(x_{\textit{min},t}\) in the original representation \(\mathcal{X}\). %
This instance is derived from the \emph{minimal} interpretable data point \(x^\prime_{\textit{min},t}\) by applying the inverse of the interpretable representation transformation function \(\IR^{-1}\), i.e., \(x_{\textit{min},t} = \IR^{-1}(x^\prime_{\textit{min},t})\). %
Therefore, \(x_{\textit{min},t}\) represents the explained instance with the smallest possible number of concepts deleted from it such that \(g_\mathit{id}(x^\prime_{\textit{min},t}) = t\). %
By assigning the probabilities predicted by the explained black box for each data point \(x_{\textit{min},t}\) to the corresponding leaf \(t\) of the surrogate, it achieves full fidelity for the minimal representation set \(X_{\textit{min},T}\), which is the backbone of model-driven explanations. %

While such an approach ensures full fidelity of model-driven explanations, the same is not guaranteed for data-driven explanations such as answers to what-if questions, e.g., ``What if concept \(x^{\prime}_i\) is absent?'' %
Root-to-leaf paths that do not condition on all of the binary interpretable features allow for more than one data point to be assigned to that leaf; e.g., for three binary features \([x^\prime_1, x^\prime_2, x^\prime_3] \in \{0, 1\}^3\), a root-to-leaf path with a \(x^\prime_1 < 0.5 \land x^\prime_3 < 0.5\) condition assigns \([0, 0, 0]\) and \([0, 1, 0]\) to this leaf. %
This observation motivates the minimal interpretable representation \(X^\prime_{\textit{min},t}\) (Definition~\ref{def:minimal_x_star}), which selects a single data point to represent each leaf thereby facilitating full fidelity of model-driven explanations without additional assumptions. %
However, for \emph{data-driven} explanations to achieve \emph{full fidelity}, the surrogate tree must faithfully model the \emph{entire} interpretable feature space, i.e., have one leaf for every data point in \(\mathcal{X}^{\prime}\), which can be thought of as \emph{extreme overfitting}. %
Since the cardinality of a binary \(d\)-dimensional space \(\mathbb{B}^d = \{0, 1\}^d\) is given by \(\vert\mathbb{B}^d\vert = 2^d\), and a complete and balanced binary decision tree of \(2^d\) width (number of leaves) is \(d\) deep, relaxing the tree complexity bound \(\Omega\) accordingly guarantees full fidelity of all the explanations -- a property captured by Corollary~\ref{cor:local_data_fidelity} (proof in Appendix~\ref{apx:proofs}).%

\begin{Corollary}%
[Full Fidelity]%
    \label{cor:local_data_fidelity}
    If the complexity bound (width) \(\Omega\) of a surrogate tree \(g\) is relaxed to equal the cardinality of the binary interpretable domain \(\mathcal{X}^\prime\), i.e., \(\Omega(g ; \; \vert\mathcal{X}^\prime\vert) = \frac{\text{width}(g)}{2^{\vert\mathcal{X}^\prime\vert}} = \frac{2^{\vert\mathcal{X}^\prime\vert}}{2^{\vert\mathcal{X}^\prime\vert}} = 1\), then the surrogate is guaranteed to achieve \textbf{full fidelity}. %
    This property applies to explanations that are both %
        \textbf{data-driven} -- %
        i.e., derived from any data point in the interpretable representation -- and %
    \textbf{model-driven} -- %
        i.e., derived from the structure of the surrogate tree. %
\end{Corollary}
 
Therefore, a surrogate tree that guarantees faithfulness of \emph{model-driven} explanations (Lemma~\ref{lemma:local_tree_fidelity}) can only deliver trustworthy counterfactuals and exemplar explanations sourced from the minimal representation set. %
This may be an attractive alternative to more complex surrogate trees that additionally guarantee faithfulness of \emph{data-driven} explanations (Corollary~\ref{cor:local_data_fidelity}). %
The latter surrogate type, which usually yields deeper trees, can deliver a broader spectrum of trustworthy explanations: tree structure-based explanations, feature importance, decision rules (root-to-leaf paths), answers to what-if questions and exemplar explanations based on \emph{any} data point, in addition to counterfactuals. %

\section{Qualitative, Quantitative and User-based Evaluation\label{sec:experiments}}

Next, we assess the explanatory power of {\sc LIMEtree} with a multi-tier evaluation approach that consists of an assessment guided by XAI desiderata (Section~\ref{sec:experiments:desiderata}) as well as \emph{functionally-grounded} (Section~\ref{sec:experiments:numerical}) and \emph{human-grounded} (Section~\ref{sec:experiments:user}) experiments~\cite{doshi2017towards,sokol2020explainability}. %
The first judges our approach against a number of criteria important for XAI systems; %
the second involves a (synthetic) proxy task in which we compare the (numerical) fidelity of LIME with multiple variants of {\sc LIMEtree} on image and tabular data; %
the third reports results of a pilot user study, %
which is based on image classification to %
enable straightforward qualitative evaluation of explanations by means of visual inspection, thus alleviating the need for technical expertise. %

\subsection{Desiderata\label{sec:experiments:desiderata}}%

XAI systems can generally be decomposed into two operationally distinct parts, one responsible for %
explanation \emph{generation} and another for its \emph{presentation}; this separation allows us %
to better identify, evaluate and report the unique desiderata important at each stage~\cite{sokol2024does}. %
Given that {\sc LIMEtree} is a surrogate explainer, the insights that it generates are post-hoc, therefore they may not reflect the true behaviour of the underlying black box~\cite{rudin2019stop}. %
This discrepancy -- empirically measured as \emph{fidelity} -- is an important indicator of explanation truthfulness, which should always be communicated to the explainees, especially in high stakes applications. %
While {\sc LIMEtree} can achieve full fidelity without sacrificing explanation comprehensibility, this desideratum is limited to IRs that are deterministic. %
To take advantage of this property it is therefore important to design an IR that addresses the explainability needs of a particular use case, which may require additional effort to build such a bespoke module despite the explainer itself being model-agnostic~\cite{sokol2020towards,mittelstadt2019explaining,sokol2022what}. %
More broadly, the truthfulness is a major advantage of our approach given that it allows to retrofit explainability into pre-existing black boxes. %
Whatever explanation type, presentation format and communication medium are chosen, this property guarantees that the explanatory insights are based on an accurate reflection of the black-box model's behaviour. %

Before reviewing desiderata of specific explanation types, we discuss a set of general properties that are expected of all explanatory insights~\cite{sokol2020explainability}. %
{\sc{}LIMEtree} excels when it comes to explanation plurality and diversity -- especially so given their consistency -- allowing the explainees to explore distinct aspects of the underlying black box without running into spuriously contradictory observations, %
further improving the trustworthiness of its explanations. %
While some of them are inherently static, others can be operationalised within an \emph{interactive explanatory protocol}~\cite{sokol2020one}, enabling the explainees to customise and personalise them in a natural way -- refer to Section~\ref{sec:discussion} for examples. %
This breadth of explanatory insights and access to their source -- the surrogate tree structure (see Figure~\ref{fig:mort}) -- enables their contextualisation, %
which makes them particularly appealing since good explanations do not only communicate \emph{what} information is used by a predictive model but also \emph{how} it is used~\cite{rudin2019stop}. %

By simultaneously accounting for multiple classes, {\sc LIMEtree} offers a more comprehensive picture of the explained model's predictive behaviour and facilitates user-driven exploration, which, as noted in Section~\ref{sec:intro}, can mitigate automation bias, especially so for counterfactuals~\cite{byrne2023good}. %
Also, recall that our method is compatible with \emph{hypothesis-driven} XAI since the breadth of its insights allows the explainees to consider multiple congruent explanations for different predictions of a given instance instead of only receiving a justification of the top prediction~\cite{miller2023explainable}. %
Given that our method operates as a surrogate, we can freely tweak and tune the target, breadth and scope of its explanations by adjusting its configuration, %
which further adds to its flexibility~\cite{sokol2020explainability,sokol2019blimey,sokol2020towards,sokol2022what,sokol2020one}. %

While {\sc LIMEtree} offers a broad spectrum of explanation types -- whose diversity makes it appealing to a wide range of audiences -- we anticipate the counterfactuals to be the most attractive given their ubiquity in XAI~\cite{miller2018explanation}. %
Notably, these insights are ante-hoc with respect to the surrogate tree, therefore their truthfulness is guaranteed in this regard~\cite{sokol2023reasonable}. %
Their generation procedure allows to account for plausibility and actionability of their conditional part as well as other (human-centred) properties that may be desired~\cite{sokol2020explainability,keane2021if,sokol2020one,sokol2018glass}. %
Counterfactual explanations are known to be intrinsically comprehensible given their parsimony and low complexity, making them an attractive choice across a diverse range of applications~\cite{sokol2020explainability,miller2018explanation}. %

\subsection{Synthetic Experiments\label{sec:experiments:numerical}}%

We evaluate the trustworthiness and comprehensibility of {\sc LIMEtree} explanations using the two components of the optimisation objective \(\mathcal{O}\) (Equation~\ref{eq:lime:objective}) -- \emph{fidelity} \(\mathcal{L}\) and \emph{complexity} \(\Omega\) -- as computational proxies. %
The former measures the faithfulness of the surrogate with respect to the black box, i.e., its ability to mimic the black box, which is the \emph{only} metric capable of reporting the reliability of all the diverse explanation types extracted from the surrogate. %
To this end, we employ the formulations of fidelity used by both LIME (Equation~\ref{eq:lime:loss}) and {\sc LIMEtree} (Equation~\ref{eq:blimey_loss}); %
we compute this property when modelling the top three classes predicted by the black box for each test instance. %
We additionally analyse the \emph{complexity} of {\sc LIMEtree} surrogates %
calculated as the tree depth normalised by the dimensionality of the IR (Equation~\ref{eq:blimey_complexity}); %
we then %
compare it to the corresponding measure for {\sc LIME} surrogates, which is computed by counting the number of non-zero coefficients of the underlying linear models, i.e., their size (Equation~\ref{eq:lime:complexity}). %

We study %
three variants of {\sc LIMEtree}, all of which \emph{minimise fidelity} but differ in complexity constraints and post-processing: %
\begin{description}
    \item[TREE] optimises a surrogate tree for complexity, i.e., it determines the shallowest tree that offers the desired level of fidelity;%
    \item[\underline{TREE}]
    is a variant of TREE
    whose %
    predictions are post-processed to guarantee full fidelity of model-driven explanations; and %
    \item[TREE\textsuperscript{\textdagger}] constructs a surrogate tree without any complexity constraints, allowing the algorithm to build a complete tree that guarantees full fidelity of both model- and data-driven explanations. %
\end{description}
These realisations of {\sc LIMEtree} are the most relevant %
given that each one offers a surrogate %
with distinct fidelity characteristics %
that %
lead to certain types of tree-based explanations achieving desired properties as explained earlier in Section~\ref{sec:fidelity}. %
We compare the fidelity of these explainers to {\sc LIME} with disabled feature selection, which allows it to achieve maximum fidelity at the expense of explanation size. %
Our study is limited to fidelity and complexity since XAI lacks metrics suitable for \emph{multi-class explainability} or for cases when \emph{multiple explanation types} are derived from a single source as well as for explanations that rely on \emph{probabilities} rather than crisp predictions (to mitigate automation bias)~\citep{sokol2025all}. %
{\sc LIME} is our only baseline given the general lack of multi-class explainers or methods whose underlying surrogate model can be directly accessed. %

\begin{table}[tb]
    \centering
        \setlength{\tabcolsep}{2.85pt} %
        \scriptsize
        
    \caption{%
Fidelity loss (mean \(\pm\) standard deviation, smaller is better, best results in bold) computed: %
    (n\textsuperscript{th} top) separately for each of the top three black-box predictions with the {\sc LIME} loss (Equation~\ref{eq:lime:loss}); and %
    (top n) collectively for the top one, two and three black-box predictions with the {\sc LIMEtree} loss (Equation~\ref{eq:blimey_loss}). %
        We report results for %
        \protect\subref*{tab:fidelity:image} three image and \protect\subref*{tab:fidelity:tabular} two tabular data sets with %
        four surrogates: {\sc LIME} and three variants of {\sc LIMEtree} ({\sc TREE}, \underline{\sc TREE} \& {\sc TREE}\textsuperscript{\textdagger}). %
The percentage shown after the explainer name specifies the tree complexity \(\Omega\) -- i.e., its depth divided by its maximum possible depth determined by the number of features in the interpretable representation -- at which loss is computed; %
{\sc TREE}\textsuperscript{\textdagger} is equivalent to {\sc TREE}@100\% (and {\sc \underline{TREE}}@100\% for deterministic IRs). %
See Figure~\ref{fig:loss} for examples of the loss behaviour. %
        }%
        \label{tab:fidelity}%
        
\begin{adjustwidth}{-\extralength}{0cm}
\subfloat[%
    Image data sets and the corresponding (pre-trained) neural networks~\cite{models}: ImageNet~\cite{imagenet} (1,659 samples, 256\(\times\)256 pixels, 1,000 classes) + Inception v3 (77\% accuracy); CIFAR-10~\cite{krizhevsky2009learning} (9,714 samples, 32\(\times\)32 pixels, 10 classes) + ResNet 56 (94\% accuracy); and CIFAR-100~\cite{krizhevsky2009learning} (9,665 samples, 32\(\times\)32 pixels, 100 classes) + RepVGG (77\% accuracy). %
    We use all validation set images for which an interpretable representation can be built; %
however, %
for ImageNet we first pre-select images that are square and at least 256\(\times\)256, which we resize to these dimensions. %
    The results are scaled up by \(10^{2}\). %
]{%
\makebox[\fulllength][c]{
    \begin{tabular}{@{}rl c@{} llll c llll c llll@{}}%
        \toprule%
        \multicolumn{2}{c}{%
        \multirow{2}{*}{%
        \(\times 10^{-2}\)%
        }
        }
        && \multicolumn{4}{c}{ImageNet + Inception v3}
        && \multicolumn{4}{c}{CIFAR-10 + ResNet 56}
        && \multicolumn{4}{c}{CIFAR-100 + RepVGG} \\
        \cmidrule{4-7} \cmidrule{9-12} \cmidrule{14-17}
        & & & LIME & TREE@66\% & \underline{TREE}@75\% & TREE\textsuperscript{\textdagger} %
        &&    LIME & TREE@66\% & \underline{TREE}@75\% & TREE\textsuperscript{\textdagger} %
        &&    LIME & TREE@66\% & \underline{TREE}@75\% & TREE\textsuperscript{\textdagger} \\
        \midrule%
        \multirow{3}{*}{\rotatebox[origin=c]{90}{\parbox[b]{.9cm}{\centering n\textsuperscript{th} top\vspace*{-2pt}}}} & \parbox{\widthof{2\textsuperscript{nd}}}{1\textsuperscript{st}} %
        && \(3.67\pm2.18\) & \(\pmb{0.60}\pm0.61\) & \(0.64\pm0.73\)       & \(\pmb{0}\pm0\) %
        && \(7.34\pm2.96\) & \(\pmb{2.17}\pm1.25\)       & \(2.77\pm1.66\) & \(\pmb{0}\pm0\)
        && \(3.33\pm1.80\) & \(\pmb{0.59}\pm0.56\)       & \(0.66\pm0.63\) & \(\pmb{0}\pm0\) \\
        & \parbox{\widthof{2\textsuperscript{nd}}}{2\textsuperscript{nd}} %
        && \(1.14\pm1.77\) & \(\pmb{0.24}\pm0.42\) & \(0.25\pm0.40\)       & \(\pmb{0}\pm0\) %
        && \(3.91\pm3.98\) & \(\pmb{1.28}\pm1.31\)       & \(1.69\pm1.76\) & \(\pmb{0}\pm0\)
        && \(0.97\pm1.46\) & \(\pmb{0.24}\pm0.36\)       & \(0.26\pm0.40\) & \(\pmb{0}\pm0\) \\
        & \parbox{\widthof{2\textsuperscript{nd}}}{3\textsuperscript{rd}} %
        && \(0.63\pm1.36\) & \(\pmb{0.13}\pm0.25\) & \(0.16\pm0.33\)       & \(\pmb{0}\pm0\) %
        && \(2.57\pm3.37\) & \(\pmb{0.89}\pm1.15\)       & \(1.10\pm1.44\) & \(\pmb{0}\pm0\)
        && \(0.56\pm1.13\) & \(\pmb{0.14}\pm0.29\)       & \(0.16\pm0.32\) & \(\pmb{0}\pm0\) \\
        \cmidrule{4-17}
        \multirow{3}{*}{\rotatebox[origin=c]{90}{\parbox[b]{.6cm}{\centering top n\vspace*{-2pt}}}} & 1 %
        && \(3.67\pm2.18\) & \(\pmb{0.60}\pm0.61\) & \(0.64\pm0.73\)       & \(\pmb{0}\pm0\) %
        && \(7.34\pm2.96\) & \(\pmb{2.17}\pm1.25\)       & \(2.77\pm1.66\) & \(\pmb{0}\pm0\) %
        && \(3.33\pm1.80\) & \(\pmb{0.59}\pm0.56\)       & \(0.66\pm0.63\) & \(\pmb{0}\pm0\) \\ %
        & 2 %
        && \(2.41\pm1.40\) & \(\pmb{0.42}\pm0.42\) & \(0.44\pm0.45\)       & \(\pmb{0}\pm0\) %
        && \(5.63\pm2.69\) & \(\pmb{1.73}\pm1.03\)       & \(2.23\pm1.42\) & \(\pmb{0}\pm0\)
        && \(2.15\pm1.15\) & \(\pmb{0.41}\pm0.36\)       & \(0.46\pm0.40\) & \(\pmb{0}\pm0\) \\
        & 3 %
        && \(2.72\pm1.58\) & \(\pmb{0.48}\pm0.47\) & \(0.53\pm0.50\)       & \(\pmb{0}\pm0\) %
        && \(6.91\pm3.26\) & \(\pmb{2.17}\pm1.28\)       & \(2.78\pm1.73\) & \(\pmb{0}\pm0\)
        && \(2.42\pm1.29\) & \(\pmb{0.48}\pm0.41\)       & \(0.54\pm0.45\) & \(\pmb{0}\pm0\) \\
        \bottomrule%
    \end{tabular}
}
    \label{tab:fidelity:image}%
}%
\par%
\subfloat[%
    Tabular data sets and the corresponding models (trained with scikit-learn~\cite{pedregosa2011scikit}): Wine~\cite{wine} (36 samples, 13 features, 3 classes) + Logistic Regression (93\% balanced accuracy); and Forest Covertypes~\cite{covertypes} (2,500 samples, 54 features, 7 classes) + Multilayer Perceptron (86\% balanced accuracy). %
    For Wine we use all the test set samples; %
    given their small number we repeated the study on the entire data set (178 samples) with comparable results. %
    The Forest Covertypes test set has 116,203 samples, from which we draw a stratified subset of size 2,500. %
    The results are scaled up by \(10^{1}\). %
]{%
\makebox[\fulllength][c]{
    \begin{tabular}{@{}rl c@{} llll c llll@{}}%
        \toprule%
        \multicolumn{2}{c}{%
        \multirow{2}{*}{%
        \(\times 10^{-1}\)%
        }
        }
        && \multicolumn{4}{c}{Wine + Logistic Regression}%
        && \multicolumn{4}{c}{Forest Covertypes + Multilayer Perceptron} \\
        \cmidrule{4-7}
        \cmidrule{9-12}
        & & & LIME & TREE@66\% & \underline{TREE}@100\% & TREE\textsuperscript{\textdagger} %
        &&    LIME & TREE@66\% & \underline{TREE}@100\% & TREE\textsuperscript{\textdagger} \\
        \midrule%
        \multirow{3}{*}{\rotatebox[origin=c]{90}{\parbox[b]{.9cm}{\centering n\textsuperscript{th} top\vspace*{-2pt}}}} & \parbox{\widthof{2\textsuperscript{nd}}}{1\textsuperscript{st}} %
        && \(0.29\pm0.27\) & \(\pmb{0.08}\pm0.11\) & \(5.54\pm3.43\)       & \(\pmb{0.07}\pm0.11\) %
        && \(0.59\pm0.26\) & \(\pmb{0.06}\pm0.06\)       & \(4.56\pm2.12\) & \(\pmb{0.06}\pm0.06\) \\
        & \parbox{\widthof{2\textsuperscript{nd}}}{2\textsuperscript{nd}} %
        && \(0.14\pm0.16\) & \(\pmb{0.03}\pm0.04\) & \(2.35\pm3.26\)       & \(\pmb{0.03}\pm0.04\) %
        && \(0.51\pm0.29\) & \(\pmb{0.05}\pm0.05\)       & \(1.88\pm1.21\) & \(\pmb{0.05}\pm0.05\) \\
        & \parbox{\widthof{2\textsuperscript{nd}}}{3\textsuperscript{rd}} %
        && \(0.20\pm0.28\) & \(\pmb{0.07}\pm0.12\) & \(3.73\pm4.18\)       & \(\pmb{0.06}\pm0.11\) %
        && \(0.13\pm0.21\) & \(\pmb{0.02}\pm0.04\)       & \(0.57\pm0.94\) & \(\pmb{0.02}\pm0.04\) \\
        \cmidrule{4-12}%
        \multirow{3}{*}{\rotatebox[origin=c]{90}{\parbox[b]{.6cm}{\centering top n\vspace*{-2pt}}}} & 1 %
        && \(0.29\pm0.27\) & \(\pmb{0.08}\pm0.11\) & \(5.54\pm3.43\)       & \(\pmb{0.07}\pm0.11\) %
        && \(0.59\pm0.26\) & \(\pmb{0.06}\pm0.06\)       & \(4.56\pm2.12\) & \(\pmb{0.06}\pm0.06\) \\ %
        & 2 %
        && \(0.22\pm0.19\) & \(\pmb{0.06}\pm0.07\) & \(3.94\pm2.67\)       & \(\pmb{0.05}\pm0.07\) %
        && \(0.55\pm0.26\) & \(\pmb{0.06}\pm0.05\)       & \(3.22\pm1.04\) & \(\pmb{0.06}\pm0.05\) \\
        & 3 %
        && \(0.32\pm0.29\) & \(\pmb{0.09}\pm0.12\) & \(5.80\pm3.56\)       & \(\pmb{0.08}\pm0.12\) %
        && \(0.62\pm0.29\) & \(\pmb{0.07}\pm0.06\)       & \(3.51\pm1.09\) & \(\pmb{0.07}\pm0.06\) \\
        \bottomrule%
    \end{tabular}
}
    \label{tab:fidelity:tabular}%
}%
\end{adjustwidth}
\end{table}

Table~\ref{tab:fidelity}, which reports the results of our evaluation, also summarises our experimental setup. %
We use a collection of popular multi-class image %
and tabular %
data sets; %
with the former we rely on a selection of pre-trained neural networks, and with the latter we split the data into stratified 80\% training and 20\% test sets, %
and fit the models ourselves. %
{\sc LIME} and {\sc LIMEtree} are implemented following best practice described in the literature~\cite{sokol2019blimey,sokol2020towards,garreau2020explaining,sokol2020fatf,sokol2022fatf,sokol2022what}. %
For images we use an IR built upon SLIC (edge-based) segmentation~\cite{achanta2012slic} with black colour occlusion as the information removal proxy; %
given its deterministic transformation function we operate directly on the binary interpretable domain and %
generate its full set of instances instead of their random sample to enable the surrogate to reach full fidelity. %
For tabular data we sample 10,000 instances around the explained data point in the original domain -- using \emph{mixup}, which is an explicitly local sampler that accounts for class labels~\cite{zhang2017mixup,sokol2022fatf} -- since the corresponding IR transformation function is non-deterministic; %
we use quartile-based discretisation applied to the data sample followed by binarisation as our interpretable domain. %
For images we use cosine distance measured in the IR, and for tabular data we use Euclidean distance measured in the original domain; %
we use the exponential kernel for both, %
with its optimal parameter determined experimentally for each data set. %
Our code is available on GitHub\coderepo. %

In our experiments %
{\sc LIME} produces three independent linear surrogates, one per class; %
each {\sc LIMEtree} variant is either built as a \emph{single} surrogate that models all of the classes simultaneously (n\textsuperscript{th} top), or a \emph{separate} surrogate is constructed for a one-, two- and three-class problem (top n). %
In deployment, however, {\sc LIMEtree} fits only a single multi-output tree, whereas LIME requires as many models as explained classes. %
As a result, since both methods follow the same steps except for the surrogate model training phase, %
our method tends to be faster for relatively small trees given that they are fitted to binary data with feature thresholds fixed at \(\sfrac{1}{2}\) -- up to the depth of 20 in our experiments -- and becomes negligibly slower for large trees -- requiring 250 milliseconds more than LIME for trees as deep as 40 -- but these measures will fluctuate with the number of explained classes and the IR dimensionality. %
Since the number of interpretable features should be kept low to improve human comprehensibility of the explanations, which directly limits the surrogate tree depth, we expect {\sc LIMEtree} to be faster in practice~\cite{sokol2020towards}. %

To assess explanation quality we measure multi-class fidelity with the {\sc LIMEtree} loss as well as the fidelity of each class separately with the {\sc LIME} loss. %
The experimental results, summarised in Table~\ref{tab:fidelity}, %
show that our base method -- {\sc TREE} -- provides more faithful explanations than {\sc LIME} at \(\sfrac{2}{3}\) of its complexity for tabular and image data. %
\underline{{\sc TREE}} -- which post-processes the surrogate tree to facilitate \emph{full fidelity of model-driven explanations} when the IR transformation function is deterministic -- also surpasses {\sc LIME} at \(\sfrac{3}{4}\) of its complexity for image data given their compliant IR, but its performance is degraded for tabular data even at full tree complexity (100\%) due to the stochasticity of the underlying IR. %
\underline{{\sc TREE}} requires higher complexity, i.e., deeper trees, than {\sc TREE} to achieve comparable fidelity since the post-processing step makes the surrogate faithful with respect to the \emph{minimal interpretable data points} but at the same time sub-optimal for the remainder of the interpretable space, %
which is especially detrimental for stochastic IRs where each minimal interpretable data point corresponds to multiple instances in the original data domain. %

\begin{figure}[t]%
    \centering
\begin{adjustwidth}{-\extralength}{0cm}
    \subfloat[CIFAR-100 (image data set).]{%
\makebox[.49\fulllength][c]{
        \includegraphics[width=.49\fulllength]{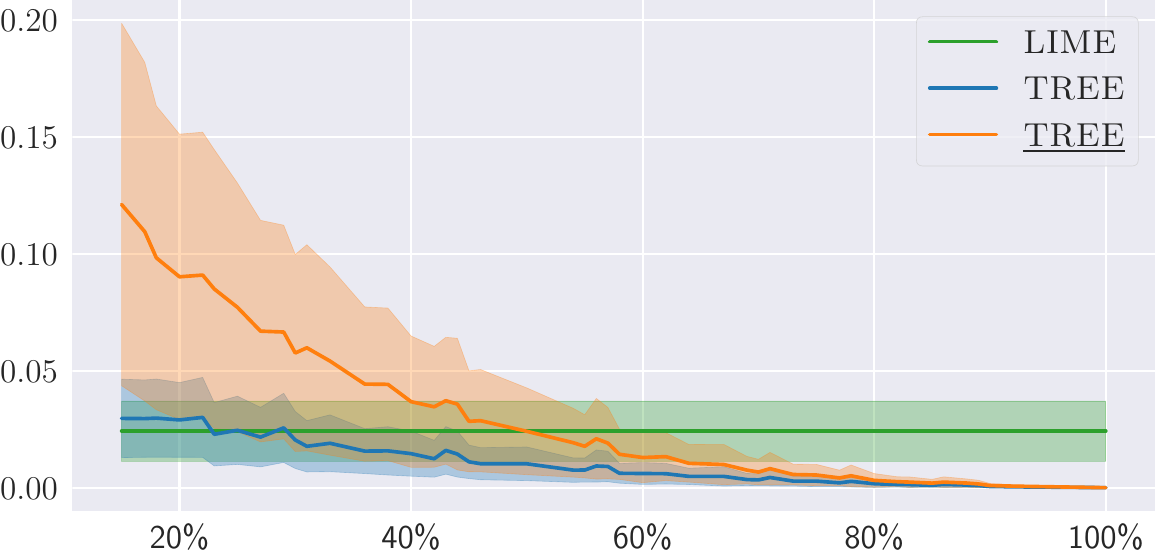}%
}%
\label{fig:loss:cifar100}%
    }%
    \hspace{.015\fulllength}
    \subfloat[Forest Covertypes (tabular data set).]{%
\makebox[.49\fulllength][c]{
        \includegraphics[width=.49\fulllength]{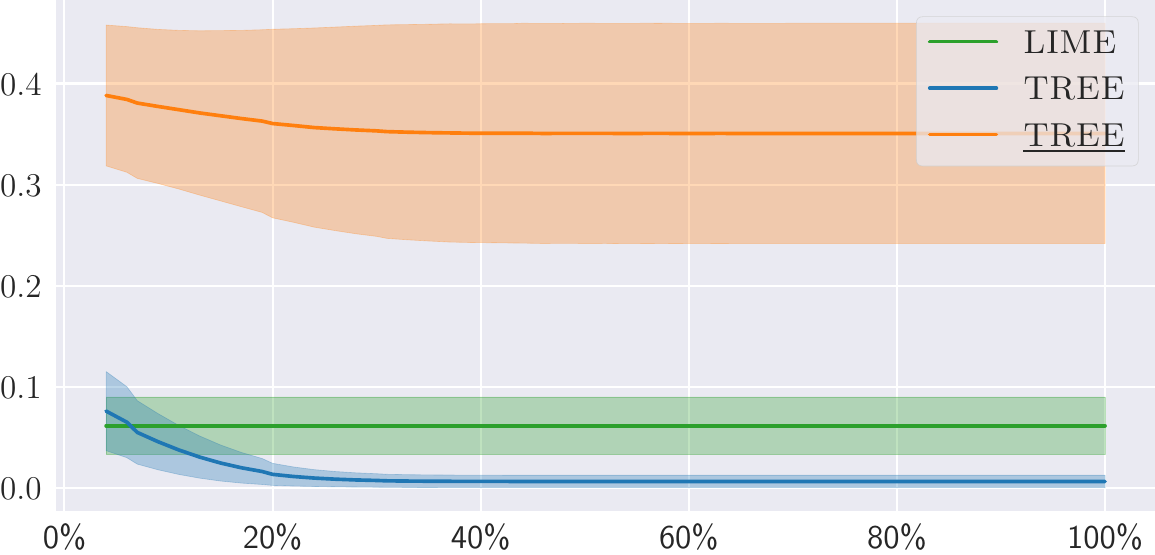}%
}\label{fig:loss:forest}}%
\end{adjustwidth}
    \caption{%
Behaviour of the {\sc LIMEtree} loss (fidelity \(\mathcal{L}\) and its standard deviation, y-axis) %
computed for the top three classes of the \protect\subref*{fig:loss:cifar100}~CIFAR-100 and \protect\subref*{fig:loss:forest}~Forest Covertypes data sets and %
plotted against surrogate complexity (\(\Omega\), x-axis) %
given as %
the ratio between the depth of the tree and its maximum depth (complete tree) determined by the number of features of the interpretable domain. %
    We report results for %
    three surrogate variants: %
    {\sc LIME}, %
    {\sc TREE} and %
    \underline{{\sc TREE}}; %
    the plots are %
    representative of the other data sets used in our experiments (see Appendix~\ref{apx:loss} for the remaining figures) and complement the fidelity at fixed tree complexity levels (66\%, 75\% and 100\%) reported in Table~\ref{tab:fidelity}. %
    {\sc LIME} complexity %
    is constant and given by the number of features in the interpretable representation, i.e., 100\% equivalent. %
    \label{fig:loss}}
\end{figure}

The version of {\sc LIMEtree} without a depth bound -- {\sc TREE}\textsuperscript{\textdagger}, which is equivalent to {\sc TREE}@100\% (and {\sc \underline{TREE}}@100\% for deterministic interpretable representations) -- achieves full fidelity across the board for a deterministic IR (images), where it faithfully models the entire interpretable data space by constructing one leaf per instance, but fails to do so for a non-deterministic IR (tabular) because in this case each tree leaf has to model multiple distinct data points. %
By allowing deeper trees we reduce the impurity of their leaves, which improves the overall performance of the surrogates -- an intuitive relation, and trade-off, between the complexity of the trees and their fidelity, two representative examples of which are shown in Figure~\ref{fig:loss}. %
Appendix~\ref{apx:loss} provides the complete collection of plots depicting the behaviour of the {\sc LIME} and {\sc LIMEtree} loss for all the data sets used in our experiments. %

\subsection{Pilot User Study\label{sec:experiments:user}}%

To assess real-life usefulness of our approach we ran a pilot user study. %
We recruited eight participants (six males and two females) evenly distributed across the 18--45 age range with diverse skills and backgrounds; %
six of them had a machine learning background and three were familiar with AI explainability. %
The participants were not compensated for their involvement in the user study. %
We exposed them to {\sc LIME} (Figure~\ref{fig:lime}) and {\sc LIMEtree} (Figure~\ref{fig:mort}) explanations in a random order without revealing the method's name. %
The study consisted of two sections, one per explainer, displaying an image split into three segments, with each part enclosing a unique object, e.g., a cat, a dog and a ball. %
The two most pertinent black-box predictions for each object were then explained with both methods -- %
e.g., \emph{tabby} and \emph{tiger cat} for the cat object, \emph{golden retriever} and \emph{Labrador retriever} for the dog object, and \emph{tennis ball} and \emph{croquet ball} for the ball object -- %
yielding six LIME explanations and a single multi-output tree spanning all six predictions. %
The participants were offered a brief tutorial illustrating how to parse the tree structure to obtain a variety of explanations. %

\begin{figure}[t]%
    \centering
\begin{adjustwidth}{-\extralength}{0cm}
        \includegraphics[width=\fulllength]{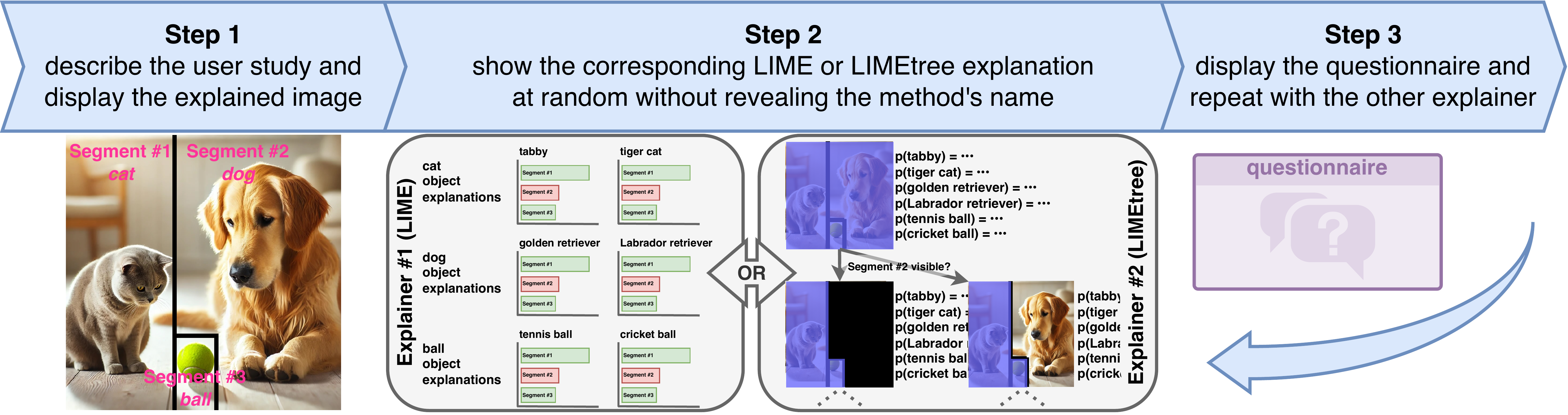}%
\end{adjustwidth}
        \caption{High-level overview of the user study flow.\label{fig:usflow}}%
\end{figure}

The participants were then asked about the expected behaviour of the black box in relation to any two out of the three displayed objects for each explainer -- six questions in total as the relations are assumed to be non-reflexive. %
For example, ``How does the presence of \emph{the cat object} affect the model's confidence of a presence of \emph{the dog object}?'', with three possible answers: \emph{confidence decreases}, \emph{confidence not affected} and \emph{confidence increases}. %
This question formulation was chosen to avoid a bias towards either explainer since we could neither ask for importance or influence of each object on a particular prediction ({\sc LIME}'s domain), nor the relation between an object and a prediction, e.g., a what-if question ({\sc LIMEtree}'s domain). %
Before viewing the explanations, the participants were asked to answer a similar set of questions using only their intuition, which allowed us to assess whether the explainees still relied on their intuition when explicitly asked to work with the explanations. %
Figure~\ref{fig:usflow} provides the %
high-level overview of the flow of our user study. %

Our findings indicate a negligible overlap between the responses based on the participants' intuition and both explainers; %
they also show that {\sc LIMEtree} helped the participants to answer 25\% more of the questions correctly as compared to {\sc LIME}. %
All of the participants indicated that using LIME was either \emph{easy} or \emph{very easy}, and at the same time rated the process of manually extracting {\sc LIMEtree} explanations as either \emph{difficult} or \emph{very difficult}, despite many of the explainees having AI background. %
This disparity in conjunction with subpar performance when using LIME suggests that the explainees misinterpreted its explanations and were overconfident~\cite{small2023helpful,xuan2023users}; %
good performance when working with {\sc LIMEtree} despite the difficulty in using its explanations, on the other hand, is promising given that the process of extracting them can be easily automated. %

\section{Discussion\label{sec:discussion}}%

{\sc LIMEtree} explanations are versatile and appealing but achieving their full fidelity presupposes a \emph{deterministic} IR transformation function (Lemma~\ref{lemma:local_tree_fidelity}) and a \emph{complete} surrogate tree (Corollary~\ref{cor:local_data_fidelity}). %
This is not a problem for image and text data since the corresponding IRs can be built to be deterministic and of low dimensionality (given by the number of desired human-comprehensible concepts). %
The IR of tabular data, however, %
is inherently non-deterministic~\cite{sokol2020towards} -- due to the many-to-one mapping introduced by discretisation and binarisation (refer back to Section~\ref{sec:background}) -- %
with its dimensionality equal to the size of the original feature space. %
Nonetheless, since uniquely for tabular data the surrogate tree can be trained directly on their original representation, thus implicitly constructing a locally faithful and meaningful IR instead of relying on an external one~\cite{sokol2019blimey,sokol2020towards}, the surrogate can be overfitted to maximise its fidelity. %
While {\sc LIMEtree} offers a close approximation in both cases, full fidelity cannot be guaranteed since even a \emph{complete} surrogate tree is unable to achieve full coverage for non-deterministic IRs. %
The consequences of this shortcoming %
can be seen in our experimental results (refer to Table~\ref{tab:fidelity}), which show that a complete surrogate tree -- labelled {\sc{}TREE}\textsuperscript{\textdagger} -- can reach full fidelity for image but not for tabular data. %

In practice, full fidelity of surrogates based on deterministic IRs (Lemma~\ref{lemma:local_tree_fidelity}) is achieved by adjusting the \emph{sample size} \(|X^{\prime}|\) and relaxing the tree \emph{complexity bound} \(\Omega\). %
Recall that %
a \(d\)-dimensional binary interpretable representation \(\mathcal{X}^\prime \equiv \mathbb{B}^d = \{0, 1\}^d\) has \(\vert\mathcal{X}^\prime\vert = 2^d\) unique instances, %
and the width, i.e., the number of leaves, of a complete, balanced binary decision tree of depth \(d\) is \(2^d\) (Corollary~\ref{cor:local_data_fidelity}). %
Therefore, we can use all of these data points -- there is no benefit from oversampling -- to easily train a local surrogate with its complexity bound \(\Omega\) removed to allow complete trees of depth \(d\), i.e., with one leaf per instance, guaranteeing \emph{full fidelity} and access to a diverse range of faithful and comprehensible explanations. %
The depth bound and the sample size can be adjust dynamically prior to training the surrogate to ensure its optimality since the size of the interpretable domain is known beforehand. %

Since for images as well as text each dimension of the IR captures a human-comprehensible concept, their number is expected to be low, %
especially that %
tokens in text excerpts and segments in images do not have to be adjacent to constitute a single concept. %
For every additional feature in the interpretable space, the number of sampled data points doubles and the tree depth is incremented by one in order to provide the interpretable domain and the surrogate tree with enough capacity to preserve the full fidelity guarantee. %
While this exponential growth in the number of interpretable data points may seem overwhelming, training decision trees on binary data spaces is fast given the predetermined \(\sfrac{1}{2}\) split at every node. %
The exponential growth of the width of the surrogate tree that guarantees its full fidelity increases its complexity and can have adverse effects on the comprehensibility of some explanation types, however, as we show next, it does not affect the most important and versatile explanation kinds. %

\begin{figure}[t]
    \centering
\begin{adjustwidth}{-\extralength}{0cm}
    \subfloat[Tree-based feature importance explanation, which is shared between all the three explained classes.]{%
\makebox[0.2285\fulllength][c]{
        \includegraphics[height=0.2244\fulllength]{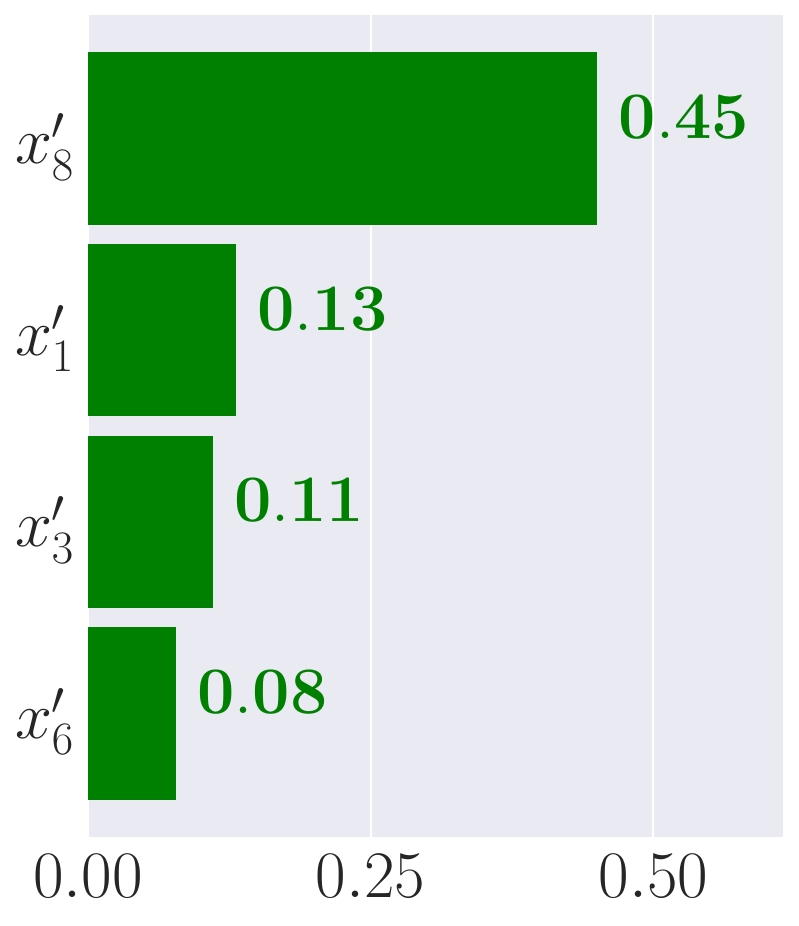}%
\label{fig:examples:fi}%
}}%
    \hspace{0.03025\textwidth}%
    \subfloat[The shortest -- i.e., highest number of occlusions with 6 out of 8 segments removed -- exemplar explanation of \emph{tennis ball} (97\%).]{%
\makebox[0.2285\fulllength][c]{
        \includegraphics[height=0.2244\fulllength]{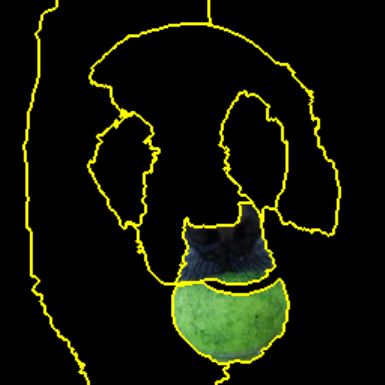}%
\label{fig:examples:ex}%
}}%
    \hspace{0.03025\textwidth}%
    \subfloat[What-if / counterfactual explanation: ``If segment \#8 (representing the ball) is removed, the black box predicts \emph{golden retriever} (97\%).'']{%
\makebox[0.2285\fulllength][c]{
        \includegraphics[height=0.2244\fulllength]{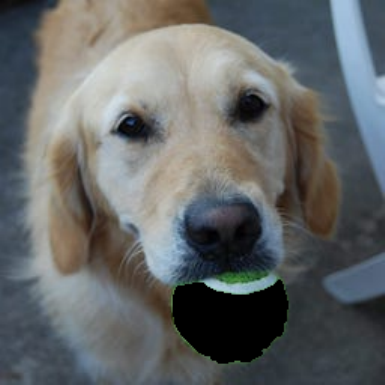}%
\label{fig:examples:cf}%
}}%
    \hspace{0.03025\textwidth}%
    \subfloat[Visualisation of an explanation based on a root-to-leaf decision rule that maximises the probability of \emph{Labrador retriever} (98\%).]{%
\makebox[0.2285\fulllength][c]{
        \includegraphics[height=0.2244\fulllength]{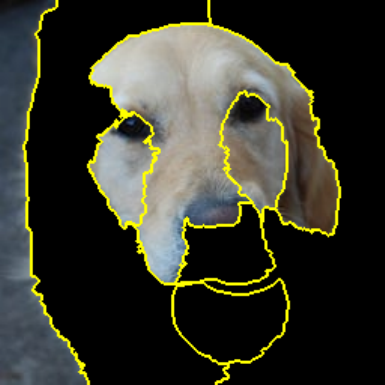}%
\label{fig:examples:rtl}%
}}%
\end{adjustwidth}
    \caption{%
Examples of four {\sc LIMEtree} explanation types complementing the tree structure visualisation shown in Figure~\ref{fig:mort}: %
\protect\subref*{fig:examples:fi}~feature importance, %
\protect\subref*{fig:examples:ex}~exemplar, %
\protect\subref*{fig:examples:cf}~what-if /counterfactual and %
\protect\subref*{fig:examples:rtl}~decision rule. %
These insights allow to uncover the heuristic used by the black box to differentiate between the three explained classes, which is not feasible with the LIME explanations displayed in Figure~\ref{fig:lime}. %
Panels~\protect\subref*{fig:examples:ex}, \protect\subref*{fig:examples:cf} \& \protect\subref*{fig:examples:rtl} show explanations generated to maximise the predicted probability of one of the classes; %
they are presented here with appealing visualisations, but they can also be communicated via the underlying logical expressions, e.g., \(x^\prime_1 = 0 \land x^\prime_2 = 0 \land x^\prime_3 = 1 \land x^\prime_4 = 1 \land x^\prime_5 = 1 \land x^\prime_6 = 1 \land x^\prime_7 = 0 \land x^\prime_8 = 0\) for Panel~\protect\subref*{fig:examples:rtl}. %
Note that {\sc LIMEtree} explanations can be customised to an individual explainee's needs, which can be seen in Panels~\protect\subref*{fig:examples:ex}, \protect\subref*{fig:examples:cf} \& \protect\subref*{fig:examples:rtl}; %
the user can ask for certain image segments (i.e., interpretable features) to be \emph{preserved} and other \emph{discarded} as well as for the \emph{smallest} or \emph{biggest} possible occlusion, at the same time requesting to maximise the probability of a selected class (according to the black box). %
\label{fig:examples}}
\end{figure}

Guaranteeing full fidelity of a surrogate tree requires relaxing its complexity bound \(\Omega\), which the optimisation objective \(\mathcal{O}\) tries to minimise (Equation~\ref{eq:lime:objective}). %
Since in this setting a moderate number of interpretable features may yield a relatively large tree, the increased complexity of the resulting explanations is concerning. %
While a complex surrogate tree may render the explanations based on its structure, e.g., model visualisations, incomprehensible, these are not the most appealing explanation types and their appreciation often requires AI expertise. %
The (interpretable) feature importance, what-if explanations, \emph{counterfactuals} and exemplars are not affected by the tree size in any way and remain highly compact and comprehensible %
-- see Figure~\ref{fig:examples} for some examples %
and Appendix~\ref{apx:examples} for a more comprehensive overview of diverse explanation types.
Notably, a complete surrogate tree with full fidelity will produce more counterfactual explanations for every data point, making it more interpretable. %

The decision rules -- logical conditions extracted from root-to-leaf paths -- may indeed become overwhelmingly long, in fact as long as the tree depth, however this does not impact all the data types equally and an appropriate presentation medium can alleviate this issue regardless of the tree complexity. %
For image and text data such rules will always be comprehensible, no matter their length, since they %
cannot have more literals than the dimensionality of the underlying interpretable domain, i.e., the number of segments for images and word-based tokens for text. %
Presenting this rule in the former case corresponds to displaying an image with its various segments occluded -- e.g., see %
Figure~\ref{fig:examples:rtl} %
-- and in the latter producing a text excerpt with selected tokens removed. %
For tabular data, however, these rules may become relatively long and incomprehensible since this domain lacks a similar human-friendly representation; the exception is root-to-leaf paths that impose multiple logical conditions on a single feature (in the original domain), allowing for their compression. %
Regardless of the presentation medium, a general criticism of rule-based explanations is the difficulty of understanding how each logical condition affects the prediction, making them less appealing than other explanation types.%

In view of these observations, if explanations based on the structure of the surrogate tree are not required for image and text data, and additionally rule-based explanations are not needed for tabular data, the model complexity \(\Omega\) does not have to be minimised. %
It can therefore be removed from the optimisation objective \(\mathcal{O}\) given in Equation~\ref{eq:lime:objective}, %
paving the way for full explanation fidelity. %
{\sc LIMEtree} therefore delivers %
a practical surrogate explainer with strong guarantees and well-understood limitations. %
Since it can be used with \emph{any} AI model and data type -- albeit with some constraints for tabular data -- it promises to become an invaluable tool for inspecting, debugging and explaining black-box predictive systems. %

\section{Conclusion and Future Work}

In this paper we introduced the concept of \emph{multi-class explainability} and proposed a \emph{surrogate explainer} -- called {\sc LIMEtree} -- based on \emph{multi-output regression trees} that is compatible with this paradigm. %
We then analysed its various properties and guarantees, and showed how it can achieve full fidelity. %
Next, we demonstrated how {\sc LIMEtree} improves upon LIME and discussed the benefits of using trees as surrogate models. %
We supported these claims with an assessment of its properties based on XAI desiderata as well as a collection of quantitative experiments and a pilot user study. %
At a higher level, the \emph{multi-class explainability} paradigm delivers more comprehensive insights into the functioning of opaque predictive models than are otherwise available with current XAI conceptualisations. %
Additionally, our implementation of this paradigm in the form of a \emph{surrogate explainer} %
enables its straightforward adoption across many application domains %
given our tool's compatibility with diverse data types %
as well as its clear guarantees and limitations. %
Together, our contributions advance both the conceptual and practical frontiers of explainable artificial intelligence. %

In future work we will %
implement methods to algorithmically extract human-centred explanations from (surrogate) trees %
-- looking into the interactive aspects of this process -- %
and evaluate them with large-scale user studies. %
We will also investigate alternative interpretable representations of tabular data that would allow {\sc LIMEtree} to achieve better fidelity guarantees for this data domain (e.g., by providing a deterministic IR transformation function). %
Finally, we %
plan to explore other techniques capable of realising the multi-class explainability paradigm as well as look into expanding this concept to better account for %
prediction uncertainty output by probabilistic classifiers -- a perspective that is largely neglected by the XAI literature. %

\vspace{6pt}

\authorcontributions{%
Conceptualisation, K.S.;
methodology, K.S.;
software, K.S.;
validation, K.S.;
formal analysis, K.S.;
investigation, K.S.;
resources, P.F.;
writing -- original draft preparation, K.S.;
writing -- review and editing, K.S. and P.F.;
visualisation, K.S.;
supervision, P.F.;
funding acquisition, P.F.
}

\funding{%
This research was supported by %
the TAILOR project, funded by EU Horizon 2020 research and innovation programme (grant agreement number 952215).%
}

\institutionalreview{%
Not applicable.%
}

\informedconsent{%
Not applicable.%
}

\dataavailability{%
The following data sets were used for this research: %
ImageNet (\url{https://image-net.org/});
CIFAR-10 \& CIFAR-100 (\url{https://www.cs.toronto.edu/~kriz/cifar.html});
Wine (\url{https://archive.ics.uci.edu/dataset/109/wine}); and
Forest Covertypes (\url{https://archive.ics.uci.edu/dataset/31/covertype}).
The source code needed to reproduce our experiments is available on GitHub (\coder).
}

\acknowledgments{%
We would like to acknowledge contributions of Alexander Hepburn and Raul Santos-Rodriguez, who helped with the development of the code used for the experiments and offered insightful feedback.%
}

\conflictsofinterest{%
We declare no conflicts of interest.%
}

\abbreviations{Abbreviations}{
The following abbreviations are used in this manuscript:\\

\noindent 
\begin{tabular}{@{}ll}
AI & Artificial Intelligence\\
CPU & Central Processing Unit \\
GAM & Generalised Additive Model\\
GPU & Graphics Processing Unit\\
IR & Interpretable Representation\\
LIME & Local Interpretable Model-agnostic Explanations\\
XAI & eXplainable Artificial Intelligence
\end{tabular}
}

\begin{adjustwidth}{-\extralength}{0cm}

\reftitle{References}

\bibliography{template}

\begin{thebibliography}{999}

\bibitem[Rudin(2019)]{rudin2019stop}
Rudin, C.
\newblock Stop explaining black box machine learning models for high stakes decisions and use interpretable models instead.
\newblock {\em Nature Machine Intelligence} {\bf 2019}, {\em 1},~206--215.

\bibitem[Sokol and Flach(2021)]{sokol2021explainability}
Sokol, K.; Flach, P.
\newblock Explainability is in the mind of the beholder: {Establishing} the foundations of explainable artificial intelligence.
\newblock {\em arXiv Preprint} {\bf 2021},  \href{http://arxiv.org/abs/2112.14466}{{\normalfont [2112.14466]}}.

\bibitem[Longo et~al.(2024)Longo, Brcic, Cabitza, Choi, Confalonieri, Del~Ser, Guidotti, Hayashi, Herrera, Holzinger, Jiang, Khosravi, Lecue, Malgieri, P{\'a}ez, Samek, Schneider, Speith, and Stumpf]{longo2024explainable}
Longo, L.; Brcic, M.; Cabitza, F.; Choi, J.; Confalonieri, R.; Del~Ser, J.; Guidotti, R.; Hayashi, Y.; Herrera, F.; Holzinger, A.;  et~al.
\newblock Explainable artificial intelligence ({XAI}) 2.0: {A} manifesto of open challenges and interdisciplinary research directions.
\newblock {\em Information Fusion} {\bf 2024}, {\em 106},~102301.

\bibitem[Guidotti et~al.(2018)Guidotti, Monreale, Ruggieri, Turini, Giannotti, and Pedreschi]{guidotti2018survey}
Guidotti, R.; Monreale, A.; Ruggieri, S.; Turini, F.; Giannotti, F.; Pedreschi, D.
\newblock A survey of methods for explaining black box models.
\newblock {\em ACM Computing Surveys (CSUR)} {\bf 2018}, {\em 51},~1--42.

\bibitem[Miller(2019)]{miller2018explanation}
Miller, T.
\newblock Explanation in artificial intelligence: {Insights} from the social sciences.
\newblock {\em Artificial Intelligence} {\bf 2019}, {\em 267},~1--38.

\bibitem[Wachter et~al.(2017)Wachter, Mittelstadt, and Russell]{wachter2017counterfactual}
Wachter, S.; Mittelstadt, B.; Russell, C.
\newblock Counterfactual explanations without opening the black box: {Automated} decisions and the {GPDR}.
\newblock {\em Harvard Journal of Law \& Technology} {\bf 2017}, {\em 31},~841.

\bibitem[Poyiadzi et~al.(2020)Poyiadzi, Sokol, Santos-Rodriguez, De~Bie, and Flach]{poyiadzi2020face}
Poyiadzi, R.; Sokol, K.; Santos-Rodriguez, R.; De~Bie, T.; Flach, P.
\newblock {FACE}: {Feasible} and actionable counterfactual explanations.
\newblock In Proceedings of the AAAI/ACM Conference on AI, Ethics, and Society,  2020, pp. 344--350.

\bibitem[Romashov et~al.(2022)Romashov, Gjoreski, Sokol, Martinez, and Langheinrich]{romashov2022baycon}
Romashov, P.; Gjoreski, M.; Sokol, K.; Martinez, M.V.; Langheinrich, M.
\newblock {BayCon}: {Model-agnostic} {Bayesian} counterfactual generator.
\newblock In Proceedings of the IJCAI,  2022, pp. 740--746.

\bibitem[Waa et~al.(2018)Waa, Robeer, Diggelen, Brinkhuis, and Neerincx]{waa2018contrastive}
Waa, J.v.d.; Robeer, M.; Diggelen, J.v.; Brinkhuis, M.; Neerincx, M.
\newblock Contrastive explanations with local foil trees.
\newblock In Proceedings of the 2018 ICML Workshop on Human Interpretability in Machine Learning (WHI 2018),  2018.

\bibitem[Verma et~al.(2024)Verma, Boonsanong, Hoang, Hines, Dickerson, and Shah]{verma2024counterfactual}
Verma, S.; Boonsanong, V.; Hoang, M.; Hines, K.; Dickerson, J.; Shah, C.
\newblock Counterfactual explanations and algorithmic recourses for machine learning: {A} review.
\newblock {\em ACM Computing Surveys (CSUR)} {\bf 2024}, {\em 56}.

\bibitem[Byrne(2023)]{byrne2023good}
Byrne, R.M.
\newblock Good explanations in explainable artificial intelligence ({XAI}): {Evidence} from human explanatory reasoning.
\newblock In Proceedings of the IJCAI,  2023, pp. 6536--6544.

\bibitem[Miller(2023)]{miller2023explainable}
Miller, T.
\newblock Explainable {AI} is dead, long live explainable {AI}! {Hypothesis-driven} decision support using evaluative {AI}.
\newblock In Proceedings of the 2023 ACM Conference on Fairness, Accountability, and Transparency,  2023, pp. 333--342.

\bibitem[Weld and Bansal(2019)]{weld2018challenge}
Weld, D.S.; Bansal, G.
\newblock The challenge of crafting intelligible intelligence.
\newblock {\em Communications of the ACM} {\bf 2019}, {\em 62},~70--79.

\bibitem[Craven and Shavlik(1995)]{craven1996extracting}
Craven, M.; Shavlik, J.W.
\newblock Extracting tree-structured representations of trained networks.
\newblock In Proceedings of the Advances in Neural Information Processing Systems,  1995, pp. 24--30.

\bibitem[Breiman et~al.(1984)Breiman, Friedman, Stone, and Olshen]{breiman1984classification}
Breiman, L.; Friedman, J.; Stone, C.J.; Olshen, R.A.
\newblock {\em Classification and regression trees}; CRC Press,  1984.

\bibitem[Sokol(2021)]{sokol2021towards}
Sokol, K.
\newblock Towards intelligible and robust surrogate explainers: {A} decision tree perspective.
\newblock PhD thesis, University of Bristol,  2021.

\bibitem[Saeed and Omlin(2023)]{saeed2023explainable}
Saeed, W.; Omlin, C.
\newblock Explainable {AI} ({XAI}): {A} systematic meta-survey of current challenges and future opportunities.
\newblock {\em Knowledge-Based Systems} {\bf 2023}, {\em 263},~110273.

\bibitem[Retzlaff et~al.(2024)Retzlaff, Angerschmid, Saranti, Schneeberger, Roettger, Mueller, and Holzinger]{retzlaff2024post}
Retzlaff, C.O.; Angerschmid, A.; Saranti, A.; Schneeberger, D.; Roettger, R.; Mueller, H.; Holzinger, A.
\newblock Post-hoc vs ante-hoc explanations: {xAI} design guidelines for data scientists.
\newblock {\em Cognitive Systems Research} {\bf 2024}, {\em 86},~101243.

\bibitem[Karimi et~al.(2021)Karimi, Sch{\"o}lkopf, and Valera]{karimi2021algorithmic}
Karimi, A.; Sch{\"o}lkopf, B.; Valera, I.
\newblock Algorithmic recourse: {From} counterfactual explanations to interventions.
\newblock In Proceedings of the 2021 ACM Conference on Fairness, Accountability, and Transparency,  2021, pp. 353--362.

\bibitem[Sokol and Flach(2020)]{sokol2020explainability}
Sokol, K.; Flach, P.
\newblock Explainability fact sheets: {A} framework for systematic assessment of explainable approaches.
\newblock In Proceedings of the 2020 ACM Conference on Fairness, Accountability, and Transparency,  2020, pp. 56--67.

\bibitem[Guidotti(2024)]{guidotti2024counterfactual}
Guidotti, R.
\newblock Counterfactual explanations and how to find them: {Literature} review and benchmarking.
\newblock {\em Data Mining and Knowledge Discovery} {\bf 2024}, {\em 38},~2770--2824.

\bibitem[Meske et~al.(2022)Meske, Bunde, Schneider, and Gersch]{meske2022explainable}
Meske, C.; Bunde, E.; Schneider, J.; Gersch, M.
\newblock Explainable artificial intelligence: {Objectives}, stakeholders, and future research opportunities.
\newblock {\em Information Systems Management} {\bf 2022}, {\em 39},~53--63.

\bibitem[Ribeiro et~al.(2016)Ribeiro, Singh, and Guestrin]{ribeiro2016should}
Ribeiro, M.T.; Singh, S.; Guestrin, C.
\newblock ``Why should {I} trust you?'': {Explaining} the predictions of any classifier.
\newblock In Proceedings of the 22\textsuperscript{nd} ACM SIGKDD International Conference on Knowledge Discovery and Data Mining,  2016, pp. 1135--1144.

\bibitem[Sokol et~al.(2019)Sokol, Hepburn, Santos-Rodriguez, and Flach]{sokol2019blimey}
Sokol, K.; Hepburn, A.; Santos-Rodriguez, R.; Flach, P.
\newblock {bLIMEy}: {Surrogate} prediction explanations beyond {LIME}.
\newblock In Proceedings of the 2019 Workshop on Human-Centric Machine Learning (HCML 2019) at the 33\textsuperscript{rd} Conference on Neural Information Processing Systems (NeurIPS 2019),  2019.

\bibitem[Sokol and Flach(2024)]{sokol2020towards}
Sokol, K.; Flach, P.
\newblock Interpretable representations in explainable {AI}: {From} theory to practice.
\newblock {\em Data Mining and Knowledge Discovery} {\bf 2024}, pp. 1--39.

\bibitem[Sokol et~al.(2022)Sokol, Hepburn, Santos-Rodriguez, and Flach]{sokol2022what}
Sokol, K.; Hepburn, A.; Santos-Rodriguez, R.; Flach, P.
\newblock What and how of machine learning transparency: {Building} bespoke explainability tools with interoperable algorithmic components.
\newblock {\em Journal of Open Source Education} {\bf 2022}, {\em 5},~175.

\bibitem[Carlevaro et~al.(2023)Carlevaro, Lenatti, Paglialonga, and Mongelli]{carlevaro2023multi}
Carlevaro, A.; Lenatti, M.; Paglialonga, A.; Mongelli, M.
\newblock Multi-class counterfactual explanations using support vector data description.
\newblock {\em IEEE Transactions on Artificial Intelligence} {\bf 2023}.

\bibitem[Hastie and Tibshirani(1986)]{hastie1986generalized}
Hastie, T.; Tibshirani, R.
\newblock Generalized additive models.
\newblock {\em Statistical Science} {\bf 1986}, {\em 1},~297--310.

\bibitem[Lou et~al.(2012)Lou, Caruana, and Gehrke]{lou2012intelligible}
Lou, Y.; Caruana, R.; Gehrke, J.
\newblock Intelligible models for classification and regression.
\newblock In Proceedings of the 18\textsuperscript{th} ACM SIGKDD International Conference on Knowledge Discovery and Data Mining,  2012, pp. 150--158.

\bibitem[Zhang et~al.(2019)Zhang, Tan, Koch, Lou, Chajewska, and Caruana]{zhang2019axiomatic}
Zhang, X.; Tan, S.; Koch, P.; Lou, Y.; Chajewska, U.; Caruana, R.
\newblock Axiomatic interpretability for multiclass additive models.
\newblock In Proceedings of the 25\textsuperscript{th} ACM SIGKDD International Conference on Knowledge Discovery and Data Mining,  2019, pp. 226--234.

\bibitem[Shi et~al.(2019)Shi, Zhang, Li, and Fan]{shi2019explaining}
Shi, S.; Zhang, X.; Li, H.; Fan, W.
\newblock Explaining the predictions of any image classifier via decision trees.
\newblock {\em arXiv Preprint} {\bf 2019},  \href{http://arxiv.org/abs/1911.01058}{{\normalfont [1911.01058]}}.

\bibitem[Tolomei et~al.(2017)Tolomei, Silvestri, Haines, and Lalmas]{tolomei2017interpretable}
Tolomei, G.; Silvestri, F.; Haines, A.; Lalmas, M.
\newblock Interpretable predictions of tree-based ensembles via actionable feature tweaking.
\newblock In Proceedings of the 23\textsuperscript{rd} ACM SIGKDD International Conference on Knowledge Discovery and Data Mining,  2017, pp. 465--474.

\bibitem[Sokol and Flach(2018)]{sokol2018glass}
Sokol, K.; Flach, P.A.
\newblock {Glass-Box}: {Explaining} {AI} decisions with counterfactual statements through conversation with a voice-enabled virtual assistant.
\newblock In Proceedings of the IJCAI,  2018, pp. 5868--5870.

\bibitem[Sokol and Flach(2020)]{sokol2020one}
Sokol, K.; Flach, P.
\newblock One explanation does not fit all: {The} promise of interactive explanations for machine learning transparency.
\newblock {\em KI-K{\"u}nstliche Intelligenz} {\bf 2020}, pp. 1--16.

\bibitem[Flach(2012)]{flach2012machine}
Flach, P.
\newblock {\em Machine learning: {The} art and science of algorithms that make sense of data}; Cambridge University Press,  2012.

\bibitem[Breiman(2001)]{breiman2001random}
Breiman, L.
\newblock Random forests.
\newblock {\em Machine Learning} {\bf 2001}, {\em 45},~5--32.

\bibitem[Laugel et~al.(2018)Laugel, Renard, Lesot, Marsala, and Detyniecki]{laugel2018defining}
Laugel, T.; Renard, X.; Lesot, M.J.; Marsala, C.; Detyniecki, M.
\newblock Defining locality for surrogates in post-hoc interpretablity.
\newblock In Proceedings of the 2018 ICML Workshop on Human Interpretability in Machine Learning (WHI 2018),  2018.

\bibitem[Zhang et~al.(2019)Zhang, Song, Sun, Tan, and Udell]{zhang2019should}
Zhang, Y.; Song, K.; Sun, Y.; Tan, S.; Udell, M.
\newblock ``{Why} should you trust my explanation?'' {Understanding} uncertainty in {LIME} explanations.
\newblock In Proceedings of the AI for Social Good Workshop at the 36\textsuperscript{th} International Conference on Machine Learning (ICML 2019),  2019.

\bibitem[Doshi-Velez and Kim(2018)]{doshi2017towards}
Doshi-Velez, F.; Kim, B.
\newblock Considerations for evaluation and generalization in interpretable machine learning.
\newblock {\em Explainable and Interpretable Models in Computer Vision and Machine Learning} {\bf 2018}, pp. 3--17.

\bibitem[Sokol and Vogt(2024)]{sokol2024does}
Sokol, K.; Vogt, J.E.
\newblock What does evaluation of explainable artificial intelligence actually tell us? {A} case for compositional and contextual validation of {XAI} building blocks.
\newblock In Proceedings of the Extended Abstracts of the 2024 CHI Conference on Human Factors in Computing Systems,  2024, pp. 1--8.

\bibitem[Mittelstadt et~al.(2019)Mittelstadt, Russell, and Wachter]{mittelstadt2019explaining}
Mittelstadt, B.; Russell, C.; Wachter, S.
\newblock Explaining explanations in {AI}.
\newblock In Proceedings of the 2019 ACM Conference on Fairness, Accountability, and Transparency,  2019, pp. 279--288.

\bibitem[Sokol and Vogt(2023)]{sokol2023reasonable}
Sokol, K.; Vogt, J.E.
\newblock {(Un)reasonable} allure of ante-hoc interpretability for high-stakes domains: {Transparency} is necessary but insufficient for comprehensibility.
\newblock In Proceedings of the 3\textsuperscript{rd} Workshop on Interpretable Machine Learning in Healthcare (IMLH) at 2023 International Conference on Machine Learning (ICML),  2023.

\bibitem[Keane et~al.(2021)Keane, Kenny, Delaney, and Smyth]{keane2021if}
Keane, M.T.; Kenny, E.M.; Delaney, E.; Smyth, B.
\newblock If only we had better counterfactual explanations: {Five} key deficits to rectify in the evaluation of counterfactual {XAI} techniques.
\newblock In Proceedings of the IJCAI,  2021, pp. 4466--4474.

\bibitem[Sokol and H\"ullermeier(2025)]{sokol2025all}
Sokol, K.; H\"ullermeier, E.
\newblock All you need for counterfactual explainability is principled and reliable estimate of aleatoric and epistemic uncertainty.
\newblock {\em arXiv Preprint} {\bf 2025},  \href{http://arxiv.org/abs/2502.17007}{{\normalfont [2502.17007]}}.

\bibitem[Chen(2021)]{models}
Chen, Y.
\newblock {PyTorch} {CIFAR} models.
\newblock \url{https://github.com/chenyaofo/pytorch-cifar-models},  2021.

\bibitem[Deng et~al.(2009)Deng, Dong, Socher, Li, Li, and Fei-Fei]{imagenet}
Deng, J.; Dong, W.; Socher, R.; Li, L.J.; Li, K.; Fei-Fei, L.
\newblock {ImageNet}: {A} large-scale hierarchical image database.
\newblock In Proceedings of the 2009 IEEE Conference on Computer Vision and Pattern Recognition. IEEE,  2009, pp. 248--255.

\bibitem[Krizhevsky and Hinton(2009)]{krizhevsky2009learning}
Krizhevsky, A.; Hinton, G.
\newblock Learning multiple layers of features from tiny images,  2009.

\bibitem[Pedregosa et~al.(2011)Pedregosa, Varoquaux, Gramfort, Michel, Thirion, Grisel, Blondel, Prettenhofer, Weiss, Dubourg, Vanderplas, Passos, Cournapeau, Brucher, Perrot, and Duchesnay]{pedregosa2011scikit}
Pedregosa, F.; Varoquaux, G.; Gramfort, A.; Michel, V.; Thirion, B.; Grisel, O.; Blondel, M.; Prettenhofer, P.; Weiss, R.; Dubourg, V.;  et~al.
\newblock {scikit-learn}: {Machine} learning in {Python}.
\newblock {\em Journal of Machine Learning Research} {\bf 2011}, {\em 12},~2825--2830.

\bibitem[Aeberhard and Forina(1991)]{wine}
Aeberhard, S.; Forina, M.
\newblock {Wine}.
\newblock UCI Machine Learning Repository,  1991.

\bibitem[Blackard(1998)]{covertypes}
Blackard, J.
\newblock {Forest} {Covertypes}.
\newblock UCI Machine Learning Repository,  1998.

\bibitem[Garreau and Luxburg(2020)]{garreau2020explaining}
Garreau, D.; Luxburg, U.
\newblock Explaining the explainer: {A} first theoretical analysis of {LIME}.
\newblock In Proceedings of the International Conference on Artificial Intelligence and Statistics. PMLR,  2020, pp. 1287--1296.

\bibitem[Sokol et~al.(2020)Sokol, Hepburn, Poyiadzi, Clifford, Santos-Rodriguez, and Flach]{sokol2020fatf}
Sokol, K.; Hepburn, A.; Poyiadzi, R.; Clifford, M.; Santos-Rodriguez, R.; Flach, P.
\newblock {FAT Forensics}: {A} {Python} toolbox for implementing and deploying fairness, accountability and transparency algorithms in predictive systems.
\newblock {\em Journal of Open Source Software} {\bf 2020}, {\em 5},~1904.

\bibitem[Sokol et~al.(2022)Sokol, Santos-Rodriguez, and Flach]{sokol2022fatf}
Sokol, K.; Santos-Rodriguez, R.; Flach, P.
\newblock {FAT Forensics}: {A} {Python} toolbox for algorithmic fairness, accountability and transparency.
\newblock {\em Software Impacts} {\bf 2022}, {\em 14},~100406.

\bibitem[Achanta et~al.(2012)Achanta, Shaji, Smith, Lucchi, Fua, and S{\"u}sstrunk]{achanta2012slic}
Achanta, R.; Shaji, A.; Smith, K.; Lucchi, A.; Fua, P.; S{\"u}sstrunk, S.
\newblock {SLIC} superpixels compared to state-of-the-art superpixel methods.
\newblock {\em IEEE Transactions on Pattern Analysis and Machine Intelligence} {\bf 2012}, {\em 34},~2274--2282.

\bibitem[Zhang et~al.(2017)Zhang, Cisse, Dauphin, and Lopez-Paz]{zhang2017mixup}
Zhang, H.; Cisse, M.; Dauphin, Y.N.; Lopez-Paz, D.
\newblock {mixup}: {Beyond} empirical risk minimization.
\newblock In Proceedings of the International Conference on Learning Representations,  2017.

\bibitem[Small et~al.(2023)Small, Xuan, Hettiachchi, and Sokol]{small2023helpful}
Small, E.; Xuan, Y.; Hettiachchi, D.; Sokol, K.
\newblock Helpful, misleading or confusing: {How} humans perceive fundamental building blocks of artificial intelligence explanations.
\newblock In Proceedings of the ACM CHI 2023 Workshop on Human-Centered Explainable AI (HCXAI),  2023.

\bibitem[Xuan et~al.(2025)Xuan, Small, Sokol, Hettiachchi, and Sanderson]{xuan2023users}
Xuan, Y.; Small, E.; Sokol, K.; Hettiachchi, D.; Sanderson, M.
\newblock Comprehension is a double-edged sword: Over-interpreting unspecified information in intelligible machine learning explanations.
\newblock {\em International Journal of Human-Computer Studies} {\bf 2025}, {\em 193},~103376.

\end{thebibliography}

\end{adjustwidth}

\pagebreak

\appendixtitles{yes} %
\appendixstart
\appendix

\setcounter{algocf}{0}%
\renewcommand\thealgocf{A\arabic{algocf}}
\setlength{\algotitleheightrule}{0.5pt} %

\setcounter{figure}{0}%
\renewcommand\thefigure{\thesection\arabic{figure}}%

\begin{algorithm}[t]
\SetAlgoLined
\LinesNumbered
\KwData{%
\(\bullet\)~explained data point \(\mathring{x}\) %
\(\bullet\)~set of classes to be explained \(C \subseteq [1,\ldots,n]\) %
\(\bullet\)~black-box model \(f\) %
\(\bullet\)~interpretable representation transformation function \(\IR\) and its inverse \(\IR^{-1}\) %
\(\bullet\)~samples number \(s\) %
\(\bullet\)~distance function \(\ell\) %
\(\bullet\)~kernel \(\kappa\) %
\(\bullet\)~tree depth bound \(z\) %
\(\bullet\)~fidelity expected of the local surrogate \(\epsilon\).%
}
\KwResult{local surrogate multi-output regression tree.}
\vspace{1em}
  \(\mathring{x}^\prime \gets \IR(\mathring{x})\) build interpretable representation of the explained instance\;
  \(X^\prime \leftarrow\) sample \(s\) data points from the interpretable domain \(\mathcal{X}^\prime\) (or generate a complete sample)\label{algo:LIMEtree_optimisation_-1}\; %
  \(X \gets \IR^{-1}(X^\prime)\) transform the sample into the original domain \(\mathcal{X}\)\;
  Predict the probabilities of \(X\) with the black-box model \(f\)\label{algo:LIMEtree_optimisation_0}\;%
  Compute the distance between \(\mathring{x}^\prime\) and the sample \(X^\prime\) using \(\ell\)\;%
  Compute the weights by kernelising the distance with \(\kappa\)\;
  \For(\tcp*[f]{off-the-shelf tree learning algorithm}){\(i \in [1, \ldots, \min(z, \vert \mathcal{X}^\prime \vert)]\)}{\label{algo:LIMEtree_optimisation_1}%
    Build the \(i\)\textsuperscript{th} level of the surrogate multi-output regression tree \(g\) using the weighted data sample \(X^\prime\) for the specified subset \(C\) of class probabilities predicted by the black box in step~\ref{algo:LIMEtree_optimisation_0} as the target;
\tcp*[f]{use binary splits}\\
    Evaluate the fidelity loss \(\mathcal{L}\) of the surrogate\;
    Break the loop when the surrogate achieves the user-defined fidelity loss \(\epsilon\), i.e., \(\mathcal{L} \leq \epsilon\), making it optimal\;%
  }\label{algo:LIMEtree_optimisation_3}
  Return the optimal multi-output regression tree surrogate\label{algo:LIMEtree_optimisation_2}\;
 \caption{The {\sc TREE} (vanilla) variant of {\sc LIMEtree}.\label{algo:LIMEtree}}
\end{algorithm}

\section[\appendixname~\thesection]{{\sc LIMEtree} Algorithms\label{apx:algos}}

Algorithm~\ref{algo:LIMEtree} captures the vanilla variant of the {\sc LIMEtree} explainer that operates on the interpretable representation (see Section~\ref{sec:background}) -- referred to as {\sc TREE} -- and Algorithm~\ref{algo:LIMEtree_star} outlines the post-processing procedure -- called \underline{{\sc TREE}} -- applied to achieve full fidelity of the tree structure-based explanations (referred to as \emph{model-driven explanations} throughout this paper). %
{\sc TREE} can be built upon most off-the-shelf tree learning methods that allow for binary splits. %
While it is relatively lightweight, manipulating data points (e.g., images) via the \(\IR\) and \(\IR^{-1}\) functions and querying the black-box model \(f\) may become a bottleneck. %
The explainee has no control over the computational and memory complexity of querying the black box, which is executed \(s\) times, where \(s \in \mathbb{N}^+\) is the number of sampled data points. %
Given the recent advances in dedicated AI hardware, this step should not be a burden when utilising GPUs (Graphics Processing Units) and manageable with just CPUs (Central Processing Units). %

Transforming the interpretable representation (binary vectors) into the original data (e.g, image) domain may require a considerable amount of operational memory: %
the explained instance (e.g., image) has to be duplicated for every data point sampled from the interpretable domain and its feature (e.g., pixel) values need to be altered to reflect the removed concepts (e.g., segment occlusions). %
The efficiency of these two steps can be significantly improved with batch processing and parallelisation, therefore reducing the use of memory and improving the processing time. %
Other steps in Algorithm~\ref{algo:LIMEtree} are relatively efficient: discretising continuous features or segmenting an image, sampling a binary matrix from the interpretable domain and fitting a multi-output regression tree to binary data with feature thresholds fixed at \(\sfrac{1}{2}\). %
For tabular data, {\sc TREE} may be adapted to operate on the original domain instead, using this representation to sample data and compute distances; %
the explainer is highly efficient for this data type. %

\begin{algorithm}[t]
\SetAlgoLined
\LinesNumbered
\KwData{%
\(\bullet\)~black-box model \(f\) %
\(\bullet\)~local surrogate multi-output regression tree \(g\) %
\(\bullet\)~interpretable representation transformation function \(\IR\) and its inverse \(\IR^{-1}\) %
\(\bullet\)~set of classes to be explained \(C \subseteq [1,\ldots,n]\). %
}
\KwResult{surrogate multi-output regression tree with \emph{full fidelity} of model-driven explanations.}%
\vspace{1em}
  \(T \leftarrow\) identify the set of leaves of the surrogate model \(g\)\;
  \(X_T^{\prime} \leftarrow\) compose the set of data points constituting the minimal interpretable representation (see Definition~\ref{def:minimal_x_star})\label{algo:LIMEtree_star_optimisation_0}\;%
  \(X_T \gets \IR^{-1}(X_T^{\prime})\) transform this set into the original domain \(\mathcal{X}\)\label{algo:LIMEtree_star_optimisation_00}\;
  Predict the probabilities of \(X_T\) with the black-box model \(f\)\label{algo:LIMEtree_star_optimisation_1}\;%
  \For{\(t \in T\)}{
    Replace the probabilities predicted by the leaf \(t\) of the surrogate \(g\) with the black-box predictions (step~\ref{algo:LIMEtree_star_optimisation_1}) of this leaf's minimal data point \(x_t \in X_T\) (steps~\ref{algo:LIMEtree_star_optimisation_0} \& \ref{algo:LIMEtree_star_optimisation_00}) for the specified subset of classes \(C\)\;%
  }
  Return the \emph{modified} multi-output surrogate regression tree \(g\) with \emph{full fidelity} of model-driven explanations\; %
 \caption{The \underline{{\sc TREE}} variant of {\sc LIMEtree}.\label{algo:LIMEtree_star}}
\end{algorithm}

\section[\appendixname~\thesection]{Proofs\label{apx:proofs}}

\begin{proof}%
[Proof of Lemma~\ref{lemma:local_tree_fidelity} (Structural Fidelity)]%
Full (\emph{empirical}) fidelity is achieved when both the explained model~\(f\) (which is assumed to be non-stochastic) and its surrogate~\(g\) deliver identical predictions across all the selected classes~\(C\) for a predetermined set of data points~\(X\). %
\emph{Structural} fidelity narrows down the scope of consideration to instances~\(X^\prime_{\textit{min}, T}\) %
(from the binary interpretable representation space) %
that are described by the logical conditions specified by the leaves of the surrogate tree~\(T\): %
\[
f_c(\IR^{-1}(x^\prime)) = g_c(x^\prime) \qquad \forall c \in C \;\; \forall x^\prime \in X^\prime_{\textit{min}, T} %
\text{.}
\]
When the interpretable representation transformation function is deterministic, %
i.e., \(\IR(\IR^{-1}(x^\prime)) = x^\prime \;\; \forall x^\prime \in \mathcal{X}^\prime\), %
post-processing the surrogate with Algorithm~\ref{algo:LIMEtree_star} guarantees %
its outputs to align with those of the explained model for instances from the \emph{minimal representation} \(X^\prime_{\textit{min}, T}\) (Definition~\ref{def:minimal_x_star}), which are the backbone of model-driven explanations. %
Therefore, this surrogate achieves \emph{full structural} fidelity: %
\[
f_c(\IR^{-1}(x^\prime)) = g_c(x^\prime) \qquad \forall c \in C \;\; \forall x^\prime \in X^\prime_{\textit{min}, T} %
\text{.}
\]
\end{proof}%

\begin{proof}%
[Proof of Corollary~\ref{cor:local_data_fidelity} (Full Fidelity)]%
Growing a complete, \(d\)-deep binary surrogate tree~\(T\) for a \(d\)-dimensional binary interpretable representation~\(\mathcal{X}^\prime\) allows each data point to be assigned a unique tree leaf: %
\[
X^\prime_{\textit{min}, T} \equiv \mathcal{X}^\prime %
\text{.}
\]
Consequently, when %
the interpretable representation transformation function is deterministic, the predictions of the surrogate will be identical to those of the explained model (which is assumed to be non-stochastic) across the entire interpretable representation space. %
Therefore, this surrogate achieves \emph{full} fidelity (both with respect to model- and data-driven explanations): %
\[
f_c(\IR^{-1}(x^\prime)) = g_c(x^\prime) \qquad \forall c \in C \;\; \forall x^\prime \in \mathcal{X}^\prime %
\text{.}
\]
\end{proof}

\section[\appendixname~\thesection]{Loss Behaviour\label{apx:loss}}

\begin{figure}%
\begin{adjustwidth}{-\extralength}{0cm}
    \centering
    \begin{minipage}{0.49\fulllength}
    \centering
    \subfloat[LIME loss for the 1\textsuperscript{st} class.]{%
\makebox[.49\fulllength][c]{%
    \includegraphics[width=.49\fulllength]{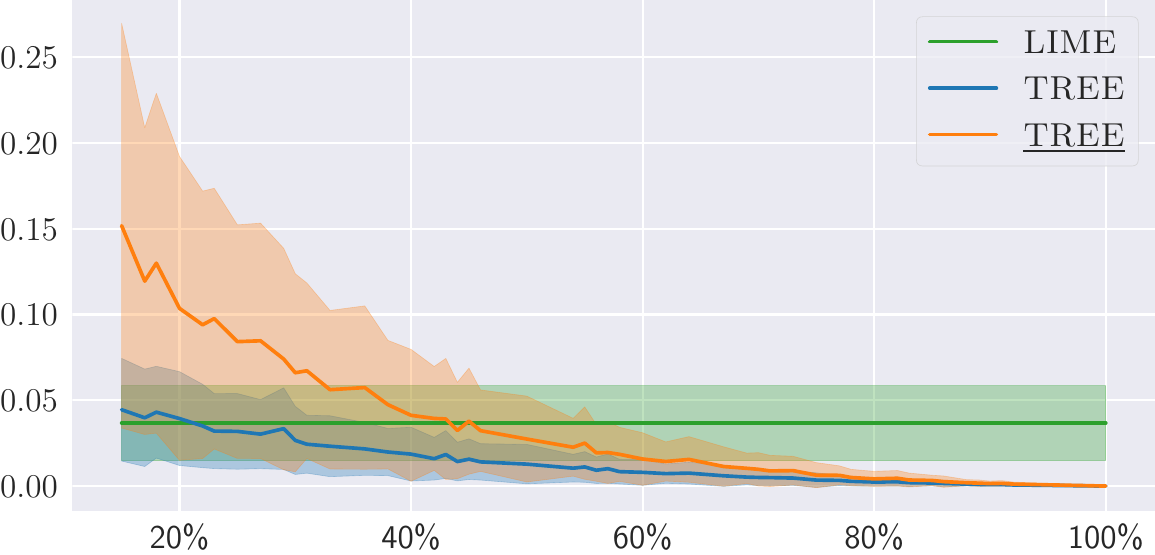}%
        \vspace{\baselineskip}%
}\label{fig:tree_depth_fidelity:loss:1}}%
\par%
    \subfloat[LIME loss for the 2\textsuperscript{nd} class.]{%
\makebox[.49\fulllength][c]{%
    \includegraphics[width=.49\fulllength]{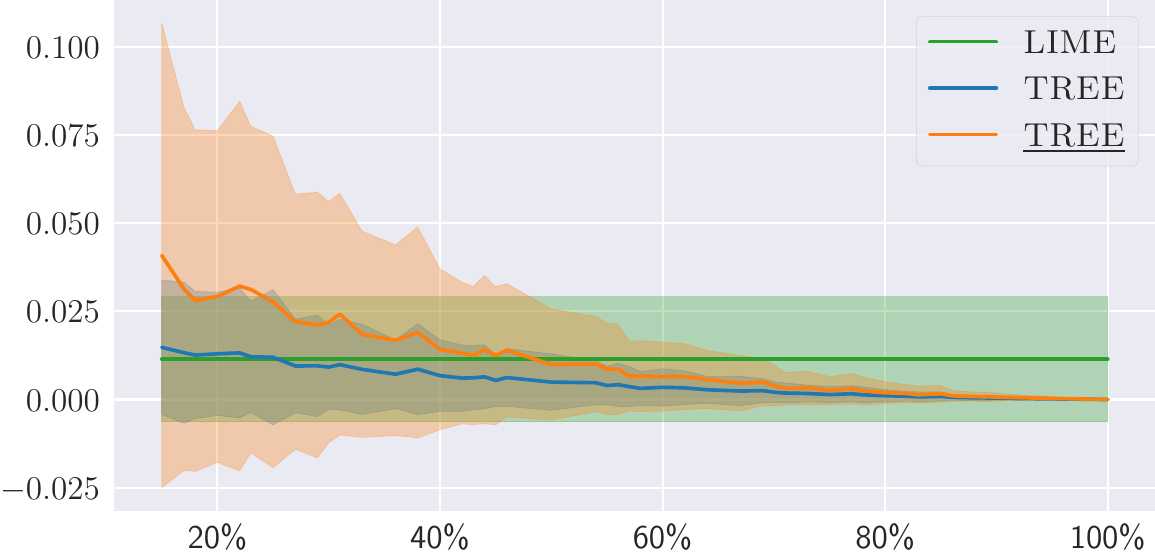}%
        \vspace{\baselineskip}%
}\label{fig:tree_depth_fidelity:loss:2}}%
\par
    \subfloat[LIME loss for the 3\textsuperscript{rd} class.]{%
\makebox[.49\fulllength][c]{%
    \includegraphics[width=.49\fulllength]{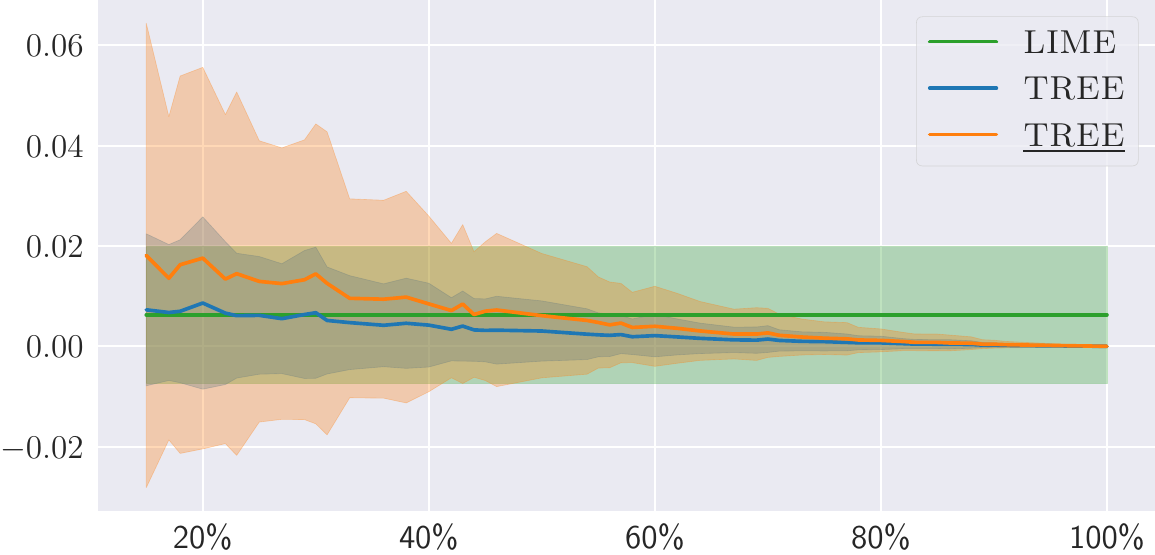}%
        \vspace{\baselineskip}%
}\label{fig:tree_depth_fidelity:loss:3}}%
\end{minipage}
    \hspace{0.015\fulllength}%
\begin{minipage}{0.49\fulllength}
    \centering

    \subfloat[{\sc LIMEtree} loss for the top class.]{%
\makebox[.49\fulllength][c]{%
    \includegraphics[width=.49\fulllength]{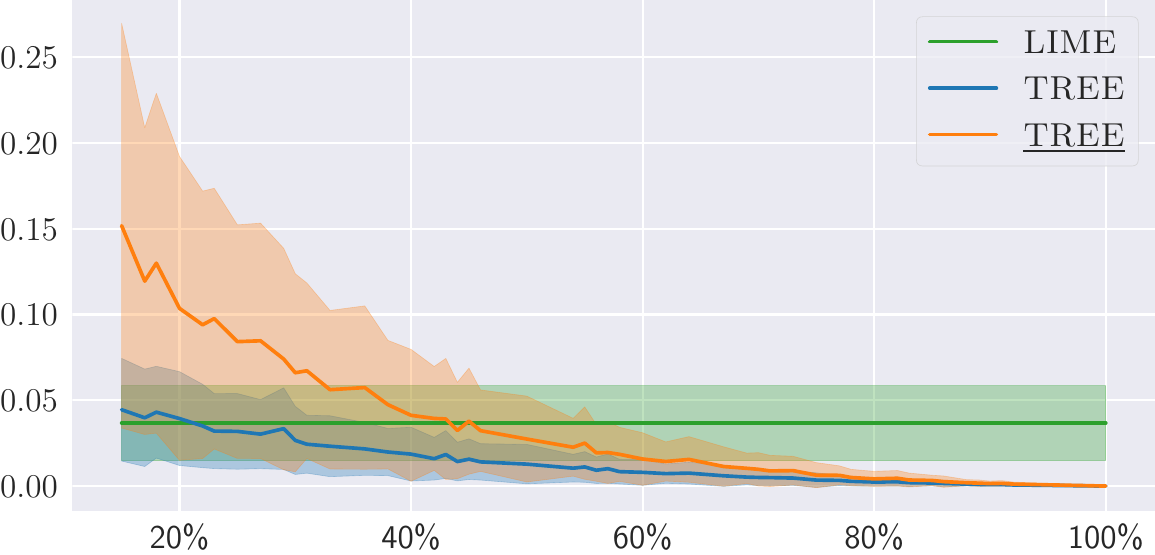}%
        \vspace{\baselineskip}%
}\label{fig:tree_depth_fidelity:ltl:1}}%
\par%
    \subfloat[{\sc LIMEtree} loss for the top 2 classes.]{%
\makebox[.49\fulllength][c]{%
    \includegraphics[width=.49\fulllength]{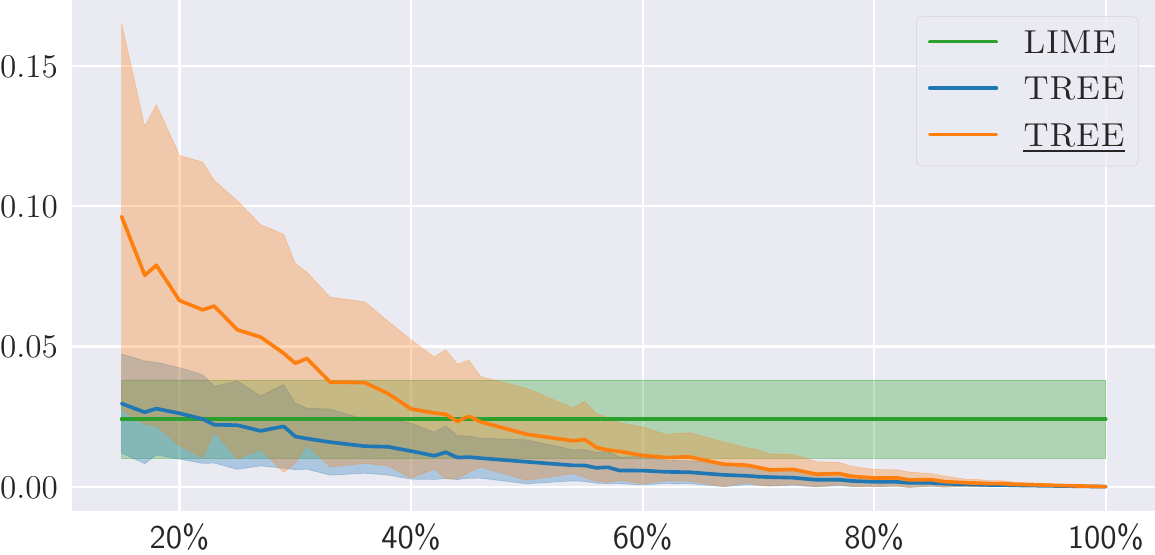}%
        \vspace{\baselineskip}%
}\label{fig:tree_depth_fidelity:ltl:2}}%
\par%
    \subfloat[{\sc LIMEtree} loss for the top 3 classes.]{%
\makebox[.49\fulllength][c]{%
    \includegraphics[width=.49\fulllength]{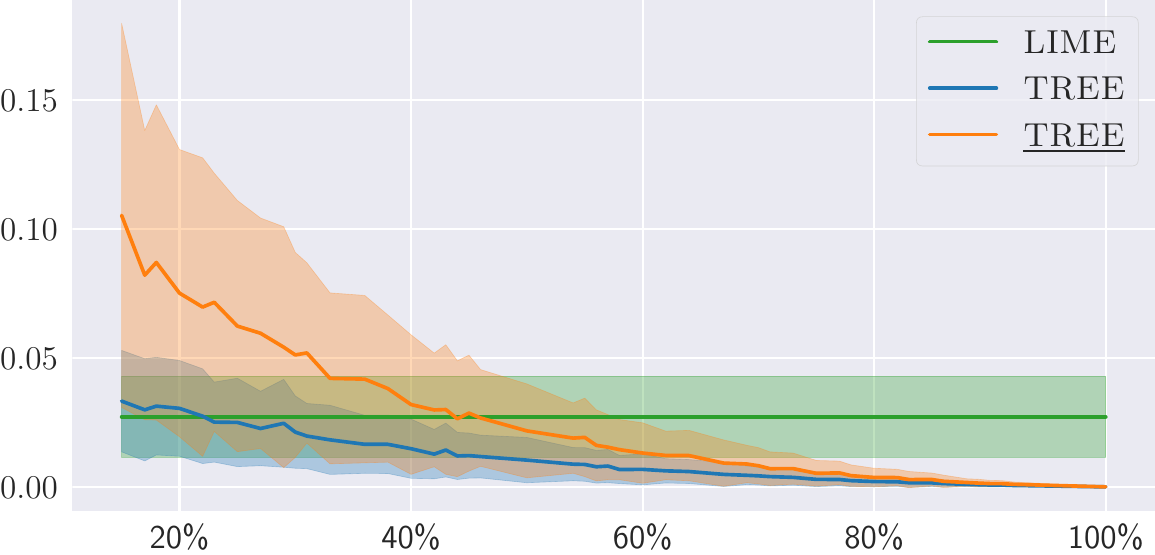}%
        \vspace{\baselineskip}%
}\label{fig:tree_depth_fidelity:ltl:3}}%
\end{minipage}
\end{adjustwidth}
    \caption{%
    Fidelity of a surrogate \(\mathcal{L}\) (y-axis) built for the ImageNet data set (see Section~\ref{sec:experiments} for information about the experimental setup) and plotted against its complexity \(\Omega\) (x-axis) expressed as the ratio between the depth of the tree and its maximum depth determined by the number of features in the interpretable representation, which is equivalent to the depth of a complete tree (Equation~\ref{eq:blimey_complexity}). %
    The complexity of a linear surrogate is fixed and given by the number of features found in the interpretable representation, i.e., 100\%. %
    Panels~\protect\subref*{fig:tree_depth_fidelity:loss:1}, \protect\subref*{fig:tree_depth_fidelity:loss:2} \& \protect\subref*{fig:tree_depth_fidelity:loss:3} depict fidelity measured with the LIME loss (Equation~\ref{eq:lime:loss}), and Panels~\protect\subref*{fig:tree_depth_fidelity:ltl:1}, \protect\subref*{fig:tree_depth_fidelity:ltl:2} \& \protect\subref*{fig:tree_depth_fidelity:ltl:3} show the same property calculated with the {\sc LIMEtree} loss (Equation~\ref{eq:blimey_loss}) for different configurations of the top three classes predicted by a black box. %
    Note different scales on the y-axes. %
    The results are shown for three surrogate models: %
    {\sc LIME} -- a linear surrogate fitted to all interpretable features; %
    {\sc TREE} -- a tree surrogate optimised for fidelity and complexity (Algorithm~\ref{algo:LIMEtree} in Appendix~\ref{apx:algos}); and %
    \underline{{\sc TREE}} -- a {\sc TREE} surrogate post-processed to achieve full fidelity of model-driven explanations (Algorithm~\ref{algo:LIMEtree_star} in Appendix~\ref{apx:algos}).%
    \label{fig:tree_depth_fidelity}}%
\end{figure}

\begin{figure}%
\begin{adjustwidth}{-\extralength}{0cm}
    \centering
    \begin{minipage}{0.49\fulllength}
    \centering
    \subfloat[LIME loss for the 1\textsuperscript{st} class.]{%
\makebox[.49\fulllength][c]{%
    \includegraphics[width=.49\fulllength]{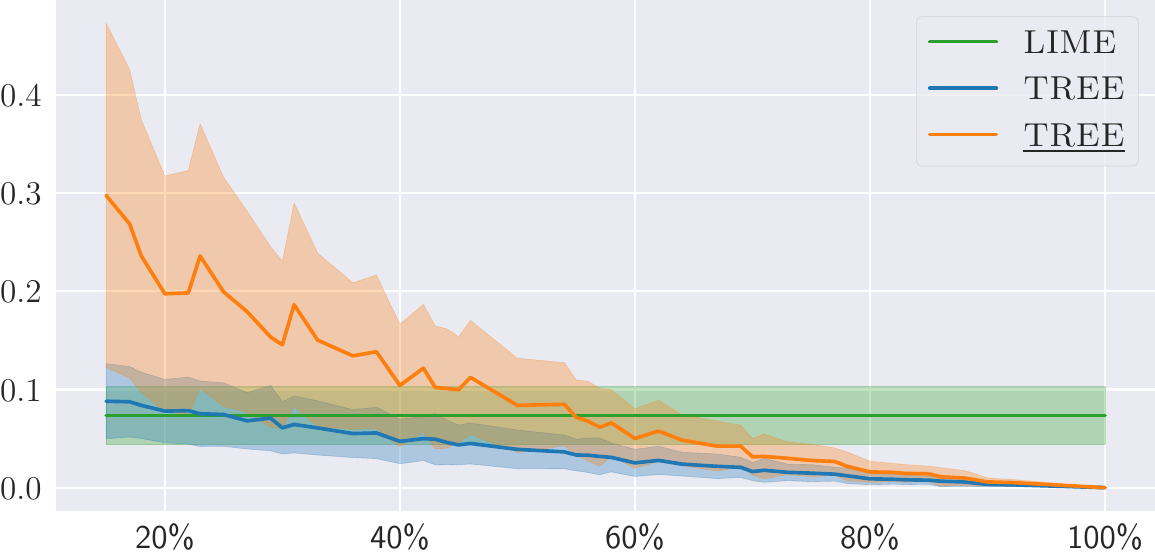}%
}\label{fig:tree_depth_fidelity-cifar10:loss:1}}%
\par%
    \subfloat[LIME loss for the 2\textsuperscript{nd} class.]{%
\makebox[.49\fulllength][c]{%
    \includegraphics[width=.49\fulllength]{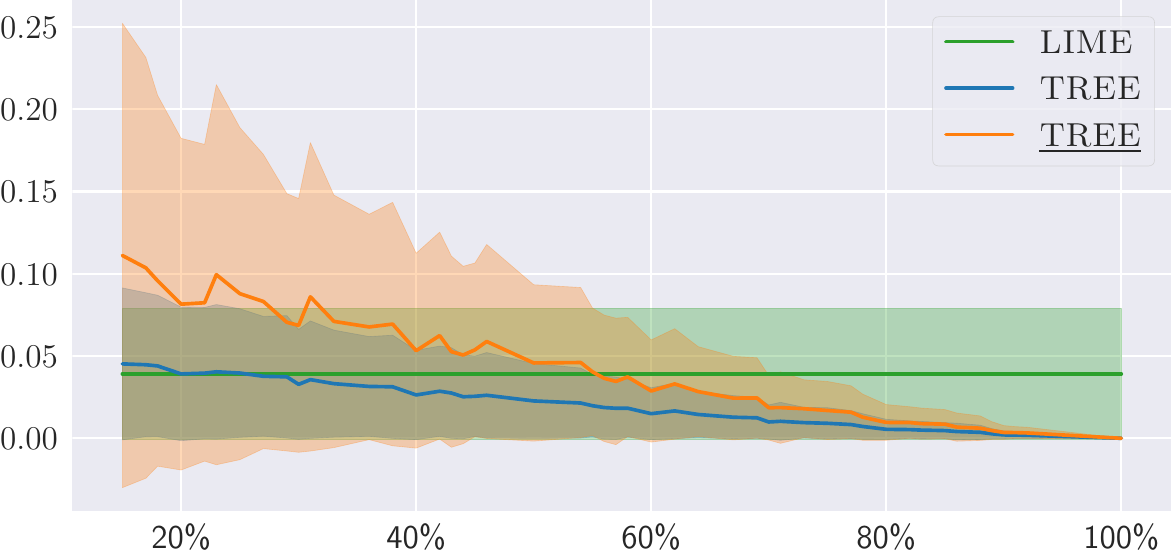}%
}\label{fig:tree_depth_fidelity-cifar10:loss:2}}%
\par%
    \subfloat[LIME loss for the 3\textsuperscript{rd} class.]{%
\makebox[.49\fulllength][c]{%
    \includegraphics[width=.49\fulllength]{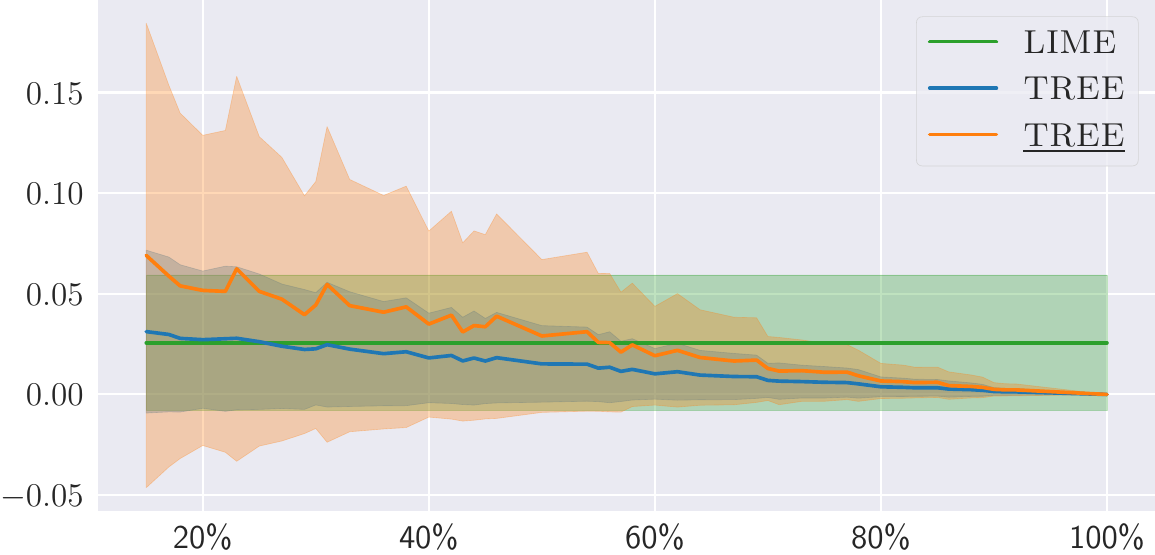}%
}\label{fig:tree_depth_fidelity-cifar10:loss:3}}%
\end{minipage}
    \hspace{0.015\fulllength}%
\begin{minipage}{0.49\fulllength}
    \centering
    \subfloat[{\sc LIMEtree} loss for the top class.]{%
\makebox[.49\fulllength][c]{%
    \includegraphics[width=.49\fulllength]{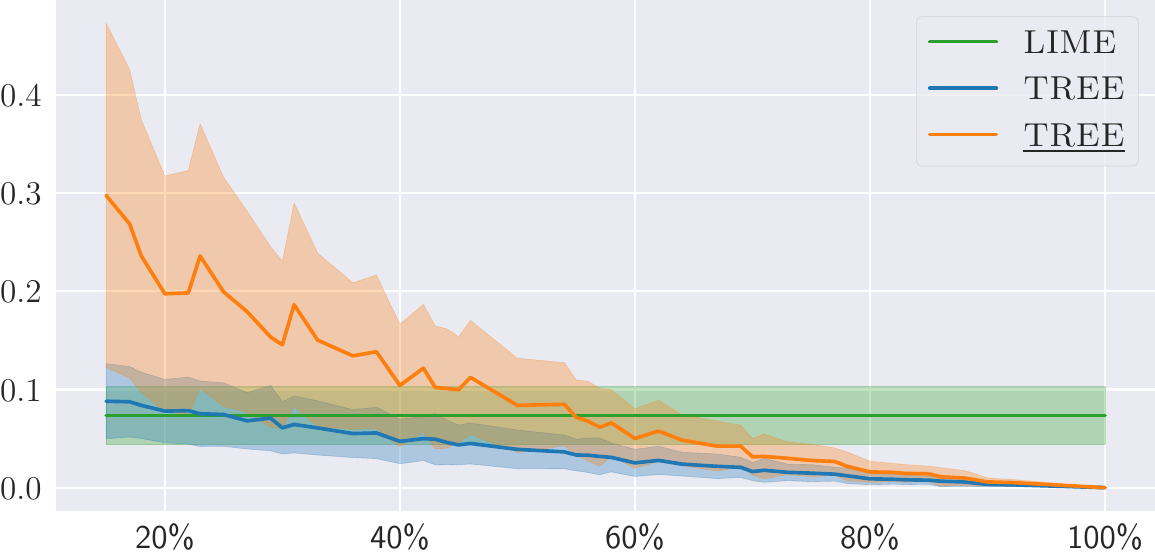}%
}\label{fig:tree_depth_fidelity-cifar10:ltl:1}}%
\par%
    \subfloat[{\sc LIMEtree} loss for the top 2 classes.]{%
\makebox[.49\fulllength][c]{%
    \includegraphics[width=.49\fulllength]{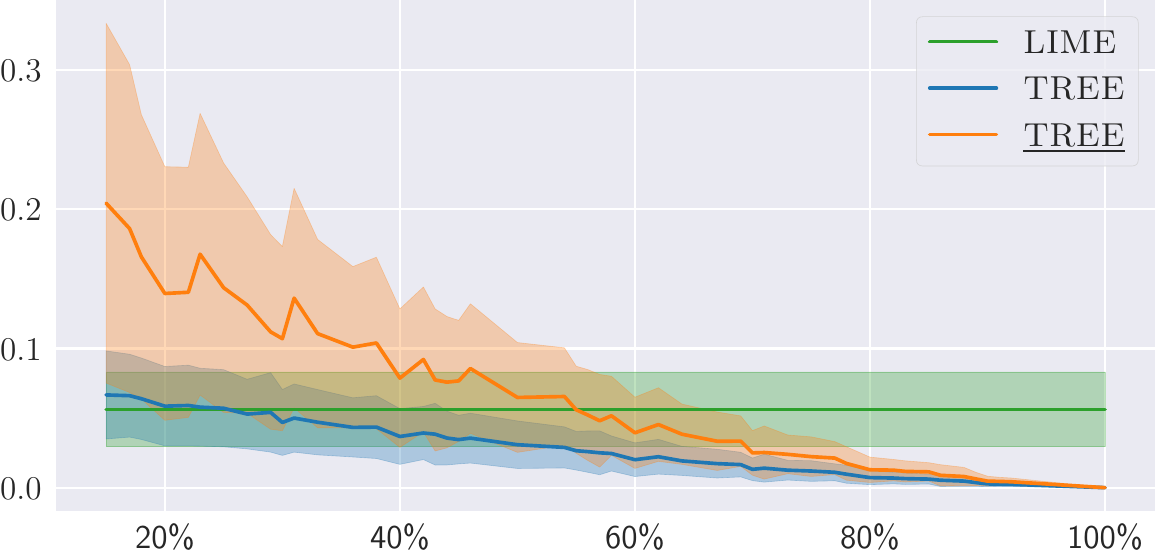}%
}\label{fig:tree_depth_fidelity-cifar10:ltl:2}}%
\par%
    \subfloat[{\sc LIMEtree} loss for the top 3 classes.]{%
\makebox[.49\fulllength][c]{%
    \includegraphics[width=.49\fulllength]{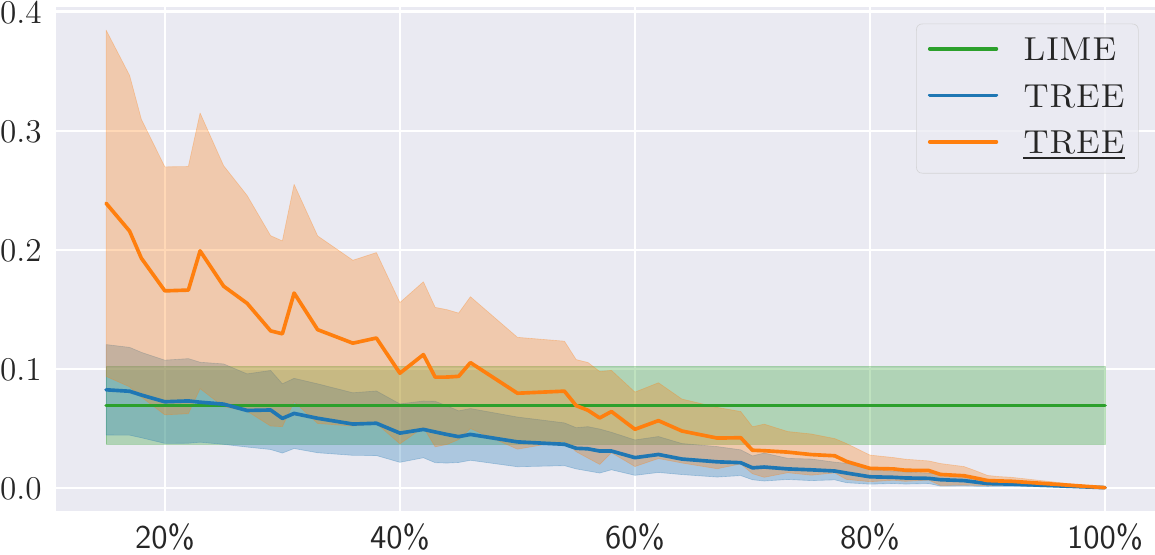}%
}\label{fig:tree_depth_fidelity-cifar10:ltl:3}}%
\end{minipage}
\end{adjustwidth}
    \caption{%
    Fidelity of a surrogate \(\mathcal{L}\) (y-axis) built for the CIFAR-10 data set (see Section~\ref{sec:experiments} for information about the experimental setup) and plotted against its complexity \(\Omega\) (x-axis) expressed as the ratio between the depth of the tree and its maximum depth determined by the number of features in the interpretable representation, which is equivalent to the depth of a complete tree (Equation~\ref{eq:blimey_complexity}). %
    The caption of Figure~\ref{fig:tree_depth_fidelity} provides further information about the details of the plot. %
    \label{fig:tree_depth_fidelity-cifar10}}%
\end{figure}

\begin{figure}%
\begin{adjustwidth}{-\extralength}{0cm}
    \centering
    \begin{minipage}{0.49\fulllength}
    \centering
    \subfloat[LIME loss for the 1\textsuperscript{st} class.]{%
\makebox[.49\fulllength][c]{%
    \includegraphics[width=.49\fulllength]{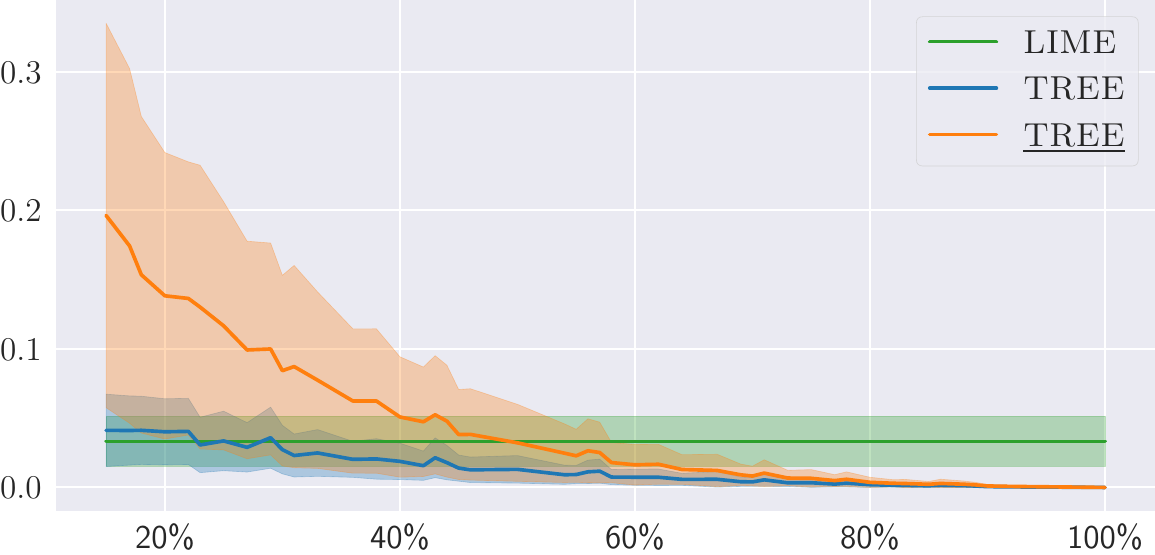}%
}\label{fig:tree_depth_fidelity-cifar100:loss:1}}%
\par%
    \subfloat[LIME loss for the 2\textsuperscript{nd} class.]{%
\makebox[.49\fulllength][c]{%
    \includegraphics[width=.49\fulllength]{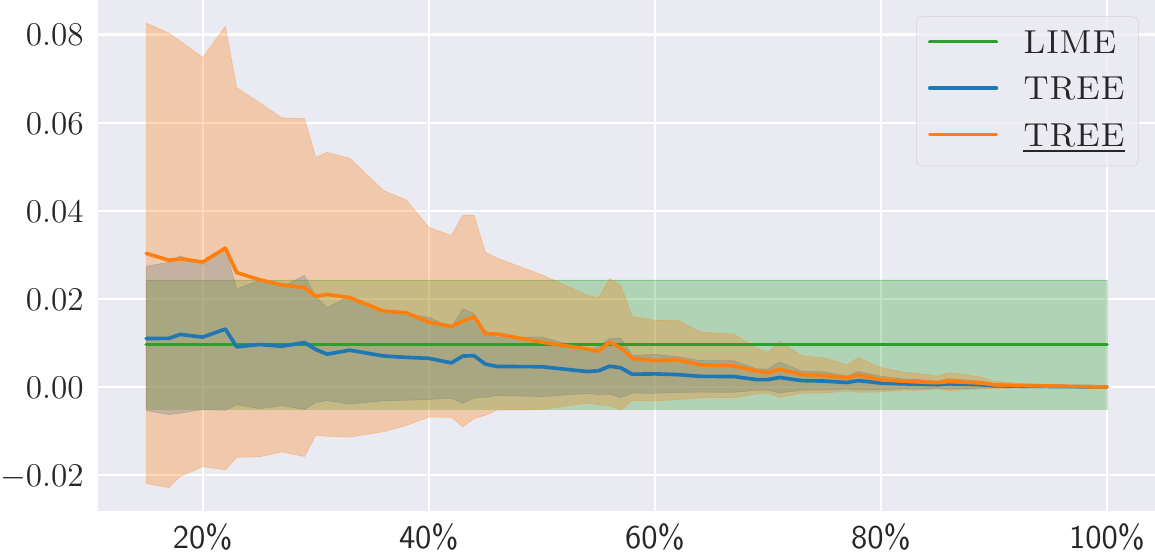}%
}\label{fig:tree_depth_fidelity-cifar100:loss:2}}%
\par%
    \subfloat[LIME loss for the 3\textsuperscript{rd} class.]{%
\makebox[.49\fulllength][c]{%
    \includegraphics[width=.49\fulllength]{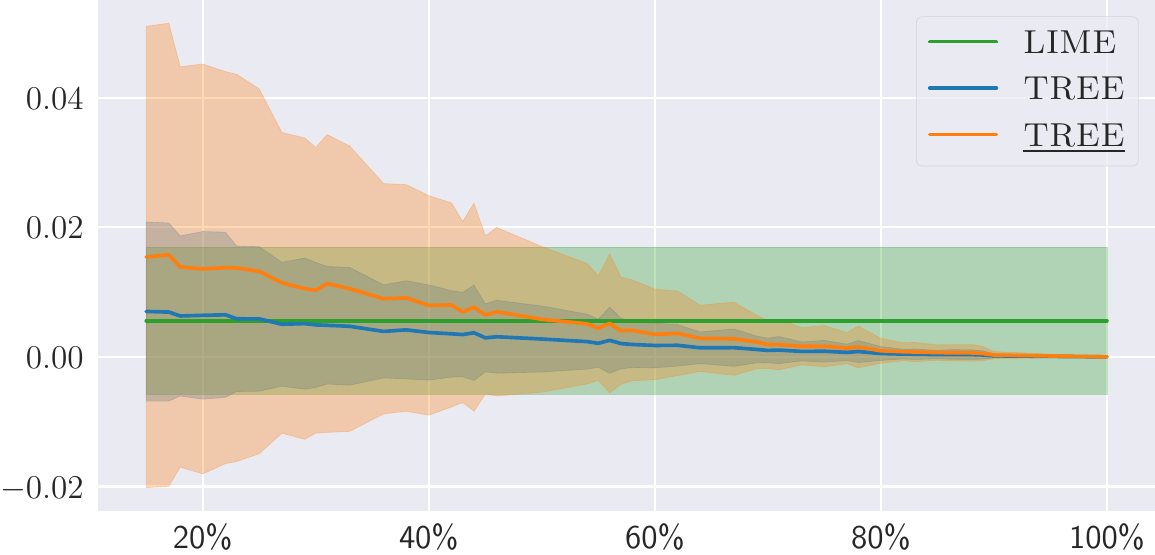}%
}\label{fig:tree_depth_fidelity-cifar100:loss:3}}%
\end{minipage}
    \hspace{0.015\fulllength}%
\begin{minipage}{0.49\fulllength}
    \centering
    \subfloat[{\sc LIMEtree} loss for the top class.]{%
\makebox[.49\fulllength][c]{%
    \includegraphics[width=.49\fulllength]{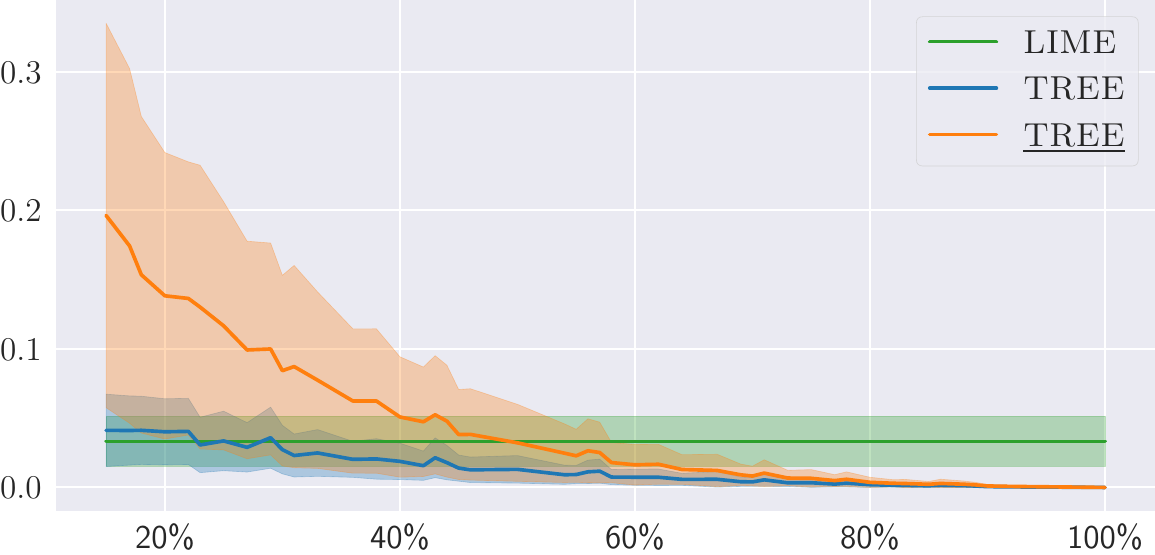}%
}\label{fig:tree_depth_fidelity-cifar100:ltl:1}}%
\par%
    \subfloat[{\sc LIMEtree} loss for the top 2 classes.]{%
\makebox[.49\fulllength][c]{%
    \includegraphics[width=.49\fulllength]{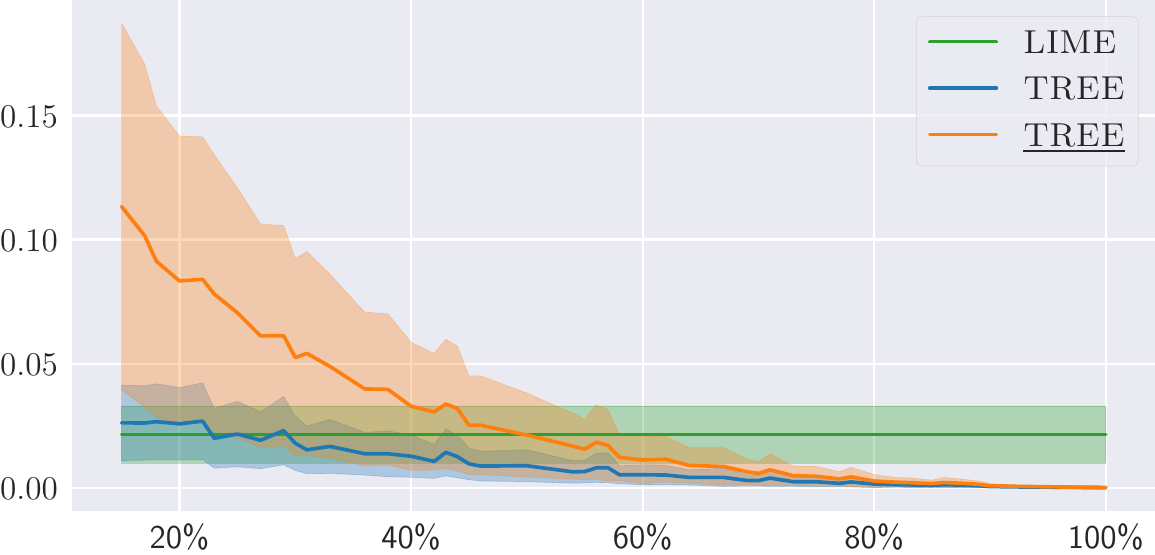}%
}\label{fig:tree_depth_fidelity-cifar100:ltl:2}}%
\par%
    \subfloat[{\sc LIMEtree} loss for the top 3 classes.]{%
\makebox[.49\fulllength][c]{%
    \includegraphics[width=.49\fulllength]{_experiments/n-rand-cifar100/loss-cls3-limet_weighted_Xrandom}%
}\label{fig:tree_depth_fidelity-cifar100:ltl:3}}%
\end{minipage}
\end{adjustwidth}
    \caption{%
    Fidelity of a surrogate \(\mathcal{L}\) (y-axis) built for the CIFAR-100 data set (see Section~\ref{sec:experiments} for information about the experimental setup) and plotted against its complexity \(\Omega\) (x-axis) expressed as the ratio between the depth of the tree and its maximum depth determined by the number of features in the interpretable representation, which is equivalent to the depth of a complete tree (Equation~\ref{eq:blimey_complexity}). %
    The caption of Figure~\ref{fig:tree_depth_fidelity} provides further information about the details of the plot. %
    \label{fig:tree_depth_fidelity-cifar100}}%
\end{figure}

\begin{figure}%
\begin{adjustwidth}{-\extralength}{0cm}
    \centering
    \begin{minipage}{0.49\fulllength}
    \centering
    \subfloat[LIME loss for the 1\textsuperscript{st} class.]{%
\makebox[.49\fulllength][c]{%
    \includegraphics[width=.49\fulllength]{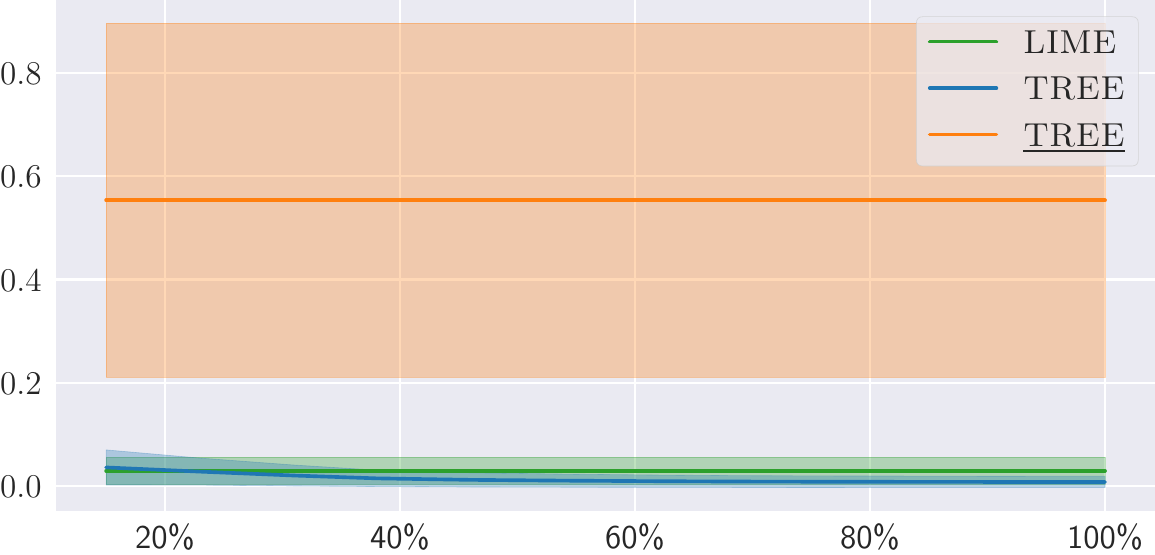}%
}\label{fig:tree_depth_fidelity-wine-test:loss:1}}%
\par%
    \subfloat[LIME loss for the 2\textsuperscript{nd} class.]{%
\makebox[.49\fulllength][c]{%
    \includegraphics[width=.49\fulllength]{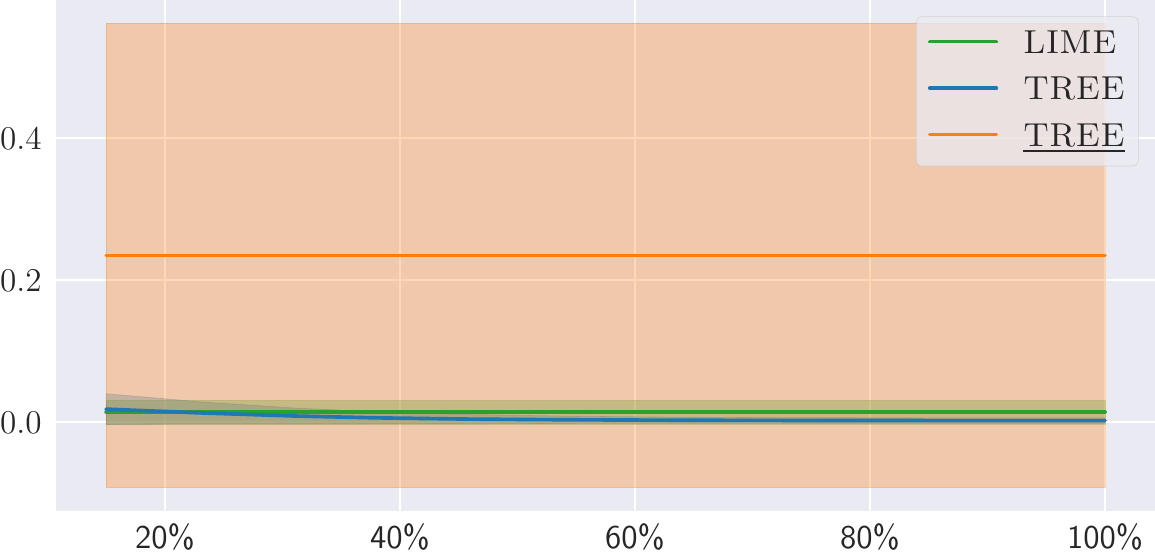}%
}\label{fig:tree_depth_fidelity-wine-test:loss:2}}%
\par%
    \subfloat[LIME loss for the 3\textsuperscript{rd} class.]{%
\makebox[.49\fulllength][c]{%
    \includegraphics[width=.49\fulllength]{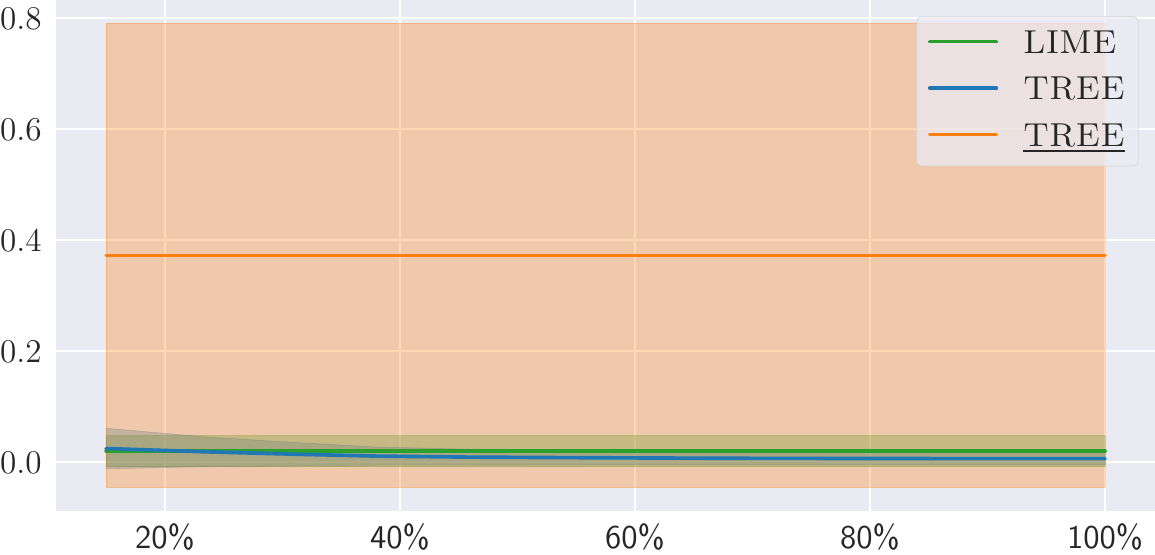}%
}\label{fig:tree_depth_fidelity-wine-test:loss:3}}%
\end{minipage}
    \hspace{0.015\fulllength}%
\begin{minipage}{0.49\fulllength}
    \centering
    \subfloat[{\sc LIMEtree} loss for the top class.]{%
\makebox[.49\fulllength][c]{%
    \includegraphics[width=.49\fulllength]{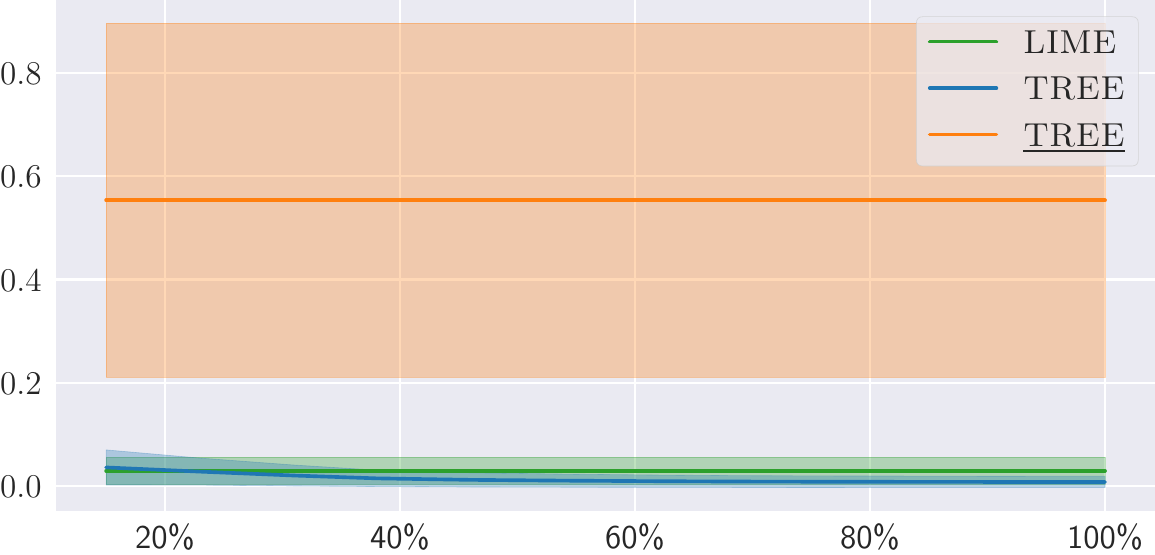}%
}\label{fig:tree_depth_fidelity-wine-test:ltl:1}}%
\par%
    \subfloat[{\sc LIMEtree} loss for the top 2 classes.]{%
\makebox[.49\fulllength][c]{%
    \includegraphics[width=.49\fulllength]{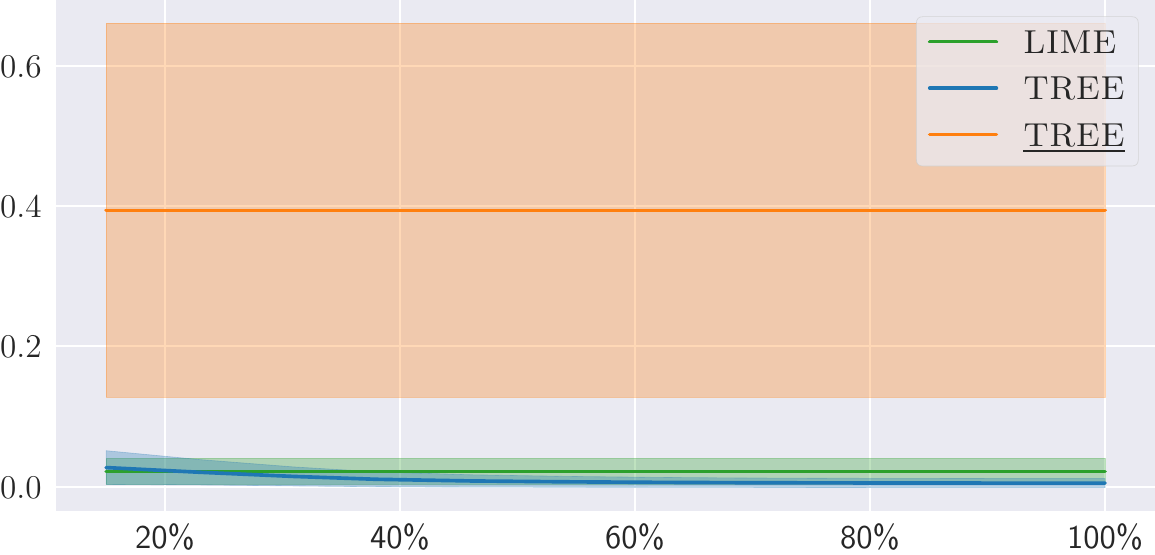}%
}\label{fig:tree_depth_fidelity-wine-test:ltl:2}}%
\par%
    \subfloat[{\sc LIMEtree} loss for the top 3 classes.]{%
\makebox[.49\fulllength][c]{%
    \includegraphics[width=.49\fulllength]{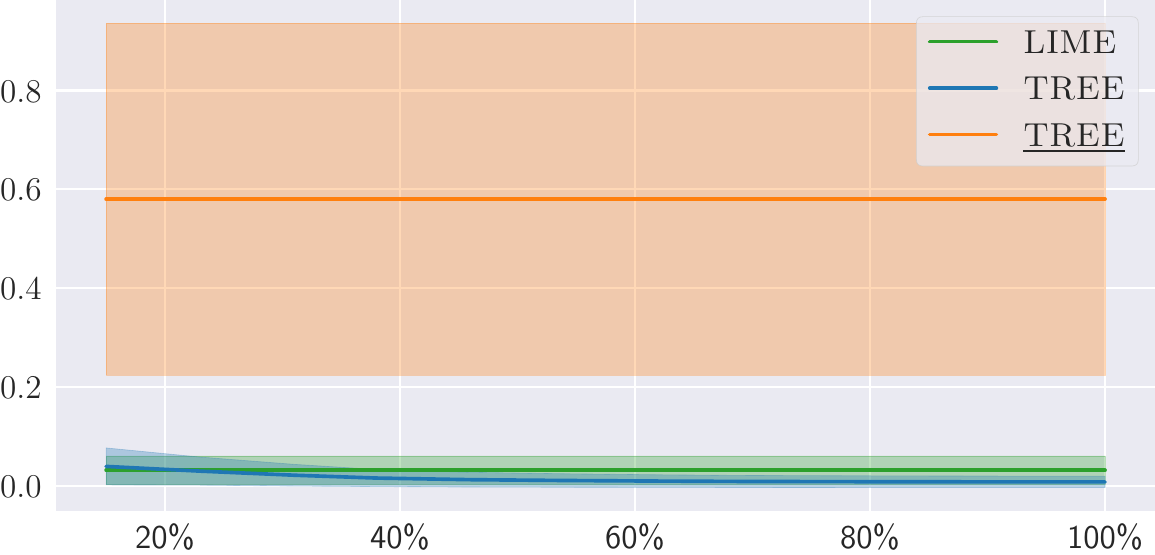}%
}\label{fig:tree_depth_fidelity-wine-test:ltl:3}}%
\end{minipage}
\end{adjustwidth}
    \caption{%
    Fidelity of a surrogate \(\mathcal{L}\) (y-axis) built for the Wine data set (see Section~\ref{sec:experiments} for information about the experimental setup) and plotted against its complexity \(\Omega\) (x-axis) expressed as the ratio between the depth of the tree and its maximum depth determined by the number of features in the interpretable representation, which is equivalent to the depth of a complete tree (Equation~\ref{eq:blimey_complexity}). %
    The caption of Figure~\ref{fig:tree_depth_fidelity} provides further information about the details of the plot. %
    \label{fig:tree_depth_fidelity-wine-test}}%
\end{figure}

\begin{figure}%
\begin{adjustwidth}{-\extralength}{0cm}
    \centering
    \begin{minipage}{0.49\fulllength}
    \centering
    \subfloat[LIME loss for the 1\textsuperscript{st} class.]{%
\makebox[.49\fulllength][c]{%
    \includegraphics[width=.49\fulllength]{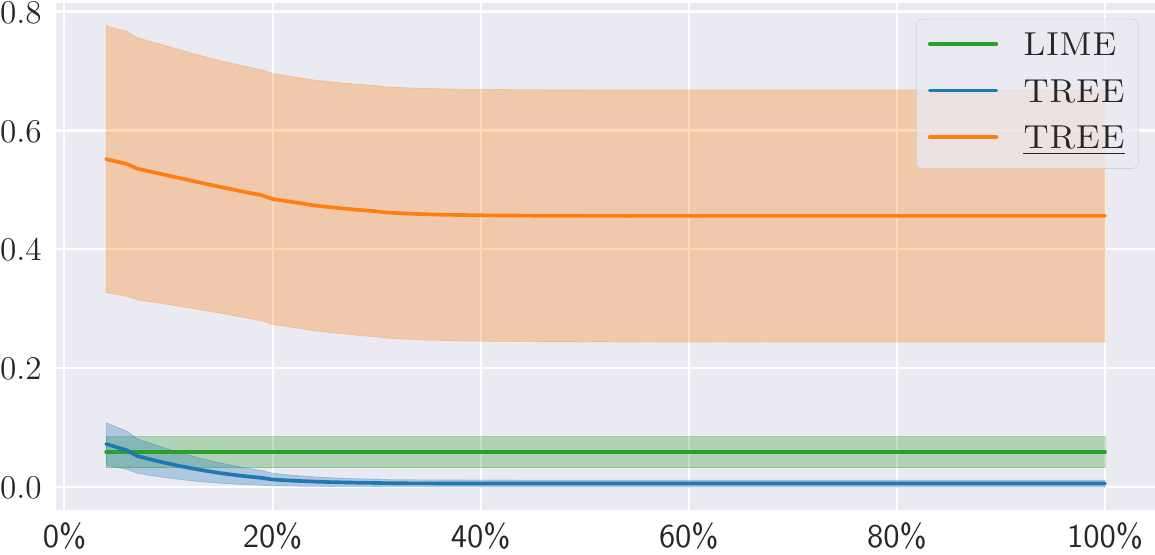}%
}\label{fig:tree_depth_fidelity-forest:loss:1}}%
\par%
    \subfloat[LIME loss for the 2\textsuperscript{nd} class.]{%
\makebox[.49\fulllength][c]{%
    \includegraphics[width=.49\fulllength]{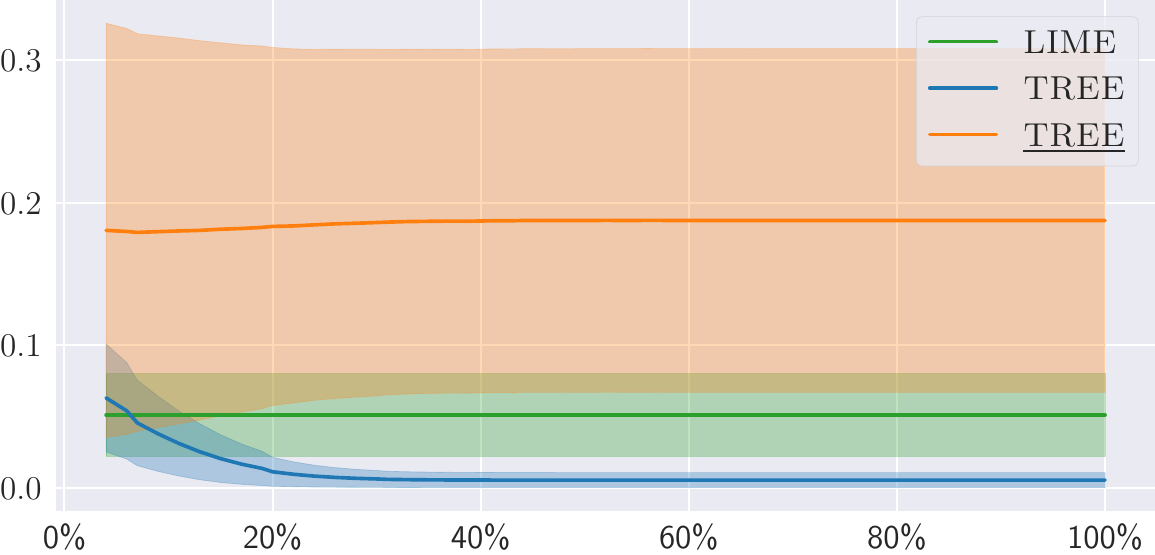}%
}\label{fig:tree_depth_fidelity-forest:loss:2}}%
\par%
    \subfloat[LIME loss for the 3\textsuperscript{rd} class.]{%
\makebox[.49\fulllength][c]{%
    \includegraphics[width=.49\fulllength]{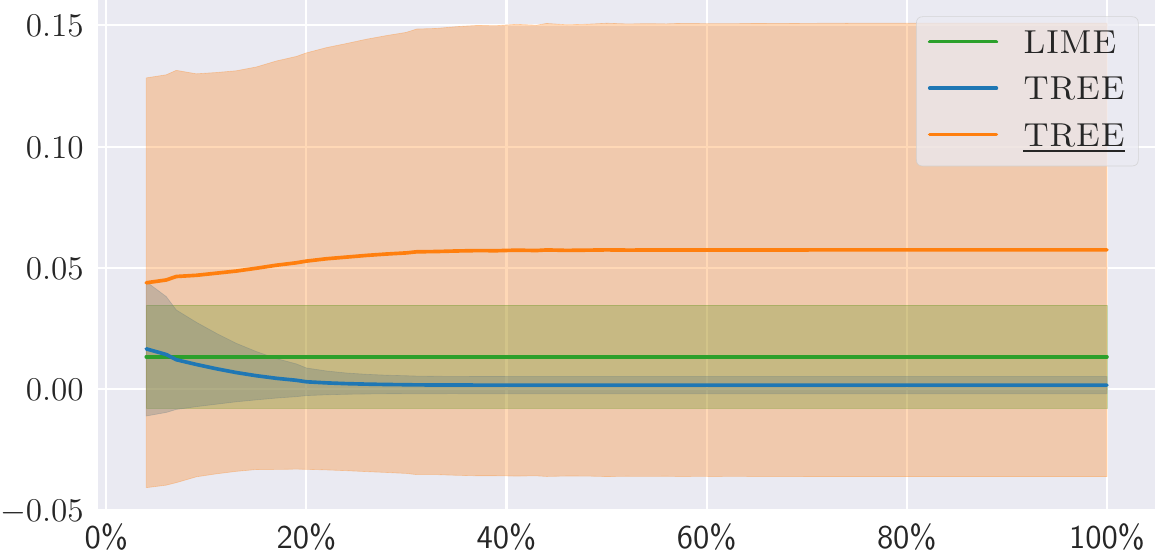}%
}\label{fig:tree_depth_fidelity-forest:loss:3}}%
\end{minipage}
    \hspace{0.015\fulllength}%
\begin{minipage}{0.49\fulllength}
    \centering
    \subfloat[{\sc LIMEtree} loss for the top class.]{%
\makebox[.49\fulllength][c]{%
    \includegraphics[width=.49\fulllength]{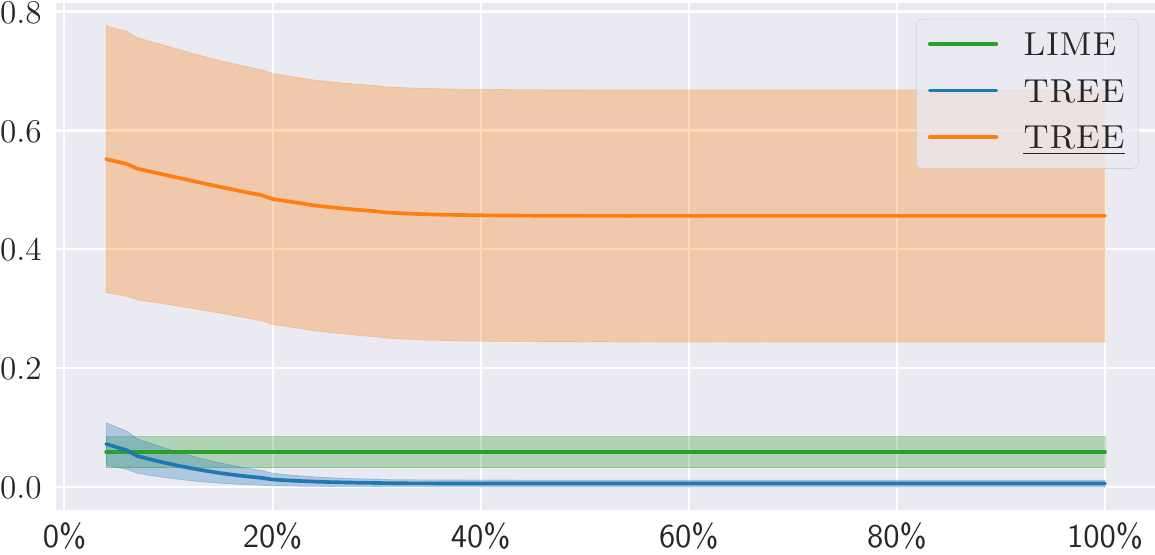}%
}\label{fig:tree_depth_fidelity-forest:ltl:1}}%
\par%
    \subfloat[{\sc LIMEtree} loss for the top 2 classes.]{%
\makebox[.49\fulllength][c]{%
    \includegraphics[width=.49\fulllength]{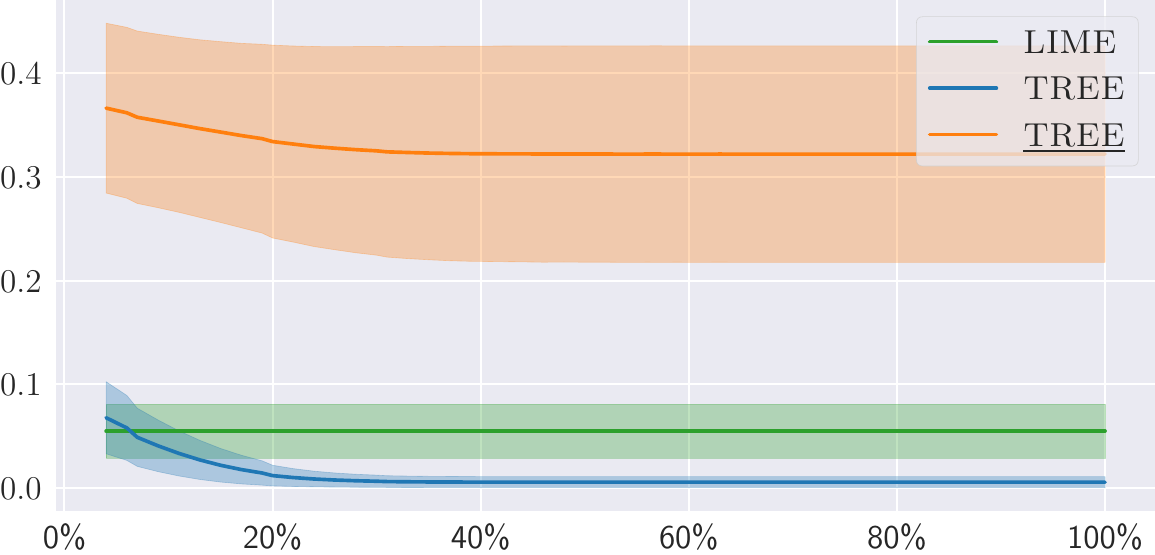}%
}\label{fig:tree_depth_fidelity-forest:ltl:2}}%
\par%
    \subfloat[{\sc LIMEtree} loss for the top 3 classes.]{%
\makebox[.49\fulllength][c]{%
    \includegraphics[width=.49\fulllength]{_experiments/forest-2500/loss-cls3-limet_weighted_Xrandom}%
}\label{fig:tree_depth_fidelity-forest:ltl:3}}%
\end{minipage}
\end{adjustwidth}
    \caption{%
    Fidelity of a surrogate \(\mathcal{L}\) (y-axis) built for the Forest Covertypes data set (see Section~\ref{sec:experiments} for information about the experimental setup) and plotted against its complexity \(\Omega\) (x-axis) expressed as the ratio between the depth of the tree and its maximum depth determined by the number of features in the interpretable representation, which is equivalent to the depth of a complete tree (Equation~\ref{eq:blimey_complexity}). %
    The caption of Figure~\ref{fig:tree_depth_fidelity} provides further information about the details of the plot. %
    \label{fig:tree_depth_fidelity-forest}}%
\end{figure}

Table~\ref{tab:fidelity} reports the fidelity loss of various {\sc LIMEtree} variants for fixed complexity levels (66\%, 75\% and 100\%). %
To better understand the relation between the complexity of the surrogate trees and their fidelity -- expanding the results shown in Figure~\ref{fig:loss} -- we plot these quantities in Figures~\ref{fig:tree_depth_fidelity}, \ref{fig:tree_depth_fidelity-cifar10}, \ref{fig:tree_depth_fidelity-cifar100}, \ref{fig:tree_depth_fidelity-wine-test} and \ref{fig:tree_depth_fidelity-forest} respectively for the ImageNet, CIFAR-10, CIFAR-100, Wine and Forest Covertypes data sets. %
It allows us to study how building surrogate trees of higher complexity influences their fidelity, and how these properties compare to the baseline given by linear surrogates (LIME) whose complexity is fixed. %
Since for the ImageNet, CIFAR-10 and CIFAR-100 data sets various images may have a different number of super-pixels, i.e., interpretable features, our formulation of the depth-based tree complexity \(\Omega\) given by Equation~\ref{eq:blimey_complexity} accounts for that by scaling the tree depth in relation to the number of segments -- this metric can be interpreted as \emph{tree completeness level}. %

\section[\appendixname~\thesection]{Examples of Diverse Explanation Types\label{apx:examples}}%

Building up on Sections~\ref{sec:limetree}, \ref{sec:experiments:desiderata} and \ref{sec:discussion}, %
this appendix offers a further discussion of various aspects of the six explanation types available for {\sc LIMEtree} %
using a concrete case study. %
To better communicate our method's explanatory power, we provide multiple examples for the top three classes predicted by a black box for the image shown in Figure~\ref{fig:lime:image} -- \emph{tennis ball} (99.28\%), \emph{golden retriever} (0.67\%) and \emph{Labrador retriever} (0.04\%) -- and compare them to the corresponding LIME explanations shown in Figure~\ref{fig:lime}.%

The LIME explanation for \emph{tennis ball} -- shown in Figure~\ref{fig:lime:ball} -- indicates that segment \#8, which depicts the ball, has an overwhelmingly positive influence on predicting this class. %
Figure~\ref{fig:lime:ball} also exemplifies the problem with high correlation of adjacent super-pixels: the next two most important segments are \#2 and \#7 -- they are neighbours of \#8 and mostly surround it -- which is likely because they include pixels that belong to the tennis ball object, e.g., the characteristic white stripe (\#2). %
The other two LIME explanations are for \emph{golden retriever} -- Figure~\ref{fig:lime:golden} -- and \emph{Labrador retriever} -- Figure~\ref{fig:lime:labrador}. %
In both cases the segment depicting the ball (\#8) has a large negative influence, which is expected, and the segment capturing the dog's face (\#3) has a large positive effect. %
Predicting between these two dog breeds is determined by the positive effect of segment \#1 on the \emph{golden retriever} class (maybe because it reveals the long coat) and the negative influence of segment \#2 on the \emph{Labrador retriever} class (possibly since it includes the white stripe of the tennis ball). %
Based on this evidence alone, it is difficult to determine the model's heuristic for telling apart these two classes; in particular, the role that segment \#2 plays.%

When it comes to {\sc LIMEtree}, we can easily calculate the \emph{importance} of interpretable features (\emph{Gini importance}~\cite{breiman2001random}) -- shown in Figure~\ref{fig:examples:fi} -- which closely resembles LIME insights. %
Since {\sc LIMEtree} models all three classes simultaneously, the importance captures the segments that help to differentiate between these classes. %
Comparing Figure~\ref{fig:examples:fi} with analogous LIME explanations shown in Figure~\ref{fig:lime} reveals a reassuring overlap, with each LIME explanation sharing at least two of its top four most important segments with the {\sc LIMEtree} explanation. %
The tree-based feature importance indicates that segment \#8 -- depicting the ball -- is the most important, owing to the dominant \emph{tennis ball} prediction (99.28\%), and is followed by segments \#1, \#3 and \#6 -- covering most of the dog. %
While informative, these insights cannot be explicitly attributed to any individual class and the feature importance values can only be positive, limiting their explanatory power.%

Since all {\sc LIMEtree} explanations are consistent -- they are derived from the same surrogate tree -- with some help of another explanation type, such as the \emph{tree structure visualisation} shown earlier in Figure~\ref{fig:mort}, we can discover the relation between each important feature -- Figure~\ref{fig:examples:fi} -- and the three explained classes. %
It is important to note that these two explanations are derived from different trees since the depth of the surrogate shown in Figure~\ref{fig:mort} was limited to two for visualisation purposes; %
this also means that we can achieve full fidelity with respect to model-driven explanations (Lemma~\ref{lemma:local_tree_fidelity}) but not data-driven explanations (Corollary~\ref{cor:local_data_fidelity}). %
Comparing the two leftmost leaves with the two rightmost ones -- the result of the root split on segment \#8 -- tells us that this segment has positive influence on the \emph{tennis ball} prediction; %
additionally, when segment \#7 is present this prediction is strengthened, nonetheless without it \emph{tennis ball} is still the most likely prediction. %
On the other hand, when the ball is absent, i.e., segment \#8 is occluded, both dog breeds are almost equally likely with the presence of segment \#3 being the deciding factor: it is \emph{Labrador retriever} if \#3 is occluded, and \emph{golden retriever} if \#3 is present (although \emph{Labrador retriever} is nearly equally likely in this case).%

\begin{figure}[t]
    \centering
    \subfloat[``Show me an example of a \emph{golden retriever}.'']{%
\makebox[0.2285\fulllength][c]{%
    \includegraphics[height=0.2244\fulllength]{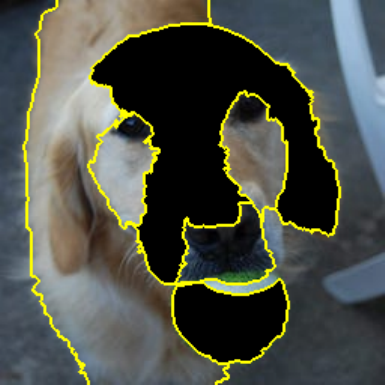}%
}\label{fig:lab_tree_exemplar_golden}}
    \hspace{0.03025\textwidth}%
    \subfloat[``Show me an example of a \emph{Labrador retriever}.'']{%
\makebox[0.2285\fulllength][c]{%
    \includegraphics[height=0.2244\fulllength]{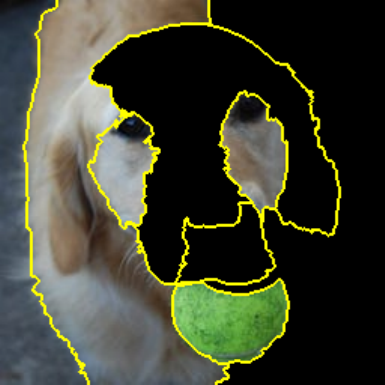}%
}\label{fig:lab_tree_exemplar_labrador}}%
    \caption{Two exemplar explanations of \protect\subref*{fig:lab_tree_exemplar_golden} a \emph{golden retriever} and \protect\subref*{fig:lab_tree_exemplar_labrador} a \emph{Labrador retriever} generated with {\sc LIMEtree}.\label{fig:lab_tree_exemplar}}%
\end{figure}

Arriving at these conclusions required us to inspect and reason over the tree structure, which cannot be expected of a lay explainee (as demonstrated by the results of our pilot user study reported in Section~\ref{sec:experiments:user}) or when the surrogate tree is large or complex. %
In such cases we can use other types of explanations, for example, \emph{what-if questions}. %
Since the tree presented in Figure~\ref{fig:mort} is not complete (see Lemma~\ref{lemma:local_tree_fidelity}), we use the black-box model instead of the surrogate to evaluate the hypothetical scenarios. %
Because segment \#8, depicting the ball, is the most important factor, we are interested in \emph{what if} this segment was not there; the new prediction is 97\% \emph{golden retriever} -- see Figure~\ref{fig:examples:cf}. %
We can also ask for \emph{exemplar} explanations of the \emph{golden retriever} and \emph{Labrador retriever} classes, which are shown in Figure~\ref{fig:lab_tree_exemplar}. %

\begin{figure}[t]
    \centering
    \subfloat[Preserving segments \#8 \& \#3 yields \emph{tennis ball} with 92\% confidence.]{%
\makebox[0.2285\fulllength][c]{%
    \includegraphics[height=0.2244\fulllength]{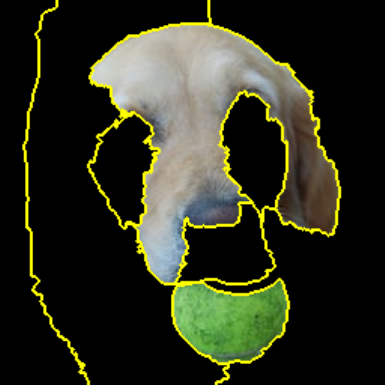}%
}\label{fig:lab_tree_ball_76}}%
    \hspace{0.03025\textwidth}%
    \subfloat[Preserving segments \#8 \& \#1 yields \emph{tennis ball} with 51\% confidence.]{%
\makebox[0.2285\fulllength][c]{%
    \includegraphics[height=0.2244\fulllength]{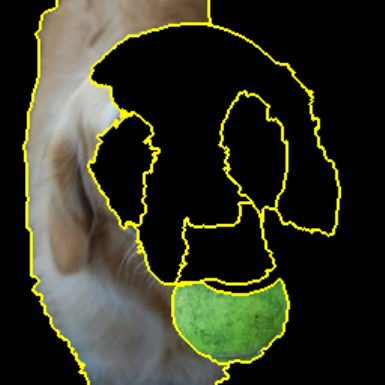}%
}\label{fig:lab_tree_ball_75}}%
    \caption{The two remaining shortest (highest number of occlusions) {\sc LIMEtree} explanations of \emph{tennis ball}; the other one -- where preserving segments \#8 \& \#7 yields \emph{tennis ball} with 97\% confidence -- is given in Figure~\ref{fig:examples:ex}.\label{fig:lab_tree_ball_left}}%
\end{figure}

In order to take full advantage of {\sc LIMEtree} explanations, we train a \emph{complete} surrogate tree (see Corollary~\ref{cor:local_data_fidelity}). %
We use it to retrieve the \emph{shortest} possible explanation, i.e., with the highest number of occluded segments, of \emph{tennis ball}. %
There are three such explanations of length two -- shown in Figures~\ref{fig:examples:ex} and \ref{fig:lab_tree_ball_left} -- with the following pairs of super-pixels preserved: \#7 \& \#8, \#3 \& \#8, and \#1 \& \#8. %
We can also generate rule explanations of \emph{Labrador retriever} based on the root-to-leaf paths found in the tree, selecting the one with the highest probability of this class. %
The resulting explanation is \(x^\prime_1 = 0 \land x^\prime_2 = 0 \land x^\prime_3 = 1 \land x^\prime_4 = 1 \land x^\prime_5 = 1 \land x^\prime_6 = 1 \land x^\prime_7 = 0 \land x^\prime_8 = 0\), yielding 98\% confidence. %
Such a representation is not particularly appealing, especially to a lay audience, but we can improve its comprehensibility by transforming it to the visual domain -- see Figure~\ref{fig:examples:rtl}.%

\begin{figure}[t]
    \centering
    \subfloat[How to get a \emph{golden retriever} prediction (54\%) without occluding segment \#8.]{%
\makebox[0.2285\fulllength][c]{%
    \includegraphics[height=0.2244\fulllength]{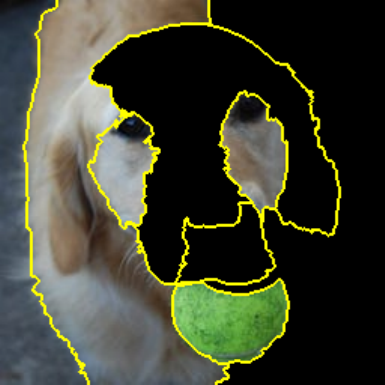}%
}\label{fig:lab_tree_custom_lab}}%
    \hspace{0.03025\textwidth}%
    \subfloat[Occluding segments \#8 \& \#1 yields \emph{tennis ball} (94\%) according to the black box.]{%
\makebox[0.2285\fulllength][c]{%
    \includegraphics[height=0.2244\fulllength]{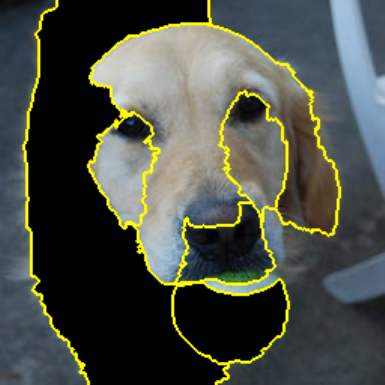}%
}\label{fig:lab_tree_custom_ball}}%
    \caption{Customised counterfactual explanations generated by {\sc LIMEtree}.\label{fig:lab_tree_custom}}%
\end{figure}

The biggest advantage of {\sc LIMEtree} is its ability to output \emph{counterfactual} explanations by using any method compatible with (regression) trees~\cite{sokol2021towards}. %
For example, we can ask the following question: ``\emph{Given} the presence of segment \#8 (the ball), what would have to change for the image to be classified as \emph{golden retriever}?'' %
Therefore, we are looking for an image occlusion pattern that preserves the ball segment (\#8) and yields \emph{golden retriever} prediction. %
{\sc LIMEtree} tells us that by discarding super-pixels \#2, \#3 and \#7 -- the smallest viable occlusion shown in Figure~\ref{fig:lab_tree_custom_lab} -- the model predicts \emph{golden retriever} (54\%). %
Since occluding segment \#8, i.e., the ball, results in 97\% \emph{golden retriever} -- see Figure~\ref{fig:examples:cf} -- another interesting question is: ``Had segment \#8 not been there, can the model still predict \emph{tennis ball}?'' %
{\sc LIMEtree} indicates that this can be achieved by occluding segments \#1 and \#8 as shown in Figure~\ref{fig:lab_tree_custom_ball}.%

\end{document}